%% file: main.tex
\documentclass[letterpaper]{article} 
\usepackage{aaai2026}  
\usepackage{times}  
\usepackage{helvet}  
\usepackage{courier}  
\usepackage[hyphens]{url}  
\usepackage{graphicx} 
\usepackage{natbib}  
\usepackage{caption} 
\usepackage{algorithm}
\usepackage{algorithmic}
\usepackage{amsmath}
\usepackage{amssymb}
\usepackage{xcolor}
\usepackage{multirow}
\usepackage{multicol}
\usepackage{graphicx}
\usepackage{pgfplots}
\usepackage{pgfplotstable}
\usepackage{tikz}
\usepackage{subcaption}
\usepackage{caption}
\usepackage{listings}
\usepackage[most]{tcolorbox}
\tcbuselibrary{listings, breakable}

\usepackage{bbding}

\usepackage{newfloat}
\usepackage{listings}
\DeclareCaptionStyle{ruled}{labelfont=normalfont,labelsep=colon,strut=off} 
\floatstyle{ruled}
\newfloat{listing}{tb}{lst}{}
\floatname{listing}{Listing}
\title{Towards Temporal Fusion Beyond the Field of View \\for Camera-based Semantic Scene Completion}
\author{
    Jongseong Bae\equalcontrib,
    Junwoo Ha\equalcontrib,
    Jinnyeong Heo\equalcontrib,
    Yeongin Lee\equalcontrib,
    Ha Young Kim\thanks{Corresponding author.}
}
\affiliations{
    Yonsei University\\

    \{js.bae,gkwnsdn0402,wlssud132,zeroin,hayoung.kim\}@yonsei.ac.kr
}

\usepackage{bibentry}
\usepackage[table,xcdraw]{xcolor}
\usepackage{amssymb}
\usepackage{booktabs}
\usepackage{graphicx}
\usepackage{tikz}
\usepackage{rotating}
\usepackage{array}
\usepackage{adjustbox}
\usepackage{hyperref}

\begin{document}

\maketitle

\begin{abstract}
Recent camera-based 3D semantic scene completion (SSC) methods have increasingly explored leveraging temporal cues to enrich the features of the current frame.
However, while these approaches primarily focus on enhancing in-frame regions, they often struggle to reconstruct critical out-of-frame areas near the sides of the ego-vehicle, although previous frames commonly contain valuable contextual information about these unseen regions.
To address this limitation, we propose the Current-Centric Contextual 3D Fusion (C3DFusion) module, which generates hidden region-aware 3D feature geometry by explicitly aligning 3D-lifted point features from both current and historical frames.
C3DFusion performs enhanced temporal fusion through two complementary techniques—historical context blurring and current-centric feature densification—which suppress noise from inaccurately warped historical point features by attenuating their scale, and enhance current point features by increasing their volumetric contribution.
Simply integrated into standard SSC architectures, C3DFusion demonstrates strong effectiveness, significantly outperforming state-of-the-art methods on the SemanticKITTI and SSCBench-KITTI-360 datasets. 
Furthermore, it exhibits robust generalization, achieving notable performance gains when applied to other baseline models.
\href{https://heo-jinnyeong.github.io/Towards-Temporal-Fusion-Beyond-the-Field-of-View-for-Camera-based-Semantic-Scene-Completion-AAAI2026/}{\textcolor{cyan}{Project Page.}}
\end{abstract}

\input{Sections/1_Introduction}
\input{Sections/2_Related_Work}
\input{Sections/3_Method}
\input{Sections/4_Experiments}
\input{Sections/5_Conclusion}

\section{Acknowledgments}
This work was supported by the National Research Foundation (NRF) grant funded by Korea Government (MSIT) (No. 2023R1A2C200337911) and the Ministry of Education of the Republic of Korea and the National Research Foundation of Korea (NRF-2024S1A5C3A03046579).

\bibliography{aaai2026}
\input{Sections/99_Appendix}

\end{document}

%% file: Sections/1_Introduction.tex
\section{Introduction}
\label{sec:intro}
3D semantic scene completion (SSC)~\cite{song2017semantic} has recently garnered significant attention as a fundamental 3D perception task, particularly in applications such as autonomous driving~\cite{hu2023planning}.  
SSC aims to simultaneously reconstruct voxelized 3D geometry and predict semantic labels for each voxel—an inherently challenging task, especially for real-world deployment.
Existing SSC methods are typically categorized by input modality.
LiDAR-based methods~\cite{xia2023scpnet,cheng2021s3cnet,yan2021sparse,yang2021semantic} have demonstrated superior performance and thus become the dominant paradigm; however, their widespread adoption is hindered by the high cost and limited scalability of LiDAR sensors.  
In contrast, camera-based methods have recently gained momentum, with rapid advances significantly narrowing the performance gap compared to their LiDAR-based counterparts.
\begin{figure}[t]
    \centering
    \includegraphics[width=1\linewidth]{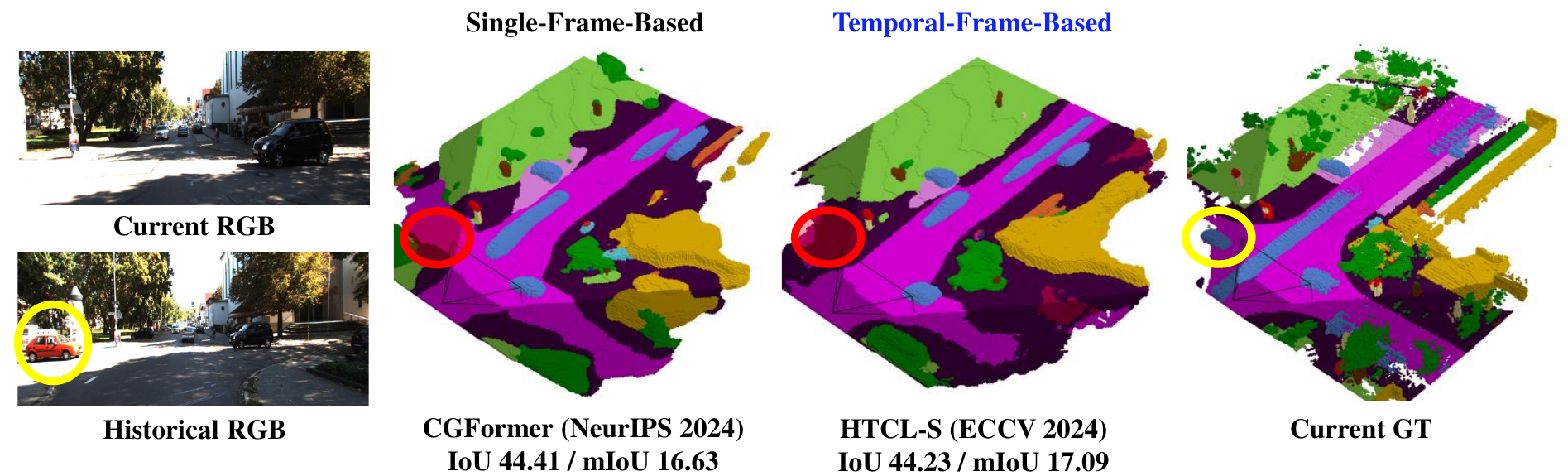}
    \caption{Existing temporal fusion models struggle to complete out-of-frame geometry in the current frame. For example, HTCL-S~\cite{li2024hierarchical}, a recent method that performs temporal fusion via 2D feature warping, fails to recover the car on the left side despite its visibility in previous frames, resulting in performance comparable to that of the single-frame-based CGFormer~\cite{yu2024context}.}
    \label{fig1}
\end{figure}

Boosted by the release of the SemanticKITTI benchmark~\cite{behley2019semantickitti}, camera-based SSC methods~\cite{li2023voxformer,huang2023tri,zhang2023occformer,miao2023occdepth,jiang2024symphonize} have been actively explored, driving continued progress in this field.  
In particular, given the temporally continuous nature of input in driving scenarios, numerous recent approaches~\cite{li2024hierarchical,ye2024cvt,li2024hierarchical1,wang2025learning,lu2025one} leverage sequences of past frames to enrich the current frame’s feature representation and improve performance.  
However, while these methods primarily focus on in-camera-view geometry from the current frame, they tend to overlook out-of-camera-view regions—blind spots that are often located near the ego-vehicle and therefore require especially precise perception for safe navigation.
Although past frames provide broader spatial context around the current viewpoint, existing methods still rely heavily on extrapolating from visible geometry to infer these blind spots, often resulting in limited accuracy (see Figure~\ref{fig1}).

In this paper, we propose an intuitive yet effective temporal geometry fusion method, Current-Centric Contextual 3D Fusion (C3DFusion), which addresses the aforementioned challenge by explicitly integrating historical and current features in 3D space.
Specifically, we map 2D features from all frames directly onto their corresponding lifted 3D points via backprojection, and align the lifted point features from historical frames to the current frame’s metric space through geometric warping using known camera poses.
To effectively fuse the historical geometry with the current geometry while mitigating geometric discrepancies between them, we introduce two complementary techniques: historical context blurring and current-centric feature densification.
The former attenuates the scale of historical point features according to their depth, reducing the influence of potentially inaccurate geometry arising from distance-dependent depth estimation errors.
The latter densifies the current lifted points by interpolating over the corresponding 2D feature and depth maps prior to backprojection, thereby enhancing their volumetric contribution to the fused geometric representation.

Our extensive experiments demonstrate the effectiveness of the proposed C3DFusion both quantitatively and qualitatively, particularly against existing camera-based SSC methods that incorporate temporal fusion.
Upon integration into a standard camera-based SSC architecture, our method achieves state-of-the-art (SOTA) performance, with IoU and mIoU scores of 47.62 and 18.98 on SemanticKITTI, and 49.28 and 21.74 on SSCBench-KITTI-360~\cite{li2024sscbenchlargescale3dsemantic}, respectively, significantly outperforming prior approaches.
Furthermore, incorporating C3DFusion into other baseline models yields consistent performance improvements, demonstrating its strong generalization capability.

Our key contributions are as follows:
\begin{itemize}
    \item To the best of our knowledge, we are the first to primarily address out-of-camera-view completion via temporal cues in camera-based SSC—a long-overlooked yet crucial factor in safety-critical driving applications.
    \item We propose a simple yet effective temporal geometry fusion method, C3DFusion, that performs perspective alignment between explicitly point-mapped historical and current frame features in the 3D metric space of the current frame.
    \item To further enhance temporal fusion, we introduce two refinement techniques—historical context blurring and current-centric feature densification—that reduce noise caused by geometric inaccuracies in warped historical features and improve geometric fidelity by emphasizing information from the current frame.
    \item Built on C3DFusion, our camera-based SSC model achieves strong SOTA performance on the SemanticKITTI and SSCBench-KITTI-360 benchmarks, while demonstrating robust generalization across diverse existing architectures.
\end{itemize}

%% file: Sections/2_Related_Work.tex
\section{Related Work}
\subsubsection{3D Semantic Scene Completion}
Since the introduction of the SSC task, early methods~\cite{chen20203d,li2020anisotropic,zhang2018efficient} focused primarily on indoor environments, using datasets such as NYUv2~\cite{silberman2012indoor}.
The release of the large-scale SemanticKITTI benchmark subsequently catalyzed research on outdoor SSC, sparking a surge of LiDAR-based approaches~\cite{roldao2020lmscnet,cheng2021s3cnet,yan2021sparse,yang2021semantic,xia2023scpnet}, which have since become the dominant paradigm.
Recently, camera-based methods have gained momentum due to the affordability and rich contextual information offered by RGB sensors.
MonoScene~\cite{cao2022monoscene} projects 2D features along optical rays for voxel-wise prediction, while TPVFormer~\cite{huang2023tri} lifts features onto multiple planes to capture diverse spatial perspectives.
OccFormer~\cite{zhang2023occformer} adopts a Lift-Splat (LSS)\cite{philion2020lift}-like strategy to construct volumetric context. 
VoxFormer~\cite{li2023voxformer} introduces an MAE-like~\cite{he2022masked} architecture with deformable attention, and Symphonies~\cite{jiang2024symphonize} models instance-level representations via learnable queries.
SOTA models such as CGFormer~\cite{yu2024context}, ScanSSC~\cite{bae2025three}, and L2COcc~\cite{wang2025l2cocc} further combine LSS-style feature lifting with deformable attention to enhance semantic reasoning.
Meanwhile, the recent release of the Occ3D-nuScenes benchmark~\cite{tian2023occ3d} has spurred a parallel line of research~\cite{huang2021bevdet,xu2024regulating,ma2024cotr,li2023fb,kim2025protoocc}, which leverages multi-camera systems as input.
In this work, we focus on the single-view RGB setting and aim to overcome its limited field of view by leveraging temporal cues, while proposing an alternative to the prevalent LSS-style feature lifting paradigm.

\subsubsection{Temporal Fusion for Camera-based 3D Perception}
A widely adopted strategy in 3D perception tasks such as object detection is to perform temporal fusion directly in the BEV feature space, either via attention mechanisms~\cite{li2024bevformer} or by warping and concatenating features across time steps~\cite{huang2022bevdet4d,wang2024panoocc,yang2023bevformer}.
In the context of SSC, various temporal fusion techniques have been proposed. 
VoxFormer-T~\cite{li2023voxformer} and SGN~\cite{mei2024camera} extend attention mechanisms to integrate features across multiple frames. HTCL-S~\cite{li2024hierarchical} and Hi-SOP~\cite{li2024hierarchical1} utilize contextual pattern affinity to temporally align features from current and past RGB frames within the 2D feature space. 
FlowScene~\cite{wang2025learning} employs optical flow between adjacent frames to guide occlusion correction and improve voxel refinement. 
CVT-Occ~\cite{ye2024cvt}, on the other hand, enhances the volumetric representation of the current frame by constructing a cost volume across temporal frames, while CF-SSC~\cite{lu2025one} takes a different approach by synthesizing future frames from past observations to better recover occluded regions.
These approaches largely concentrate on regions within the current camera's field of view and often neglect the reconstruction of out-of-camera-view areas.
In contrast, our method explicitly targets these unseen regions by projecting 3D points from both historical and current frames into a unified target space and processing them in a single pass to generate coherent voxel features.

%% file: Sections/3_Method.tex
\section{Method}
\begin{figure*}[hbt]
    \centering
    \includegraphics[width=1\linewidth]{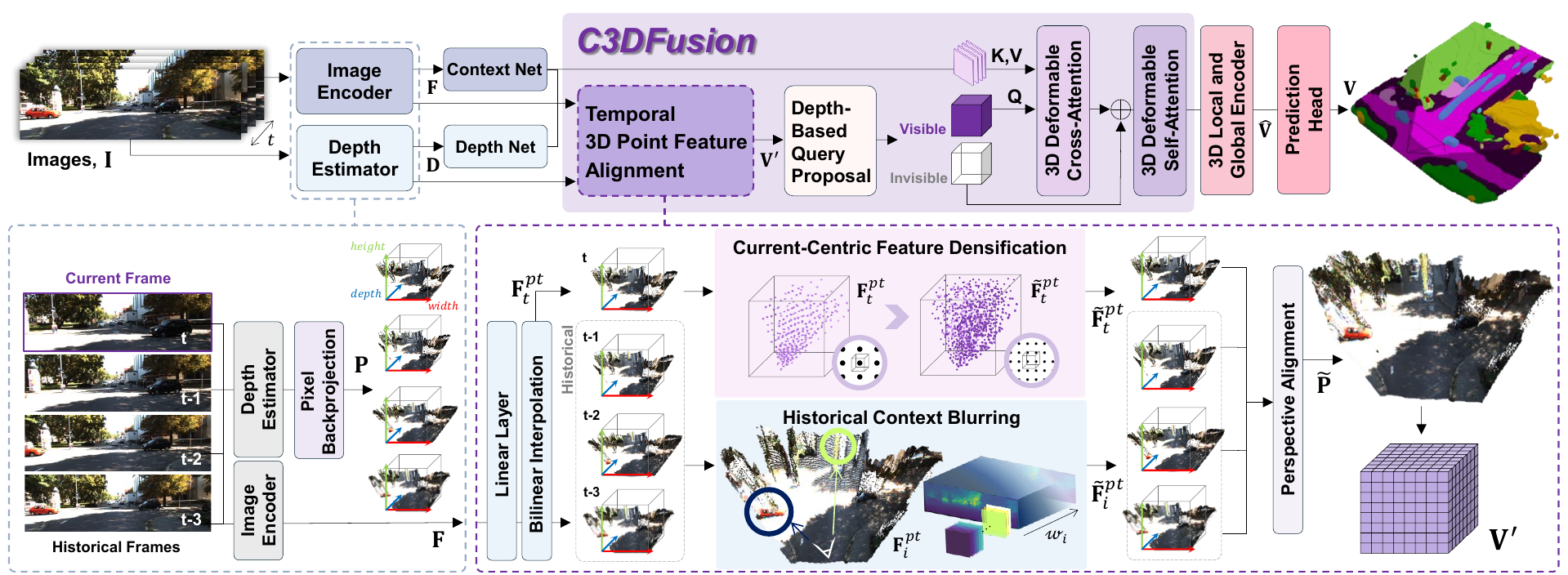}
    \caption{An overview of our model, highlighting the proposed C3DFusion. The symbol `$\oplus$' denotes feature concatenation.}
    \label{fig2}
\end{figure*}
\subsection{Overview}
Given a sequence of $n$ RGB images $\mathbf{I} = \{\mathbf{I}_i \in \mathbb{R}^{H \times W \times 3} \mid i = t - n + 1, \dots, t \}$ with resolution $(H, W)$, we aim to predict voxel-wise semantic class probabilities at time $t$ within a voxelized 3D space $\mathbf{V} \in \mathbb{R}^{X \times Y \times Z \times P}$, where $(X, Y, Z)$ denotes the spatial resolution of the volume and $P$ is the number of semantic classes, including the empty class.
Our model follows the standard architectural paradigm of modern camera-based SSC~\cite{jiang2024symphonize,yu2024context,li2024hierarchical,wang2025l2cocc}, consisting of three main components: viewing transformation, voxel processing, and semantic prediction.
Among these, the proposed C3DFusion primarily focuses on the viewing transformation stage, which plays a pivotal role in overall SSC performance, as it performs both 2D-to-3D lifting and temporal geometry fusion.
An architectural overview is provided in Figure~\ref{fig2}.

\subsection{C3DFusion}
\subsubsection{Temporal 3D Point Feature Alignment}
In camera-based SSC, mapping 2D image features into 3D space is a fundamental step.  
To achieve this, numerous methods adopt backprojection techniques in conjunction with off-the-shelf depth estimators.
Early methods~\cite{li2023voxformer,jiang2024symphonize} primarily use the resulting 3D points as voxel proposals for cross-attention with 2D image features.
More recent methods~\cite{zhang2023occformer,yu2024context,wang2025l2cocc} further advance this process by constructing volumetric geometry using the LSS strategy, which effectively generates a frustum-shaped dense feature volume in 3D space.
Since our approach targets temporal alignment in 3D space, a natural extension would be to apply temporal fusion directly to such volumetric features.
However, we hypothesize that when extended across multiple frames, these sparsely densified and long-tailed features introduce geometric noise that ultimately degrades semantic prediction accuracy—an effect we quantitatively validate in our experiments.
To handle this issue, we instead adopt a strategy that directly maps 2D image features onto 3D lifted point clouds from both historical and current frames, and appropriately aggregates them with a focus on the current perspective.

From the input image sequence $\mathbf{I}$, we extract 2D image features $\mathbf{F} = \{\mathbf{F}_i \in \mathbb{R}^{H' \times W' \times C}\}$ and corresponding depth maps $\mathbf{D} = \{\mathbf{D}_i \in \mathbb{R}^{H \times W}\}$ using an image encoder and a pretrained depth estimator, where $(H', W')$ denotes the spatial resolution of the features and $C$ is the feature dimension.  
Using the depth maps with known camera parameters, each image pixel is backprojected into 3D space to obtain a set of point clouds $\mathbf{P} = \{\mathbf{P}_i \in \mathbb{R}^{HW \times 3}\}$, where each point cloud is defined in its corresponding metric 3D coordinate system.
To obtain corresponding point features $\mathbf{F}^{pt} = \{\mathbf{F}^{pt}_i \in \mathbb{R}^{HW \times C'}\}$, we first apply a linear layer to each feature map $\mathbf{F}_i$, followed by bilinear interpolation to match the resolution of the depth maps, as follows:
\begin{equation}
\mathbf{F}^{pt}_i = \text{Flatten}(\text{Bilinear}(\text{Linear}(\mathbf{F}_i), (H, W))),
\end{equation}
where $\text{Bilinear}(\cdot, (a,b))$ denotes bilinear interpolation to resolution $(a,b)$, and $\text{Flatten}(\cdot)$ flattens the spatial dimensions of the feature map.

Given the relative poses derived from the extrinsic matrices, the historical points $\{\mathbf{P}_{t-n+1}, \dots, \mathbf{P}_{t-1}\}$ are warped into the current frame’s coordinate system, resulting in $\{\mathbf{\tilde{P}}_{t-n+1}, \dots, \mathbf{\tilde{P}}_{t-1}\}$, for fusion with the current points $\mathbf{P}_t$.  
Instead of directly merging them, we propose two techniques to improve the quality of the fused geometry: historical context blurring and current-centric feature densification.

\subsubsection{Historical Context Blurring}
Assuming the ego-vehicle moves continuously forward, the unfiltered warped historical points that remain within the current coordinate system tend to originate from farther regions in the original camera view—since nearer regions have already been passed—and therefore typically have larger depth values.
Since we use backprojection based on estimated depth maps, the accuracy of the lifted 3D geometry inherently depends on the quality of the depth estimator, which typically degrades at greater depths~\cite{poggi2020uncertainty}.
To mitigate geometric discrepancies arising from this limitation, we apply a regularization that scales the magnitude of historical point features inversely proportional to their estimated depth values.
Specifically, we apply min-max normalization to each historical depth map $\{\mathbf{D}_{t-n+1}, \dots, \mathbf{D}_{t-1}\}$ independently, subtract the normalized values from 1, and employ the resulting values as per-point weights $w_i \in \mathbb{R}^{H\times W}$ in the range $[0,1]$:
\begin{equation}
w_i = 1 - \text{MinMax}(\mathbf{D}_i),
\end{equation}
where $\text{MinMax}(\cdot)$ denotes min-max normalization.
By element-wise multiplying these weights with the point features $\mathbf{F}_i^{pt}$, we obtain the rescaled features $\mathbf{\Tilde{F}}_i^{pt}$, which are finally associated with their corresponding warped points $\mathbf{\tilde{P}}_{i}$:
\begin{equation}
\mathbf{\Tilde{F}}_i^{pt} = w_i \odot \mathbf{F}_i^{pt},
\end{equation}
where $\odot$ denotes element-wise multiplication.

\subsubsection{Current-Centric Feature Densification}
For in-camera-view regions of the current frame, there is substantial overlap with preceding frames, resulting in numerous points within these areas in the fused 3D space.
However, since each frame contributes a fixed number of points equal to the image grid size $HW$, the influence of the current frame’s features may become diluted during temporal aggregation—despite their greater temporal relevance.
To emphasize the current frame's contextual information in these regions, we increase the density of its point cloud by bilinearly interpolating the point feature $\mathbf{F}^{pt}_t$ and depth map $\mathbf{D}_t$ prior to backprojection, yielding densified current points $\mathbf{\tilde{P}}_t$:
\begin{gather}
\tilde{\mathbf{D}}_t = \text{Bilinear}(\mathbf{D}_t, (\tilde{H}, \tilde{W})),\\
\tilde{\mathbf{F}}^{pt}_t = \text{Flatten}(\text{Bilinear}(\mathbf{F}^{pt\rightarrow \text{grid}}_t, (\tilde{H}, \tilde{W}))),
\end{gather}
where $\mathbf{F}^{pt\rightarrow \text{grid}}_t$ denotes $\mathbf{F}^{pt}_t$ rearranged into its original spatial grid form $(H, W, C')$ before interpolation.
By default, we set the upsampling resolution to $(\tilde{H}, \tilde{W}) = (2H, 2W)$.

\subsubsection{Voxel Aggregation}
Given the temporally aligned point clouds from multiple frames, we define the unified point set as
$\mathbf{\tilde{P}} = \bigcup_{i=t-n+1}^{t} \mathbf{\tilde{P}}_i$, and the corresponding set of refined point features as
$\mathbf{\tilde{F}}^{pt} = \bigcup_{i=t-n+1}^{t} \mathbf{\tilde{F}}^{pt}_i$. We then discard any points that fall outside the predefined spatial boundaries of the current target voxel grid.
The remaining points are voxelized into the volume features $\mathbf{V}' \in \mathbb{R}^{X' \times Y' \times Z' \times C'}$.
For each voxel at position $(x, y, z)$ in $\mathbf{V}'$, its feature is computed by summing all point features within the voxel and dividing by the number of frames $n$ as:
\begin{equation}
\mathbf{V}'(x,y,z) = 
\begin{cases}
\frac{1}{n} \sum_{j=1}^v \mathbf{f}_{j}, & \text{if } v > 0 \\
\mathbf{0}, & \text{if } v = 0
\end{cases}
\end{equation}
where \(v\) is the number of points falling into the voxel, and \(\{\mathbf{f}_{1}, \dots, \mathbf{f}_{v}\} \subset \mathbf{\tilde{F}}^{pt}\) are the associated features.

\subsubsection{MAE-like Voxel Refinement}
Following the milestone paradigm of voxel proposal-based refinement~\cite{li2023voxformer,jiang2024symphonize,yu2024context,wang2025l2cocc}, initially occupied voxels in $\mathbf{V}'$ first undergo cross-attention to supplement lifted features with additional 2D context, while unoccupied voxels are refined via self-attention to extrapolate missing regions.
As we use 3D deformable attention~\cite{li2023dfa3d} for the cross-attention stage, current 2D image feature $\mathbf{F}_t$ and depth map $\mathbf{D}_t$ are passed through additional context and depth networks~\cite{zhang2023occformer,yu2024context, wang2025l2cocc}, yielding $\mathbf{F}_{\text{cross}} \in \mathbb{R}^{H' \times W' \times C}$ and $\mathbf{D}_{\text{cross}} \in \mathbb{R}^{H' \times W' \times B}$, where $B$ denotes the number of discretized depth bins.
These are then fed into the cross-attention module, whose output is subsequently processed by the self-attention module as follows:
\begin{gather}
    \mathbf{V}'_{\text{cross}} = \text{DeformCross}(\mathbf{V}', \mathbf{F}_{\text{cross}}, \mathbf{M}_{\text{cross}},  \mathbf{D}_{\text{cross}}),\\
    \mathbf{V}'_{\text{self}} = \text{DeformSelf}(\mathbf{V}'_{\text{cross}}, \mathbf{V}'_{\text{cross}}, \mathbf{M}_{\text{self}}),
\end{gather}
where $\text{DeformCross}(\cdot,\cdot,\cdot,\cdot)$ and $\text{DeformSelf}(\cdot,\cdot,\cdot)$ denote 3D deformable cross- and self-attention modules, taking query, key-value, and attention mask as inputs. DeformCross additionally leverages discretized depth probability.
$\mathbf{M}_{\text{cross}}$ and $\mathbf{M}_{\text{self}}$ are binary masks indicating the occupied and unoccupied voxels in $\mathbf{V}'$, respectively.

\subsection{Voxel Processing}
Once the viewing-transformed voxel feature $\text{V}'_{\text{self}}$ is obtained, we further process it using a voxel backbone network to capture geometric patterns across multiple spatial scales within the voxel space.
For this stage, we adopt the voxel processing architecture from CGFormer, a recent approach in single-frame camera-based SSC.

Specifically, given $\mathbf{V}'_{\text{self}}$, the voxel processing network is divided into two branches:
a voxel-based branch serving as a local encoder implemented with a 3D ResNet~\cite{he2016deep}, and
a TPV-based branch serving as a global encoder implemented with a 2D Swin Transformer~\cite{liu2021swin}:
\begin{gather}
\mathbf{V}'_{\text{vox}} = \text{ResNet3D}(\mathbf{V}'_{\text{self}}), \\
\mathbf{V}'_{\text{tpv}} = \left\{ \text{Swin}\left(\text{Pool}(\mathbf{V}'_{\text{self}}, \text{dim})\right) \mid \text{dim} \in \{xy, yz, zx\} \right\},
\end{gather}
where $\text{Pool}(\cdot)$ denotes a max-pooling operation along the specified dimension, resulting in a 2D feature map on the corresponding plane.

The outputs $\mathbf{V}'_{\text{vox}} \in \mathbb{R}^{X' \times Y' \times Z' \times C'}$ and $\mathbf{V}'_{\text{tpv}} = \{\mathbf{V}'_{\text{xy}}\in \mathbb{R}^{X' \times Y' \times 1 \times C'}, \mathbf{V}'_{\text{yz}}\in \mathbb{R}^{1 \times Y' \times Z' \times C'}, \mathbf{V}'_{\text{zx}}\in \mathbb{R}^{X' \times 1 \times Z' \times C'}\}$ are aggregated via weighted summation to produce the final voxel feature $\hat{\mathbf{V}}$, where the weight $\mathbf{W} \in \mathbb{R}^{X' \times Y' \times Z' \times 4}$ is generated from $\mathbf{V}'_{\text{vox}}$ using a linear layer followed by a softmax along the channel dimension:
\begin{gather}
\mathbf{W} = \text{Softmax}(\text{Linear}(\mathbf{V}'_{\text{vox}})),\\
\hat{\mathbf{V}} = \sum_{k=1}^{4} \mathbf{W}_k \odot \mathbf{V}'_k,
\end{gather}
where $\mathbf{W}_k \in \mathbb{R}^{X' \times Y' \times Z' \times 1}$ is the $k$-th slice of $\mathbf{W}$, and $\mathbf{V}'_k$ is one of the intermediate voxel features from the set $\{\mathbf{V}'_{\text{vox}}, \mathbf{V}'_{\text{xy}}, \mathbf{V}'_{\text{yz}}, \mathbf{V}'_{\text{zx}}\}$.

\subsection{Semantic Prediction}
The processed 3D feature volume $\mathbf{\hat{V}}$ is finally fed into the semantic prediction head, which consists of a 3D convolutional layer, followed by normalization and a linear projection.
This produces voxel-wise class logits $\mathbf{V}^{logit} \in \mathbb{R}^{X' \times Y' \times Z' \times P}$.
To obtain the final semantic prediction volume $\mathbf{V}$, the logits $\mathbf{V}^{logit}$ are first upsampled via trilinear interpolation and then passed through a softmax function.
This process can be summarized as:
\begin{gather}
\mathbf{V}^{logit} = \text{Linear}(\text{Norm}(\text{Conv3D}({\mathbf{\hat{V}}}))), \\
\mathbf{V} = \text{Softmax}(\text{Trilinear}(\mathbf{V}^{logit}, (X,Y,Z))),
\end{gather}
where $\text{Trilinear}(\cdot, (a,b,c))$ denotes trilinear interpolation to resolution $(a,b,c)$.

\subsection{Training Loss}
Following prior works~\cite{yu2024context,bae2025three,wang2025l2cocc}, we employ a combination of four losses: cross-entropy loss $\mathcal{L}_{ce}$, affinity losses $\mathcal{L}_{scal}^{geo}$ and $\mathcal{L}_{scal}^{sem}$, and depth loss $\mathcal{L}_d$.
The total loss $\mathcal{L}$ is defined as:
\begin{align}
\mathcal{L}=\lambda_{ce}\mathcal{L}_{ce}+\lambda_{scal}^{geo}\mathcal{L}_{scal}^{geo}+\lambda_{scal}^{sem}\mathcal{L}_{scal}^{sem}+\lambda_d\mathcal{L}_d,
\end{align}
where $\lambda_{ce}=\lambda_{scal}^{geo}=\lambda_{scal}^{sem}=1$, and $\lambda_d=0.001$.

%% file: Sections/4_Experiments.tex
\section{Experiments}
\subsection{Experimental Settings}
\subsubsection{Datasets}
We conduct experiments on two widely used SSC benchmarks: SemanticKITTI and SSCBench-KITTI360.
SemanticKITTI consists of 22 outdoor driving scenes, offering both LiDAR sweeps and stereo RGB images.
It is partitioned into 10 training scenes, 1 validation scene, and 11 test scenes.
The ground truth voxel grid has a resolution of $256\times 256\times 32$, covering a spatial extent of $51.2m\times 51.2m\times 6.4m$, meaning each voxel corresponds to a volume of $0.2m\times 0.2m\times 0.2m$.
In total, 21 semantic classes are defined: 19 semantic categories, 1 empty class, and 1 unknown class.
SSCBench-KITTI360 shares the same spatial coverage and ground truth voxel grid resolution as SemanticKITTI.
The dataset comprises 9 driving scenes, partitioned into 7 for training, 1 for validation, and 1 for testing.
In total, 20 semantic classes are defined, including 18 semantic categories, 1 free class, and 1 unknown class.

\subsubsection{Metrics}
Following widely adopted practices, we employ Intersection over Union (IoU) and mean IoU (mIoU) as quantitative evaluation metrics: IoU measures class-agnostic scene completion accuracy, while mIoU captures class-specific SSC performance.

\subsection{Performance Comparisons}
\subsubsection{Quantitative Results}
\input{Sections/Table/results_SemanticKITTI_test}
\input{Sections/Table/results_KITTI360}
The quantitative performance comparison between our proposed model and existing methods on the SemanticKITTI benchmark is presented in Table~\ref{tab:SemKITTI_test}.
The results clearly demonstrate the effectiveness of our approach, which achieves SOTA performance with an IoU of 47.62 and an mIoU of 18.98, significantly surpassing previous methods on both metrics.
We attribute this performance gain to the enhanced completion of out-of-camera-view regions—an area that constitutes the core focus of this work.
To verify this, we separately evaluate performance on these regions, as shown in the rightmost columns of the table.
Our model achieves particularly notable improvements in these challenging areas, reaching an IoU of 44.37 and an mIoU of 17.17, while other methods perform significantly worse on the same regions. 
These results validate our core hypothesis that conventional methods underestimate out-of-view (OOV) geometry, which our method successfully completes.

In addition, to demonstrate the generalizability of our method, we report results on the SSCBench-KITTI360 benchmark in Table~\ref{tab:KITTI360_test}.
Our model again achieves strong state-of-the-art performance, with an IoU of 49.28 and an mIoU of 21.74, while also delivering significant improvements in OOV regions, achieving 52.41 IoU and 17.17 mIoU.
These results further exhibit the effectiveness of our approach across different datasets.

\subsubsection{Qualitative Results}
\begin{figure*}[hbt]
    \centering
    \includegraphics[width=1\linewidth]{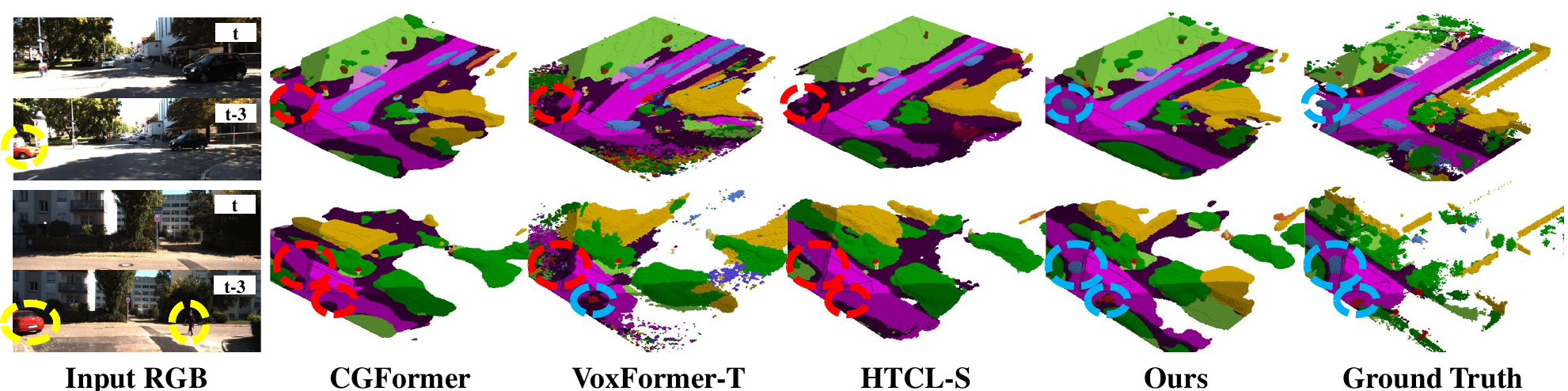}
    \caption{Visual comparison of our model against other recent camera-based methods on the SemanticKITTI validation set.}
    \label{qualitative}
\end{figure*}

Figure~\ref{qualitative} presents a qualitative comparison between our method and existing open-source camera-based approaches on the SemanticKITTI validation set. In scenarios where critical semantic objects, such as cars and persons, are invisible but located very close to the ego-vehicle in the current frame, temporal-frame-based baseline methods like VoxFormer-T and HTCL-S fail to recover these OOV regions, showing performance comparable to CGFormer, which relies only on a single frame. In contrast, our approach successfully reconstructs the missing structures, demonstrating effective use of temporal cues to accurately capture their spatial context even in the absence of direct visual observations in the current frame.

\subsection{Ablation Studies}
All ablation experiments are conducted on the validation split of the SemanticKITTI benchmark.

\subsubsection{C3DFusion}
Table~\ref{tab:ablation_architectur} presents the ablation study of the main components proposed in C3DFusion.
For the baseline, we extend CGFormer to process temporal frames by aligning each LSS-style feature volume using perspective warping, followed by a summation of the aligned volumes.
Notably, comparing the baseline with variant (a) quantitatively validates our hypothesis from the method section regarding simple temporal LSS feature fusion: replacing LSS feature alignment with point-based geometric temporal fusion results in significant improvements of 0.5 IoU and 1.87 mIoU.
Further comparisons between (a) and (b) or (c) highlight the individual contributions of the proposed historical context blurring and current-centric feature densification components to the performance gains, and applying the complete C3DFusion framework ultimately achieves the best IoU and mIoU scores, clearly demonstrating its effectiveness.

Additionally, to assess the practical generalizability of C3DFusion, we conduct further experiments by integrating it with several milestone camera-based SSC methods, as shown in Table~\ref{tab:ablation_generalization}.
For the baseline models, we select originally single-frame methods that adopt the backprojection technique—VoxFormer-S, which uses it solely for voxel proposals, and others that apply it for LSS-based 3D lifting. We extend them to temporal settings by first applying temporal LSS fusion (as in Table~\ref{tab:ablation_architectur}), then replace it with C3DFusion for performance comparison.
The results show that incorporating C3DFusion consistently leads to significant performance improvements across all models. 
While temporal LSS fusion often shows limited effectiveness in terms of mIoU—particularly with models such as VoxFormer-S and CGFormer—C3DFusion reliably boosts mIoU across various architectures. 
Remarkably, methods such as OccFormer and ScanSSC, when integrated with C3DFusion, surpass the previous SOTA mIoU of 18.22 achieved by L2COcc. 
These findings underscore the strong generalizability of C3DFusion across diverse architectures and its practical utility within camera-based SSC pipelines.

\input{Sections/Table/ablation_architecture}
\input{Sections/Table/ablation_generalization}
\input{Sections/Table/ablation_interpolation}
\input{Sections/Table/ablation_depth_gate}
\input{Sections/Table/ablation_N_of_frames}

\subsubsection{Intensity of Current-Centric Feature Densification}
Table~\ref{tab:ablation_interpolation} presents the ablation study on the intensity of current-centric feature densification by varying the target resolution for point feature interpolation. 
The best performance is achieved with a 2$\times$ interpolation factor, while further increasing the resolution results in a progressive degradation of mIoU. 
We attribute this decline primarily to geometric inaccuracies arising from the interpolation of the estimated depth map—excessive upsampling amplifies depth estimation errors, leading to less reliable geometric representations and ultimately harming performance. 

\subsubsection{Context Blurring to Current Frame}
We conduct an ablation study on extending historical context blurring to the current frame, as shown in Table~\ref{tab:ablation_depth_gate}. 
As expected, this extension results in a substantial drop in both IoU and mIoU, underscoring the critical importance of preserving current point features for accurate scene completion—even though these features also inherit geometric inaccuracies arising from the distance-dependent errors in estimated depth.

\subsubsection{Number of Frames in Sequence}
Table~\ref{tab:ablation_frames} presents an ablation study on the effect of varying the number of frames in the input sequence. 
The results show steady improvements in both IoU and mIoU as the number of frames increases, with performance peaking at 4 frames before declining. 
We infer that additional preceding frames introduce noise rather than useful contextual information. 
Accordingly, we empirically select 4 frames as the input sequence. 

\subsection{Efficiency Comparisons}
In Table~\ref{tab:efficiency}, we compare the number of parameters and memory usage of our model with other open-source camera-based methods, particularly the baseline using temporal LSS fusion strategy.
Our model demonstrates a strong performance–efficiency trade-off among temporal-frame-based methods such as VoxFormer-T and HTCL-S, significantly outperforming HTCL-S despite using fewer parameters and less memory.
Moreover, compared to the temporal LSS fusion strategy, integrating C3DFusion yields substantially better performance with comparable parameter count and memory usage, highlighting the superiority of our temporal fusion approach in terms of efficiency.
\input{Sections/Table/ablation_efficiency}

%% file: Sections/Table/results_SemanticKITTI_test.tex
\newcommand{\colorboxlabel}[2]{\raisebox{0pt}[0pt][0pt]{\tikz{\node[fill=#1, minimum width=0.28cm, minimum height=0.28cm, inner sep=0pt] {};}}~#2}
\newcommand{\rotatebigtext}[1]{\rotatebox{90}{\fontsize{10}{10}\selectfont #1}}
\newcommand{\rotatetinytext}[1]{\rotatebox{90}{\fontsize{8}{8}\selectfont #1}}
\newcommand{\rotateemptytext}[1]{\rotatebox{90}{\fontsize{2}{2}\selectfont #1}}

\definecolor{custom_road}{RGB}{255, 0, 255}
\definecolor{custom_sidewalk}{RGB}{75, 0, 75}
\definecolor{custom_parking}{RGB}{255, 150, 255}
\definecolor{custom_other_ground}{RGB}{175, 0, 75}
\definecolor{custom_building}{RGB}{255, 200, 0}
\definecolor{custom_car}{RGB}{100, 150, 245}
\definecolor{custom_truck}{RGB}{80, 30, 180}
\definecolor{custom_bicycle}{RGB}{100, 230, 245}
\definecolor{custom_motorcycle}{RGB}{30, 60, 150}
\definecolor{custom_other_veh}{RGB}{0, 0, 255}
\definecolor{custom_vegetation}{RGB}{0, 175, 0}
\definecolor{custom_trunk}{RGB}{135, 60, 0}
\definecolor{custom_terrain}{RGB}{150, 240, 80}
\definecolor{custom_person}{RGB}{255, 30, 30}
\definecolor{custom_bicyclist}{RGB}{255, 40, 200}
\definecolor{custom_motorcyclist}{RGB}{150, 30, 90}
\definecolor{custom_fence}{RGB}{255, 120, 50}
\definecolor{custom_pole}{RGB}{255, 240, 150}
\definecolor{custom_traf_sign}{RGB}{255, 0, 0}

\begin{table*}[hbt!]
\centering
\resizebox{\textwidth}{!}{%
\begin{tabular}{lccl|clclclclclclclclclclclclclclclclclclcl|cc}
\toprule
 &
   &
  \multicolumn{1}{l}{} &
  \multicolumn{1}{l|}{} &
  \multicolumn{2}{l}{} &
  \multicolumn{2}{l}{} &
  \multicolumn{2}{l}{} &
  \multicolumn{2}{l}{} &
  \multicolumn{2}{l}{} &
  \multicolumn{2}{l}{} &
  \multicolumn{2}{l}{} &
  \multicolumn{2}{l}{} &
  \multicolumn{2}{l}{} &
  \multicolumn{2}{l}{} &
  \multicolumn{2}{l}{} &
  \multicolumn{2}{l}{} &
  \multicolumn{2}{l}{} &
  \multicolumn{2}{l}{} &
  \multicolumn{2}{l}{} &
  \multicolumn{2}{l}{} &
  \multicolumn{2}{l}{} &
  \multicolumn{2}{l}{} &
  \multicolumn{2}{l|}{} &
  \multicolumn{2}{l}{} \\
 &
   &
  \multicolumn{1}{l}{} &
  \multicolumn{1}{l|}{} &
  \multicolumn{2}{l}{\rotatebigtext{road} \newline \textcolor{white}{\rotatetinytext{1}} \newline \rotatetinytext{(15.30\%)}} &
  \multicolumn{2}{l}{\rotatebigtext{sidewalk} \newline \textcolor{white}{\rotatetinytext{1}} \newline \rotatetinytext{(11.13\%)}} &
  \multicolumn{2}{l}{\rotatebigtext{parking} \newline \textcolor{white}{\rotatetinytext{1}}\newline \rotatetinytext{(1.12\%)}} &
  \multicolumn{2}{l}{\rotatebigtext{other-grnd.} \newline \textcolor{white}{\rotatetinytext{1}}\newline \rotatetinytext{(0.56\%)}} &
  \multicolumn{2}{l}{\rotatebigtext{building} \newline \textcolor{white}{\rotatetinytext{1}}\newline \rotatetinytext{(14.1\%)}} &
  \multicolumn{2}{l}{\rotatebigtext{car} \newline \textcolor{white}{\rotatetinytext{1}}\newline \rotatetinytext{(3.92\%)}} &
  \multicolumn{2}{l}{\rotatebigtext{truck} \newline \textcolor{white}{\rotatetinytext{1}}\newline \rotatetinytext{(0.16\%)}} &
  \multicolumn{2}{l}{\rotatebigtext{bicycle} \newline \textcolor{white}{\rotatetinytext{1}}\newline \rotatetinytext{(0.03\%)}} &
  \multicolumn{2}{l}{\rotatebigtext{motorcycle} \newline \textcolor{white}{\rotatetinytext{1}}\newline \rotatetinytext{(0.03\%)}} &
  \multicolumn{2}{l}{\rotatebigtext{other-veh.} \newline \textcolor{white}{\rotatetinytext{1}}\newline \rotatetinytext{(0.20\%)}} &
  \multicolumn{2}{l}{\rotatebigtext{vegetation} \newline \textcolor{white}{\rotatetinytext{1}}\newline \rotatetinytext{(39.3\%)}} &
  \multicolumn{2}{l}{\rotatebigtext{trunk} \newline \textcolor{white}{\rotatetinytext{1}}\newline \rotatetinytext{(0.51\%)}} &
  \multicolumn{2}{l}{\rotatebigtext{terrain} \newline \textcolor{white}{\rotatetinytext{1}}\newline \rotatetinytext{(9.17\%)}} &
  \multicolumn{2}{l}{\rotatebigtext{person} \newline \textcolor{white}{\rotatetinytext{1}}\newline \rotatetinytext{(0.07\%)}} &
  \multicolumn{2}{l}{\rotatebigtext{bicyclist} \newline \textcolor{white}{\rotatetinytext{1}}\newline \rotatetinytext{(0.07\%)}} &
  \multicolumn{2}{l}{\rotatebigtext{motorcyclist} \newline \textcolor{white}{\rotatetinytext{1}}\newline \rotatetinytext{(0.05\%)}} &
  \multicolumn{2}{l}{\rotatebigtext{fence} \newline \textcolor{white}{\rotatetinytext{1}}\newline \rotatetinytext{(3.90\%)}} &
  \multicolumn{2}{l}{\rotatebigtext{pole} \newline \textcolor{white}{\rotatetinytext{1}}\newline \rotatetinytext{(0.29\%)}} &
  \multicolumn{2}{l|}{\rotatebigtext{traf.-sign} \newline \textcolor{white}{\rotatetinytext{1}}\newline \rotatetinytext{(0.08\%)}} &
  \multicolumn{2}{c}{\textbf{OOV (val.)}} \\
\textbf{Method} &
  \textbf{Input} &
  \textbf{IoU} &
  \textbf{mIoU} &
  \multicolumn{2}{c}{\colorboxlabel{custom_road}{}} &
  \multicolumn{2}{c}{\colorboxlabel{custom_sidewalk}{}} &
  \multicolumn{2}{c}{\colorboxlabel{custom_parking}{}} &
  \multicolumn{2}{c}{\colorboxlabel{custom_other_ground}{}} &
  \multicolumn{2}{c}{\colorboxlabel{custom_building}{}} &
  \multicolumn{2}{c}{\colorboxlabel{custom_car}{}} &
  \multicolumn{2}{c}{\colorboxlabel{custom_truck}{}} &
  \multicolumn{2}{c}{\colorboxlabel{custom_bicycle}{}} &
  \multicolumn{2}{c}{\colorboxlabel{custom_motorcycle}{}} &
  \multicolumn{2}{c}{\colorboxlabel{custom_other_veh}{}} &
  \multicolumn{2}{c}{\colorboxlabel{custom_vegetation}{}} &
  \multicolumn{2}{c}{\colorboxlabel{custom_trunk}{}} &
  \multicolumn{2}{c}{\colorboxlabel{custom_terrain}{}} &
  \multicolumn{2}{c}{\colorboxlabel{custom_person}{}} &
  \multicolumn{2}{c}{\colorboxlabel{custom_bicyclist}{}} &
  \multicolumn{2}{c}{\colorboxlabel{custom_motorcyclist}{}} &
  \multicolumn{2}{c}{\colorboxlabel{custom_fence}{}} &
  \multicolumn{2}{c}{\colorboxlabel{custom_pole}{}} &
  \multicolumn{2}{c|}{\colorboxlabel{custom_traf_sign}{}} &
  \textbf{IoU} &
  \textbf{mIoU} \\ \midrule
   \multicolumn{44}{c}{\textbf{\textit{Single-Frame-Based}}} \\ \midrule
\multicolumn{1}{l|}{MonoScene} &
  \multicolumn{1}{c|}{Mono} &
  34.16 &
  11.08 &
  \multicolumn{2}{c}{54.70} &
  \multicolumn{2}{c}{27.10} &
  \multicolumn{2}{c}{24.80} &
  \multicolumn{2}{c}{5.70} &
  \multicolumn{2}{c}{14.40} &
  \multicolumn{2}{c}{18.80} &
  \multicolumn{2}{c}{3.30} &
  \multicolumn{2}{c}{0.50} &
  \multicolumn{2}{c}{0.70} &
  \multicolumn{2}{c}{4.40} &
  \multicolumn{2}{c}{14.90} &
  \multicolumn{2}{c}{2.40} &
  \multicolumn{2}{c}{19.50} &
  \multicolumn{2}{c}{1.00} &
  \multicolumn{2}{c}{1.40} &
  \multicolumn{2}{c}{0.40} &
  \multicolumn{2}{c}{11.10} &
  \multicolumn{2}{c}{3.30} &
  \multicolumn{2}{c|}{2.10} &
  31.07 &
  7.02 \\
\multicolumn{1}{l|}{TPVFormer} &
  \multicolumn{1}{c|}{Mono} &
  34.25 &
  11.26 &
  \multicolumn{2}{c}{55.10} &
  \multicolumn{2}{c}{27.20} &
  \multicolumn{2}{c}{27.40} &
  \multicolumn{2}{c}{6.50} &
  \multicolumn{2}{c}{14.80} &
  \multicolumn{2}{c}{19.20} &
  \multicolumn{2}{c}{3.70} &
  \multicolumn{2}{c}{1.00} &
  \multicolumn{2}{c}{0.50} &
  \multicolumn{2}{c}{2.30} &
  \multicolumn{2}{c}{13.90} &
  \multicolumn{2}{c}{2.60} &
  \multicolumn{2}{c}{20.40} &
  \multicolumn{2}{c}{1.10} &
  \multicolumn{2}{c}{2.40} &
  \multicolumn{2}{c}{0.30} &
  \multicolumn{2}{c}{11.00} &
  \multicolumn{2}{c}{2.90} &
  \multicolumn{2}{c|}{1.50} &
  30.76 &
  7.87 \\
\multicolumn{1}{l|}{OccFormer} &
  \multicolumn{1}{c|}{Mono} &
  34.53 &
  12.32 &
  \multicolumn{2}{c}{55.90} &
  \multicolumn{2}{c}{30.30} &
  \multicolumn{2}{c}{31.50} &
  \multicolumn{2}{c}{6.50} &
  \multicolumn{2}{c}{15.70} &
  \multicolumn{2}{c}{21.60} &
  \multicolumn{2}{c}{1.20} &
  \multicolumn{2}{c}{1.50} &
  \multicolumn{2}{c}{1.70} &
  \multicolumn{2}{c}{3.20} &
  \multicolumn{2}{c}{16.80} &
  \multicolumn{2}{c}{3.90} &
  \multicolumn{2}{c}{21.30} &
  \multicolumn{2}{c}{2.20} &
  \multicolumn{2}{c}{1.10} &
  \multicolumn{2}{c}{0.20} &
  \multicolumn{2}{c}{11.90} &
  \multicolumn{2}{c}{3.80} &
  \multicolumn{2}{c|}{3.70} &
  30.46 &
  8.68 \\
\multicolumn{1}{l|}{Symphonies} &
  \multicolumn{1}{c|}{Stereo} &
  42.19 &
  15.04 &
  \multicolumn{2}{c}{58.40} &
  \multicolumn{2}{c}{29.30} &
  \multicolumn{2}{c}{26.90} &
  \multicolumn{2}{c}{11.70} &
  \multicolumn{2}{c}{24.70} &
  \multicolumn{2}{c}{23.60} &
  \multicolumn{2}{c}{3.20} &
  \multicolumn{2}{c}{3.60} &
  \multicolumn{2}{c}{2.60} &
  \multicolumn{2}{c}{5.60} &
  \multicolumn{2}{c}{24.20} &
  \multicolumn{2}{c}{10.00} &
  \multicolumn{2}{c}{23.10} &
  \multicolumn{2}{c}{\textbf{3.20}} &
  \multicolumn{2}{c}{1.90} &
  \multicolumn{2}{c}{\textbf{2.00}} &
  \multicolumn{2}{c}{16.10} &
  \multicolumn{2}{c}{7.70} &
  \multicolumn{2}{c|}{8.00} &
  23.48 &
  6.40 \\
\multicolumn{1}{l|}{CGFormer} &
  \multicolumn{1}{c|}{Stereo} &
  44.41 &
  16.63 &
  \multicolumn{2}{c}{64.30} &
  \multicolumn{2}{c}{34.20} &
  \multicolumn{2}{c}{34.10} &
  \multicolumn{2}{c}{12.10} &
  \multicolumn{2}{c}{25.80} &
  \multicolumn{2}{c}{26.10} &
  \multicolumn{2}{c}{4.30} &
  \multicolumn{2}{c}{3.70} &
  \multicolumn{2}{c}{1.30} &
  \multicolumn{2}{c}{2.70} &
  \multicolumn{2}{c}{24.50} &
  \multicolumn{2}{c}{11.20} &
  \multicolumn{2}{c}{29.30} &
  \multicolumn{2}{c}{1.70} &
  \multicolumn{2}{c}{3.60} &
  \multicolumn{2}{c}{0.40} &
  \multicolumn{2}{c}{18.70} &
  \multicolumn{2}{c}{8.70} &
  \multicolumn{2}{c|}{9.30} &
  33.54 &
  9.06 \\
\multicolumn{1}{l|}{L2COcc-C} &
  \multicolumn{1}{c|}{Stereo} &
  44.31 &
  17.03 &
  \multicolumn{2}{c}{66.00} &
  \multicolumn{2}{c}{35.00} &
  \multicolumn{2}{c}{33.10} &
  \multicolumn{2}{c}{13.50} &
  \multicolumn{2}{c}{25.10} &
  \multicolumn{2}{c}{27.20} &
  \multicolumn{2}{c}{3.00} &
  \multicolumn{2}{c}{3.50} &
  \multicolumn{2}{c}{\underline{3.60}} &
  \multicolumn{2}{c}{4.30} &
  \multicolumn{2}{c}{25.20} &
  \multicolumn{2}{c}{11.50} &
  \multicolumn{2}{c}{30.10} &
  \multicolumn{2}{c}{1.50} &
  \multicolumn{2}{c}{2.40} &
  \multicolumn{2}{c}{0.20} &
  \multicolumn{2}{c}{20.50} &
  \multicolumn{2}{c}{9.10} &
  \multicolumn{2}{c|}{8.90} &
  32.24 &
  8.55 \\
\multicolumn{1}{l|}{ScanSSC} &
  \multicolumn{1}{c|}{Stereo} &
  44.54 &
  17.40 &
  \multicolumn{2}{c}{66.20} &
  \multicolumn{2}{c}{35.90} &
  \multicolumn{2}{c}{\underline{35.10}} &
  \multicolumn{2}{c}{12.50} &
  \multicolumn{2}{c}{25.30} &
  \multicolumn{2}{c}{27.10} &
  \multicolumn{2}{c}{3.50} &
  \multicolumn{2}{c}{3.50} &
  \multicolumn{2}{c}{3.20} &
  \multicolumn{2}{c}{6.10} &
  \multicolumn{2}{c}{25.20} &
  \multicolumn{2}{c}{11.00} &
  \multicolumn{2}{c}{30.60} &
  \multicolumn{2}{c}{1.80} &
  \multicolumn{2}{c}{\textbf{5.30}} &
  \multicolumn{2}{c}{0.70} &
  \multicolumn{2}{c}{20.50} &
  \multicolumn{2}{c}{8.40} &
  \multicolumn{2}{c|}{8.90} &
  33.60 &
  9.50 \\
\multicolumn{1}{l|}{L2COcc-D} &
  \multicolumn{1}{c|}{Stereo} &
  45.37 &
  \underline{18.18} &
  \multicolumn{2}{c}{\textbf{68.20}} &
  \multicolumn{2}{c}{\textbf{36.90}} &
  \multicolumn{2}{c}{34.60} &
  \multicolumn{2}{c}{\underline{16.20}} &
  \multicolumn{2}{c}{25.80} &
  \multicolumn{2}{c}{\underline{28.30}} &
  \multicolumn{2}{c}{4.50} &
  \multicolumn{2}{c}{\underline{4.90}} &
  \multicolumn{2}{c}{3.30} &
  \multicolumn{2}{c}{\textbf{7.20}} &
  \multicolumn{2}{c}{26.20} &
  \multicolumn{2}{c}{\underline{11.90}} &
  \multicolumn{2}{c}{\underline{32.00}} &
  \multicolumn{2}{c}{2.10} &
  \multicolumn{2}{c}{2.40} &
  \multicolumn{2}{c}{0.30} &
  \multicolumn{2}{c}{\underline{21.60}} &
  \multicolumn{2}{c}{\underline{9.60}} &
  \multicolumn{2}{c|}{\underline{9.50}} &
  31.85 &
  10.05 \\ \midrule
  \multicolumn{44}{c}{\textbf{\textit{Temporal-Frame-Based}}} \\ \midrule
\multicolumn{1}{l|}{VoxFormer-T} &
  \multicolumn{1}{c|}{Stereo} &
  43.21 &
  13.41 &
  \multicolumn{2}{c}{54.10} &
  \multicolumn{2}{c}{26.90} &
  \multicolumn{2}{c}{25.10} &
  \multicolumn{2}{c}{7.30} &
  \multicolumn{2}{c}{23.50} &
  \multicolumn{2}{c}{21.70} &
  \multicolumn{2}{c}{3.60} &
  \multicolumn{2}{c}{1.90} &
  \multicolumn{2}{c}{1.60} &
  \multicolumn{2}{c}{4.10} &
  \multicolumn{2}{c}{24.40} &
  \multicolumn{2}{c}{8.10} &
  \multicolumn{2}{c}{24.20} &
  \multicolumn{2}{c}{1.60} &
  \multicolumn{2}{c}{1.10} &
  \multicolumn{2}{c}{0.00} &
  \multicolumn{2}{c}{13.10} &
  \multicolumn{2}{c}{6.60} &
  \multicolumn{2}{c|}{5.70} &
  \underline{40.21} &
  \underline{11.58} \\
\multicolumn{1}{l|}{HTCL-S} &
  \multicolumn{1}{c|}{Stereo} &
  44.23 &
  17.09 &
  \multicolumn{2}{c}{64.40} &
  \multicolumn{2}{c}{34.80} &
  \multicolumn{2}{c}{33.80} &
  \multicolumn{2}{c}{12.40} &
  \multicolumn{2}{c}{25.90} &
  \multicolumn{2}{c}{27.30} &
  \multicolumn{2}{c}{\textbf{10.80}} &
  \multicolumn{2}{c}{1.80} &
  \multicolumn{2}{c}{2.20} &
  \multicolumn{2}{c}{5.40} &
  \multicolumn{2}{c}{25.30} &
  \multicolumn{2}{c}{10.80} &
  \multicolumn{2}{c}{31.20} &
  \multicolumn{2}{c}{1.10} &
  \multicolumn{2}{c}{3.10} &
  \multicolumn{2}{c}{0.90} &
  \multicolumn{2}{c}{21.10} &
  \multicolumn{2}{c}{9.00} &
  \multicolumn{2}{c|}{8.30} &
  33.14 &
  9.04 \\
\multicolumn{1}{l|}{Hi-SOP} &
  \multicolumn{1}{c|}{Stereo} &
  44.57 &
  17.49 &
  \multicolumn{2}{c}{63.95} &
  \multicolumn{2}{c}{34.27} &
  \multicolumn{2}{c}{\textbf{35.85}} &
  \multicolumn{2}{c}{13.77} &
  \multicolumn{2}{c}{25.91} &
  \multicolumn{2}{c}{27.35} &
  \multicolumn{2}{c}{7.18} &
  \multicolumn{2}{c}{2.99} &
  \multicolumn{2}{c}{2.59} &
  \multicolumn{2}{c}{\underline{7.19}} &
  \multicolumn{2}{c}{26.07} &
  \multicolumn{2}{c}{10.35} &
  \multicolumn{2}{c}{30.77} &
  \multicolumn{2}{c}{1.68} &
  \multicolumn{2}{c}{4.81} &
  \multicolumn{2}{c}{1.06} &
  \multicolumn{2}{c}{20.15} &
  \multicolumn{2}{c}{8.70} &
  \multicolumn{2}{c|}{7.90} &
  
  - &
  - \\
\multicolumn{1}{l|}{FlowScene} &
  \multicolumn{1}{c|}{Stereo} &
  45.20 &
  17.70 &
  \multicolumn{2}{c}{64.10} &
  \multicolumn{2}{c}{35.00} &
  \multicolumn{2}{c}{33.70} &
  \multicolumn{2}{c}{13.00} &
  \multicolumn{2}{c}{27.70} &
  \multicolumn{2}{c}{26.40} &
  \multicolumn{2}{c}{\underline{10.00}} &
  \multicolumn{2}{c}{4.20} &
  \multicolumn{2}{c}{3.10} &
  \multicolumn{2}{c}{7.00} &
  \multicolumn{2}{c}{26.30} &
  \multicolumn{2}{c}{10.00} &
  \multicolumn{2}{c}{30.20} &
  \multicolumn{2}{c}{\underline{3.10}} &
  \multicolumn{2}{c}{\underline{5.10}} &
  \multicolumn{2}{c}{\underline{1.10}} &
  \multicolumn{2}{c}{20.20} &
  \multicolumn{2}{c}{8.90} &
  \multicolumn{2}{c|}{9.10} &
  - &
  - \\
\multicolumn{1}{l|}{CF-SSC} &
  \multicolumn{1}{c|}{Stereo} &
  \underline{46.21} &
  16.40 &
  \multicolumn{2}{c}{61.30} &
  \multicolumn{2}{c}{33.30} &
  \multicolumn{2}{c}{29.20} &
  \multicolumn{2}{c}{11.90} &
  \multicolumn{2}{c}{\underline{30.40}} &
  \multicolumn{2}{c}{26.30} &
  \multicolumn{2}{c}{4.80} &
  \multicolumn{2}{c}{2.60} &
  \multicolumn{2}{c}{2.70} &
  \multicolumn{2}{c}{6.30} &
  \multicolumn{2}{c}{\underline{28.50}} &
  \multicolumn{2}{c}{11.40} &
  \multicolumn{2}{c}{28.30} &
  \multicolumn{2}{c}{1.50} &
  \multicolumn{2}{c}{1.40} &
  \multicolumn{2}{c}{0.40} &
  \multicolumn{2}{c}{17.70} &
  \multicolumn{2}{c}{7.20} &
  \multicolumn{2}{c|}{6.30} &
  - &
  - \\
\rowcolor[HTML]{EFEFEF} 
\multicolumn{1}{l|}{\cellcolor[HTML]{EFEFEF}\textbf{Ours}} &
  \multicolumn{1}{c|}{\cellcolor[HTML]{EFEFEF}\textbf{Stereo}} &
  \textbf{47.62} &
  \textbf{18.98} &
  \multicolumn{2}{c}{\cellcolor[HTML]{EFEFEF}\underline{67.00}} &
  \multicolumn{2}{c}{\cellcolor[HTML]{EFEFEF}\underline{36.30}} &
  \multicolumn{2}{c}{\cellcolor[HTML]{EFEFEF}33.20} &
  \multicolumn{2}{c}{\cellcolor[HTML]{EFEFEF}\textbf{19.30}} &
  \multicolumn{2}{c}{\cellcolor[HTML]{EFEFEF}\textbf{30.60}} &
  \multicolumn{2}{c}{\cellcolor[HTML]{EFEFEF}\textbf{29.00}} &
  \multicolumn{2}{c}{\cellcolor[HTML]{EFEFEF}3.30} &
  \multicolumn{2}{c}{\cellcolor[HTML]{EFEFEF}\textbf{5.40}} &
  \multicolumn{2}{c}{\cellcolor[HTML]{EFEFEF}\textbf{4.40}} &
  \multicolumn{2}{c}{\cellcolor[HTML]{EFEFEF}4.70} &
  \multicolumn{2}{c}{\cellcolor[HTML]{EFEFEF}\textbf{29.60}} &
  \multicolumn{2}{c}{\cellcolor[HTML]{EFEFEF}\textbf{14.70}} &
  \multicolumn{2}{c}{\cellcolor[HTML]{EFEFEF}\textbf{33.80}} &
  \multicolumn{2}{c}{\cellcolor[HTML]{EFEFEF}1.60} &
  \multicolumn{2}{c}{\cellcolor[HTML]{EFEFEF}2.80} &
  \multicolumn{2}{c}{\cellcolor[HTML]{EFEFEF}0.30} &
  \multicolumn{2}{c}{\cellcolor[HTML]{EFEFEF}\textbf{22.80}} &
  \multicolumn{2}{c}{\cellcolor[HTML]{EFEFEF}\textbf{11.40}} &
  \multicolumn{2}{c|}{\cellcolor[HTML]{EFEFEF}\textbf{10.40}} &
  \textbf{44.37} &
  \textbf{17.17} \\ \bottomrule

\end{tabular}%
}
\caption{Quantitative results on SemanticKITTI hidden test set. ‘OOV’ on the right indicates performance on out-of-view regions, evaluated on the validation set using our implementation.
\textbf{Bold} / \underline{underline} highlight the best / second-best, respectively.}
\label{tab:SemKITTI_test}
    
\end{table*}

%% file: Sections/Table/results_KITTI360.tex

\definecolor{custom_road}{RGB}{255, 0, 255}
\definecolor{custom_sidewalk}{RGB}{75, 0, 75}
\definecolor{custom_parking}{RGB}{255, 150, 255}
\definecolor{custom_other_ground}{RGB}{175, 0, 75}
\definecolor{custom_building}{RGB}{255, 200, 0}
\definecolor{custom_car}{RGB}{100, 150, 245}
\definecolor{custom_truck}{RGB}{80, 30, 180}
\definecolor{custom_bicycle}{RGB}{100, 230, 245}
\definecolor{custom_motorcycle}{RGB}{30, 60, 150}
\definecolor{custom_other_veh}{RGB}{0, 0, 255}
\definecolor{custom_vegetation}{RGB}{0, 175, 0}
\definecolor{custom_trunk}{RGB}{135, 60, 0}
\definecolor{custom_terrain}{RGB}{150, 240, 80}
\definecolor{custom_person}{RGB}{255, 30, 30}
\definecolor{custom_bicyclist}{RGB}{255, 40, 20}
\definecolor{custom_motorcyclist}{RGB}{150, 30, 90}
\definecolor{custom_fence}{RGB}{255, 120, 50}
\definecolor{custom_pole}{RGB}{255, 240, 150}
\definecolor{custom_traf_sign}{RGB}{255, 0, 0}
\definecolor{custom_other_struct}{RGB}{255, 150, 0}
\definecolor{custom_other_obj}{RGB}{50, 255, 255}

\begin{table*}[hbt!]
\centering
\resizebox{\textwidth}{!}{%
\begin{tabular}{lccc|cccccccccccccccccccccccccccccccccccc|cc}
\toprule
 &
   &
  \multicolumn{1}{l}{} &
   &
  \multicolumn{2}{c}{} &
  \multicolumn{2}{c}{} &
  \multicolumn{2}{c}{} &
  \multicolumn{2}{c}{} &
  \multicolumn{2}{c}{} &
  \multicolumn{2}{c}{} &
  \multicolumn{2}{c}{} &
  \multicolumn{2}{c}{} &
  \multicolumn{2}{c}{} &
  \multicolumn{2}{c}{} &
  \multicolumn{2}{c}{} &
  \multicolumn{2}{c}{} &
  \multicolumn{2}{c}{} &
  \multicolumn{2}{c}{} &
  \multicolumn{2}{c}{} &
  \multicolumn{2}{c}{} &
  \multicolumn{2}{c}{} &
  \multicolumn{2}{c|}{} &
  \multicolumn{2}{c}{} \\
 &
   &
   &
   &
  \multicolumn{2}{c}{\rotatebigtext{car} \newline \textcolor{white}{\rotatetinytext{1}} \newline \rotatetinytext{(2.85\%)}} &
  \multicolumn{2}{c}{\rotatebigtext{bicycle} \newline \textcolor{white}{\rotatetinytext{1}} \newline \rotatetinytext{(0.01\%)}} &
  \multicolumn{2}{c}{\rotatebigtext{motorcycle} \newline \textcolor{white}{\rotatetinytext{1}}\newline \rotatetinytext{(0.01\%)}} &
  \multicolumn{2}{c}{\rotatebigtext{truck} \newline \textcolor{white}{\rotatetinytext{1}}\newline \rotatetinytext{(0.16\%)}} &
  \multicolumn{2}{c}{\rotatebigtext{other-veh.} \newline \textcolor{white}{\rotatetinytext{1}}\newline \rotatetinytext{(5.75\%)}} &
  \multicolumn{2}{c}{\rotatebigtext{person} \newline \textcolor{white}{\rotatetinytext{1}}\newline \rotatetinytext{(0.02\%)}} &
  \multicolumn{2}{c}{\rotatebigtext{road} \newline \textcolor{white}{\rotatetinytext{1}}\newline \rotatetinytext{(14.98\%)}} &
  \multicolumn{2}{c}{\rotatebigtext{parking} \newline \textcolor{white}{\rotatetinytext{1}}\newline \rotatetinytext{(2.31\%)}} &
  \multicolumn{2}{c}{\rotatebigtext{sidewalk} \newline \textcolor{white}{\rotatetinytext{1}}\newline \rotatetinytext{(6.43\%)}} &
  \multicolumn{2}{c}{\rotatebigtext{other-grnd.} \newline \textcolor{white}{\rotatetinytext{1}}\newline \rotatetinytext{(2.05\%)}} &
  \multicolumn{2}{c}{\rotatebigtext{building} \newline \textcolor{white}{\rotatetinytext{1}}\newline \rotatetinytext{(15.67\%)}} &
  \multicolumn{2}{c}{\rotatebigtext{fence} \newline \textcolor{white}{\rotatetinytext{1}}\newline \rotatetinytext{(0.96\%)}} &
  \multicolumn{2}{c}{\rotatebigtext{vegetation} \newline \textcolor{white}{\rotatetinytext{1}}\newline \rotatetinytext{(41.99\%)}} &
  \multicolumn{2}{c}{\rotatebigtext{terrain} \newline \textcolor{white}{\rotatetinytext{1}}\newline \rotatetinytext{(7.10\%)}} &
  \multicolumn{2}{c}{\rotatebigtext{pole} \newline \textcolor{white}{\rotatetinytext{1}}\newline \rotatetinytext{(0.22\%)}} &
  \multicolumn{2}{c}{\rotatebigtext{traf.-sign} \newline \textcolor{white}{\rotatetinytext{1}}\newline \rotatetinytext{(0.06\%)}} &
  \multicolumn{2}{c}{\rotatebigtext{other-struct.} \newline \textcolor{white}{\rotatetinytext{1}}\newline \rotatetinytext{(4.33\%)}} &
  \multicolumn{2}{c|}{\rotatebigtext{other-obj.} \newline \textcolor{white}{\rotatetinytext{1}}\newline \rotatetinytext{(0.28\%)}} &
  \multicolumn{2}{c}{\textbf{OOV (test)}}
   \\
  \textbf{Method} &
  \textbf{Input} &
  \textbf{IoU} &
  \textbf{mIoU} &
  \multicolumn{2}{c}{\colorboxlabel{custom_car}{}} &
  \multicolumn{2}{c}{\colorboxlabel{custom_bicycle}{}} &
  \multicolumn{2}{c}{\colorboxlabel{custom_motorcycle}{}} &
  \multicolumn{2}{c}{\colorboxlabel{custom_truck}{}} &
  \multicolumn{2}{c}{\colorboxlabel{custom_other_veh}{}} &
  \multicolumn{2}{c}{\colorboxlabel{custom_person}{}} &
  \multicolumn{2}{c}{\colorboxlabel{custom_road}{}} &
  \multicolumn{2}{c}{\colorboxlabel{custom_parking}{}} &
  \multicolumn{2}{c}{\colorboxlabel{custom_sidewalk}{}} &
  \multicolumn{2}{c}{\colorboxlabel{custom_other_ground}{}} &
  \multicolumn{2}{c}{\colorboxlabel{custom_building}{}} &
  \multicolumn{2}{c}{\colorboxlabel{custom_fence}{}} &
  \multicolumn{2}{c}{\colorboxlabel{custom_vegetation}{}} &
  \multicolumn{2}{c}{\colorboxlabel{custom_terrain}{}} &
  \multicolumn{2}{c}{\colorboxlabel{custom_pole}{}} &
  \multicolumn{2}{c}{\colorboxlabel{custom_traf_sign}{}} &
  \multicolumn{2}{c}{\colorboxlabel{custom_other_struct}{}} &
  \multicolumn{2}{c|}{\colorboxlabel{custom_other_obj}{}} &
  \textbf{IoU} &
  \textbf{mIoU} \\ \midrule
    \multicolumn{42}{c}{\textbf{\textit{Single-Frame-Based}}} \\ \midrule
\multicolumn{1}{l|}{MonoScene} &
  \multicolumn{1}{c|}{Mono} &
  37.87 &
  12.31 &
  \multicolumn{2}{c}{19.34} &
  \multicolumn{2}{c}{0.43} &
  \multicolumn{2}{c}{0.58} &
  \multicolumn{2}{c}{8.02} &
  \multicolumn{2}{c}{2.03} &
  \multicolumn{2}{c}{0.86} &
  \multicolumn{2}{c}{48.35} &
  \multicolumn{2}{c}{11.38} &
  \multicolumn{2}{c}{28.13} &
  \multicolumn{2}{c}{3.32} &
  \multicolumn{2}{c}{32.89} &
  \multicolumn{2}{c}{3.53} &
  \multicolumn{2}{c}{26.15} &
  \multicolumn{2}{c}{16.75} &
  \multicolumn{2}{c}{6.92} &
  \multicolumn{2}{c}{5.67} &
  \multicolumn{2}{c}{4.20} &
  \multicolumn{2}{c|}{3.09} &
   - &
   - \\
\multicolumn{1}{l|}{TPVFormer} &
  \multicolumn{1}{c|}{Mono} &
  40.22 &
  13.64 &
  \multicolumn{2}{c}{21.56} &
  \multicolumn{2}{c}{1.09} &
  \multicolumn{2}{c}{1.37} &
  \multicolumn{2}{c}{8.06} &
  \multicolumn{2}{c}{2.57} &
  \multicolumn{2}{c}{2.38} &
  \multicolumn{2}{c}{52.99} &
  \multicolumn{2}{c}{11.99} &
  \multicolumn{2}{c}{31.07} &
  \multicolumn{2}{c}{3.78} &
  \multicolumn{2}{c}{34.83} &
  \multicolumn{2}{c}{4.80} &
  \multicolumn{2}{c}{30.08} &
  \multicolumn{2}{c}{17.52} &
  \multicolumn{2}{c}{7.46} &
  \multicolumn{2}{c}{5.86} &
  \multicolumn{2}{c}{5.48} &
  \multicolumn{2}{c|}{2.70} &
   - &
   - \\
\multicolumn{1}{l|}{OccFormer} &
  \multicolumn{1}{c|}{Mono} &
  40.27 &
  13.81 &
  \multicolumn{2}{c}{22.58} &
  \multicolumn{2}{c}{0.66} &
  \multicolumn{2}{c}{0.26} &
  \multicolumn{2}{c}{9.89} &
  \multicolumn{2}{c}{3.82} &
  \multicolumn{2}{c}{2.77} &
  \multicolumn{2}{c}{54.30} &
  \multicolumn{2}{c}{13.44} &
  \multicolumn{2}{c}{31.53} &
  \multicolumn{2}{c}{3.55} &
  \multicolumn{2}{c}{36.42} &
  \multicolumn{2}{c}{4.80} &
  \multicolumn{2}{c}{31.00} &
  \multicolumn{2}{c}{19.51} &
  \multicolumn{2}{c}{7.77} &
  \multicolumn{2}{c}{8.51} &
  \multicolumn{2}{c}{6.95} &
  \multicolumn{2}{c|}{4.60} &
   - &
   - \\
\multicolumn{1}{l|}{Symphonies} &
  \multicolumn{1}{c|}{Stereo} &
  44.12 &
  18.58 &
  \multicolumn{2}{c}{\underline{30.02}} &
  \multicolumn{2}{c}{1.85} &
  \multicolumn{2}{c}{5.90} &
  \multicolumn{2}{c}{\textbf{25.07}} &
  \multicolumn{2}{c}{\textbf{12.06}} &
  \multicolumn{2}{c}{\textbf{8.20}} &
  \multicolumn{2}{c}{54.94} &
  \multicolumn{2}{c}{13.83} &
  \multicolumn{2}{c}{32.76} &
  \multicolumn{2}{c}{\textbf{6.93}} &
  \multicolumn{2}{c}{35.11} &
  \multicolumn{2}{c}{8.58} &
  \multicolumn{2}{c}{38.33} &
  \multicolumn{2}{c}{11.52} &
  \multicolumn{2}{c}{14.01} &
  \multicolumn{2}{c}{9.57} &
  \multicolumn{2}{c}{\textbf{14.44}} &
  \multicolumn{2}{c|}{\textbf{11.28}} &
   34.39 &
   11.93 \\
\multicolumn{1}{l|}{CGFormer} &
  \multicolumn{1}{c|}{Stereo} &
  48.07 &
  20.05 &
  \multicolumn{2}{c}{29.85} &
  \multicolumn{2}{c}{3.42} &
  \multicolumn{2}{c}{3.96} &
  \multicolumn{2}{c}{17.59} &
  \multicolumn{2}{c}{6.79} &
  \multicolumn{2}{c}{6.63} &
  \multicolumn{2}{c}{\textbf{63.85}} &
  \multicolumn{2}{c}{17.15} &
  \multicolumn{2}{c}{\underline{40.72}} &
  \multicolumn{2}{c}{5.53} &
  \multicolumn{2}{c}{42.73} &
  \multicolumn{2}{c}{8.22} &
  \multicolumn{2}{c}{38.80} &
  \multicolumn{2}{c}{24.94} &
  \multicolumn{2}{c}{16.24} &
  \multicolumn{2}{c}{17.45} &
  \multicolumn{2}{c}{10.18} &
  \multicolumn{2}{c|}{6.77} &
   44.72 &
   15.61 \\
\multicolumn{1}{l|}{ScanSSC} &
  \multicolumn{1}{c|}{Stereo} &
  \underline{48.29} &
  \underline{20.14} &
  \multicolumn{2}{c}{29.91} &
  \multicolumn{2}{c}{3.78} &
  \multicolumn{2}{c}{4.28} &
  \multicolumn{2}{c}{14.34} &
  \multicolumn{2}{c}{\underline{9.08}} &
  \multicolumn{2}{c}{6.65} &
  \multicolumn{2}{c}{62.21} &
  \multicolumn{2}{c}{\underline{18.16}} &
  \multicolumn{2}{c}{40.19} &
  \multicolumn{2}{c}{5.16} &
  \multicolumn{2}{c}{42.68} &
  \multicolumn{2}{c}{8.83} &
  \multicolumn{2}{c}{\underline{38.84}} &
  \multicolumn{2}{c}{\underline{25.50}} &
  \multicolumn{2}{c}{16.60} &
  \multicolumn{2}{c}{\underline{19.14}} &
  \multicolumn{2}{c}{10.30} &
  \multicolumn{2}{c|}{6.89} &
   45.09 &
   15.44 \\ \midrule
    \multicolumn{42}{c}{\textbf{\textit{Temporal-Frame-Based}}} \\ \midrule
\multicolumn{1}{l|}{FlowScene} &
  \multicolumn{1}{c|}{Stereo} &
  46.98 &
  19.12 &
  \multicolumn{2}{c}{29.83} &
  \multicolumn{2}{c}{\underline{4.44}} &
  \multicolumn{2}{c}{3.78} &
  \multicolumn{2}{c}{16.71} &
  \multicolumn{2}{c}{8.71} &
  \multicolumn{2}{c}{\underline{7.77}} &
  \multicolumn{2}{c}{60.70} &
  \multicolumn{2}{c}{16.99} &
  \multicolumn{2}{c}{39.59} &
  \multicolumn{2}{c}{\underline{6.01}} &
  \multicolumn{2}{c}{\underline{43.17}} &
  \multicolumn{2}{c}{\underline{9.45}} &
  \multicolumn{2}{c}{37.32} &
  \multicolumn{2}{c}{25.14} &
  \multicolumn{2}{c}{\underline{17.35}} &
  \multicolumn{2}{c}{18.12} &
  \multicolumn{2}{c}{10.63} &
  \multicolumn{2}{c|}{7.56} &
   - &
   - \\
\multicolumn{1}{l|}{CF-SSC} &
  \multicolumn{1}{c|}{Stereo} &
  45.79 &
  19.10 &
  \multicolumn{2}{c}{28.10} &
  \multicolumn{2}{c}{3.39} &
  \multicolumn{2}{c}{\underline{6.87}} &
  \multicolumn{2}{c}{16.76} &
  \multicolumn{2}{c}{7.75} &
  \multicolumn{2}{c}{5.68} &
  \multicolumn{2}{c}{59.01} &
  \multicolumn{2}{c}{16.80} &
  \multicolumn{2}{c}{37.60} &
  \multicolumn{2}{c}{4.95} &
  \multicolumn{2}{c}{42.16} &
  \multicolumn{2}{c}{8.26} &
  \multicolumn{2}{c}{36.14} &
  \multicolumn{2}{c}{21.89} &
  \multicolumn{2}{c}{14.73} &
  \multicolumn{2}{c}{17.72} &
  \multicolumn{2}{c}{9.73} &
  \multicolumn{2}{c|}{7.14} &
   - &
   - \\
\rowcolor[HTML]{EFEFEF} 
\multicolumn{1}{l|}{\cellcolor[HTML]{EFEFEF}\textbf{Ours}} &
  \multicolumn{1}{c|}{\cellcolor[HTML]{EFEFEF}\textbf{Stereo}} &
  \textbf{49.28} &
   \textbf{21.74}&
  \multicolumn{2}{c}{\cellcolor[HTML]{EFEFEF} \textbf{31.16}} & 
  \multicolumn{2}{c}{\cellcolor[HTML]{EFEFEF} \textbf{5.39}} &
  \multicolumn{2}{c}{\cellcolor[HTML]{EFEFEF} \textbf{7.01}} &
  \multicolumn{2}{c}{\cellcolor[HTML]{EFEFEF} \underline{18.12}} &
  \multicolumn{2}{c}{\cellcolor[HTML]{EFEFEF} 8.25} &
  \multicolumn{2}{c}{\cellcolor[HTML]{EFEFEF} 5.66} &
  \multicolumn{2}{c}{\cellcolor[HTML]{EFEFEF} \underline{63.70}} &
  \multicolumn{2}{c}{\cellcolor[HTML]{EFEFEF} \textbf{19.12}} &
  \multicolumn{2}{c}{\cellcolor[HTML]{EFEFEF} \textbf{41.64}} &
  \multicolumn{2}{c}{\cellcolor[HTML]{EFEFEF} 5.09} &
  \multicolumn{2}{c}{\cellcolor[HTML]{EFEFEF} \textbf{43.93}} &
  \multicolumn{2}{c}{\cellcolor[HTML]{EFEFEF} \textbf{10.43}} &
  \multicolumn{2}{c}{\cellcolor[HTML]{EFEFEF} \textbf{40.73}} &
  \multicolumn{2}{c}{\cellcolor[HTML]{EFEFEF} \textbf{27.62}} &
  \multicolumn{2}{c}{\cellcolor[HTML]{EFEFEF} \textbf{19.30}} &
  \multicolumn{2}{c}{\cellcolor[HTML]{EFEFEF} \textbf{23.08}} &
  \multicolumn{2}{c}{\cellcolor[HTML]{EFEFEF} \underline{12.34}} &
  \multicolumn{2}{c|}{\cellcolor[HTML]{EFEFEF} \underline{8.74}} &
   \textbf{52.41} &
   \textbf{23.72} \\ \bottomrule
\end{tabular}
}

\caption{Quantitative results on SSCBench-KITTI-360 test set. ‘OOV’ on the right indicates performance on out-of-view regions, evaluated on the test set using our implementation.
\textbf{Bold} / \underline{underline} highlight the best / second-best, respectively.}
\label{tab:KITTI360_test}
    
\end{table*}

%% file: Sections/Table/ablation_architecture.tex
\begin{table}[hbt!]
\centering
\setlength\tabcolsep{10pt}\resizebox{\linewidth}{!}{
\begin{tabular}{c|c c c|c c}
\toprule
                                                           & \textbf{TPFA}                      & \textbf{HCB}                       & \textbf{CCFD}                      & \multicolumn{1}{c}{\textbf{IoU}} & \multicolumn{1}{c}{\textbf{mIoU}} \\ \midrule
\multicolumn{1}{c|}{Baseline}                              &                                    &                                    &                                    & \multicolumn{1}{c}{48.59}        & \multicolumn{1}{c}{16.58}         \\
\multicolumn{1}{c|}{(a)}                                   & \checkmark          &                                    &                                    & \multicolumn{1}{l}{49.09}                                  & 18.45                             \\
\multicolumn{1}{c|}{(b)}                                   & \checkmark          & \checkmark          &                                    & 48.99                                                      & \multicolumn{1}{c}{18.88}         \\
\multicolumn{1}{c|}{(c)}                                   & \checkmark          &                                    & \checkmark          & \multicolumn{1}{l}{48.87}                                  & 18.86                             \\
\rowcolor[HTML]{EFEFEF} 
\multicolumn{1}{c|}{\cellcolor[HTML]{EFEFEF}\textbf{Ours}} & \textbf{\checkmark} & \textbf{\checkmark} & \textbf{\checkmark} & \multicolumn{1}{l}{\cellcolor[HTML]{EFEFEF}\textbf{49.53}} & \textbf{19.31}                    \\ \bottomrule
\end{tabular}}
\caption{Ablation study of C3DFusion. `TPFA', `HCB', and `CCFD' denote temporal 3D point feature alignment, historical context blurring, and current-centric feature densification, respectively.}
\label{tab:ablation_architectur}
\end{table}

%% file: Sections/Table/ablation_generalization.tex
\begin{table}[hbt!]
\centering
\setlength\tabcolsep{15pt}\resizebox{\linewidth}{!}{
\begin{tabular}{@{}l|c c@{}}
\toprule
\multicolumn{1}{c|}{\textbf{ Method}}     & \textbf{IoU}  & \textbf{mIoU} \\ \midrule
\; VoxFormer-S                                      & 44.02         & 12.35         \\
\multicolumn{1}{>{\hspace{1em}}l|}{+ \textit{Temporal LSS fusion}} & 44.39 (+0.37) & 9.56 (-2.79) \\
\rowcolor[HTML]{EFEFEF}
\multicolumn{1}{>{\hspace{1em}}l|}{+ \textbf{\textit{C3DFusion}}}         & \textbf{45.98 (+1.96)} & \textbf{15.12 (+2.77)} \\ \midrule
\; OccFormer                                        & 36.50         & 13.46         \\
\multicolumn{1}{>{\hspace{1em}}l|}{+ \textit{Temporal LSS fusion}} & 44.48 (+7.98) & 16.91 (+3.45) \\
\rowcolor[HTML]{EFEFEF}
\multicolumn{1}{>{\hspace{1em}}l|}{+ \textbf{\textit{C3DFusion}}}         & \textbf{44.83 (+8.36)} & \textbf{18.29 (+4.83)} \\ \midrule
\; ScanSSC                                          & 45.95         & 17.12         \\
\multicolumn{1}{>{\hspace{1em}}l|}{+ \textit{Temporal LSS fusion}} & 49.31 (+3.36) & 17.67 (+0.55) \\
\rowcolor[HTML]{EFEFEF}
\multicolumn{1}{>{\hspace{1em}}l|}{+ \textbf{\textit{C3DFusion}}}         & \textbf{49.89 (+3.94)} & \textbf{18.73 (+1.61)} \\ \midrule
\; CGFormer                                             & 45.99         & 16.87         \\
\multicolumn{1}{>{\hspace{1em}}l|}{+ \textit{Temporal LSS fusion}} & 48.59 (+2.60) & 16.58 (-0.29) \\
\rowcolor[HTML]{EFEFEF}
\multicolumn{1}{>{\hspace{1em}}l|}{+ \textbf{\textit{C3DFusion (Ours)}}}         & \textbf{49.53 (+3.54)} & \textbf{19.31 (+2.44)} \\
\bottomrule
\end{tabular}}
\caption{Generalization of C3DFusion across other leading camera-based SSC models.}
\label{tab:ablation_generalization}
\end{table}

%% file: Sections/Table/ablation_interpolation.tex
\begin{table}[hbt!]
\centering
\setlength\tabcolsep{8pt}\resizebox{.7\linewidth}{!}{
\begin{tabular}{@{}cll|clcl@{}}
\toprule
\multicolumn{3}{c|}{\textbf{Interpolation Factor}}               & \multicolumn{2}{c}{\textbf{IoU}}                           & \multicolumn{2}{c}{\textbf{mIoU}}                          \\ \midrule
\multicolumn{3}{c|}{$\times$ 1}                                  & \multicolumn{2}{c}{48.99}                                  & \multicolumn{2}{c}{18.88}                                  \\
\multicolumn{3}{c|}{\cellcolor[HTML]{EFEFEF}\textbf{$\times$ 2}} & \multicolumn{2}{c}{\cellcolor[HTML]{EFEFEF}\textbf{49.53}} & \multicolumn{2}{c}{\cellcolor[HTML]{EFEFEF}\textbf{19.31}} \\
\multicolumn{3}{c|}{$\times$ 3}                                  & \multicolumn{2}{c}{48.53}                                  & \multicolumn{2}{c}{19.08}                                  \\
\multicolumn{3}{c|}{$\times$ 4}                                  & \multicolumn{2}{c}{49.21}                                  & \multicolumn{2}{c}{18.86}                                  \\ \bottomrule
\end{tabular}}
\caption{Ablation study on the intensity of current-centric feature densification, controlled by the interpolation factor.}
\label{tab:ablation_interpolation}
\end{table}

%% file: Sections/Table/ablation_depth_gate.tex
\begin{table}[hbt!]
\centering
\setlength\tabcolsep{15pt}\resizebox{.8\linewidth}{!}{
\begin{tabular}{ cc|ll@{}}
\toprule
\textbf{History}          & \textbf{Current}          & \multicolumn{1}{c}{\textbf{IoU}}                  & \multicolumn{1}{c}{\textbf{mIoU}}                 \\ \midrule
\rowcolor[HTML]{EFEFEF} 
\checkmark &                           & \multicolumn{1}{c}{\cellcolor[HTML]{EFEFEF}\textbf{49.53}} & \multicolumn{1}{c}{\cellcolor[HTML]{EFEFEF}\textbf{19.31}} \\
                          & \checkmark & 48.74                                             & 18.29                                             \\
\checkmark & \checkmark & 49.22                                             & 18.76                                             \\ \bottomrule
\end{tabular}}
\caption{Ablation study on extending historical context blurring to the current frame.}
\label{tab:ablation_depth_gate}
\end{table}



%% file: Sections/Table/ablation_N_of_frames.tex
\begin{table}[hbt!]
\centering
\setlength\tabcolsep{10pt}\resizebox{.6\linewidth}{!}{
\begin{tabular}{@{}cll|clcl@{}}

\toprule
\multicolumn{3}{c|}{\textbf{\# of Frames $n$}}                 & \multicolumn{2}{c}{\textbf{IoU}}                           & \multicolumn{2}{c}{\textbf{mIoU}}                          \\ \midrule
\multicolumn{3}{c|}{1}                                  & \multicolumn{2}{c}{45.99}                                  & \multicolumn{2}{c}{16.37}                                  \\
\multicolumn{3}{c|}{2}                                  & \multicolumn{2}{c}{47.72}                                  & \multicolumn{2}{c}{18.00}                                  \\
\multicolumn{3}{c|}{3}                                  & \multicolumn{2}{c}{48.90}                                  & \multicolumn{2}{c}{18.62}                                  \\
\multicolumn{3}{c|}{\cellcolor[HTML]{EFEFEF}\textbf{4}} & \multicolumn{2}{c}{\cellcolor[HTML]{EFEFEF}\textbf{49.53}} & \multicolumn{2}{c}{\cellcolor[HTML]{EFEFEF}\textbf{19.31}} \\
\multicolumn{3}{c|}{5}                                  & \multicolumn{2}{c}{49.20}                                  & \multicolumn{2}{c}{18.58}                                  \\
\bottomrule
\end{tabular}}
\caption{Ablation study on the number of input frames $n$.}
\label{tab:ablation_frames}
\end{table}

%% file: Sections/Table/ablation_efficiency.tex
\begin{table}[hbt!]
\centering
\setlength\tabcolsep{8pt}\resizebox{\linewidth}{!}{
\begin{tabular}{@{}cll|clcl@{}}
\toprule
\multicolumn{2}{c|}{\textbf{Method}}  & \multicolumn{1}{c}{\textbf{\# of Params.}}                & \multicolumn{1}{c|}{\textbf{Memory}}       & \multicolumn{1}{c}{\textbf{mIoU}}   
\\ \midrule
\multicolumn{2}{l|}{VoxFormer-T}   &\multicolumn{1}{c}{58 M}& \multicolumn{1}{c|}{14763 MB} & \multicolumn{1}{c}{13.35} \\
\multicolumn{2}{l|}{HTCL-S}          & \multicolumn{1}{c}{182 M} & \multicolumn{1}{c|}{34593 MB}& \multicolumn{1}{c}{17.13}  \\
\midrule
\multicolumn{2}{l|}{CGFormer}   & \multicolumn{1}{c}{163 M} & \multicolumn{1}{c|}{16013 MB} & \multicolumn{1}{c}{16.87} \\
\multicolumn{2}{>{\hspace{1em}}l|}{ +\textit{ Temporal LSS fusion}}   &\multicolumn{1}{c}{163 M}& \multicolumn{1}{c|}{22687 MB} & \multicolumn{1}{c}{16.58} \\
\rowcolor[HTML]{EFEFEF} 
\multicolumn{2}{>{\hspace{1em}}l|}{+ \textit{\textbf{C3DFusion (Ours)}}}          & \multicolumn{1}{c}{160 M} & \multicolumn{1}{c|}{23987 MB} & \multicolumn{1}{c}{19.31} \\ \bottomrule
\end{tabular}}
\caption{Efficiency comparison with existing methods.}
\label{tab:efficiency}
\end{table}

%% file: Sections/5_Conclusion.tex
\section{Conclusion}
In this work, we address the challenge of completing regions beyond the current camera’s field of view by leveraging temporal context in camera-based SSC—a critical capability for autonomous driving applications.
To this end, we propose C3DFusion, a temporal geometry fusion approach that aligns 3D point features mapped directly from 2D image features via backprojection.
We further introduce techniques such as historical context blurring and current-centric feature densification to mitigate noise from warped historical points and strengthen the contribution of current-frame features in the aggregated point cloud.
Extensive experiments demonstrate that C3DFusion not only achieves SOTA performance on standard benchmarks, but also consistently improves performance across a variety of architectures. 
Owing to its simplicity, generalizability, and strong effectiveness, we believe C3DFusion offers valuable insights for camera-based SSC and broader 3D perception tasks, and can serve as a solid foundation for future research in temporal fusion.

%% file: Sections/99_Appendix.tex
\setcounter{table}{0}
\setcounter{figure}{0}
\renewcommand{\thetable}{A.\arabic{table}}
\renewcommand{\thefigure}{A.\arabic{figure}}
\renewcommand{\thealgorithm}{A.\arabic{algorithm}}
\setcounter{page}{1}

\twocolumn[
\begin{center}
    {\LARGE \textbf{Towards Temporal Fusion Beyond the Field of View
\\for Camera-based Semantic Scene Completion}}\\
    \vspace{20pt}
    {\Large\textbf{Supplementary Material}}
    \vspace{50pt}
\end{center}
]

\section{Implementation Details}

\subsection{Training Setup}
We train our model for 25 epochs on 4 NVIDIA A6000 GPUs with a batch size of 4.
Optimization is performed using the AdamW~\cite{loshchilov2017decoupled} with $\beta_1 = 0.9$ and $\beta_2 = 0.99$.
The learning rate follows a cosine annealing schedule, reaching a maximum of $3 \times 10^{-4}$. 
To stabilize the early training phase, we apply a cosine warmup over the first 5\% of total iterations. 
All training settings align with those employed in the CGFormer~\cite{yu2024context} baseline.

\subsection{Model Architecture}
Building upon prior works~\cite{cao2022monoscene,huang2023tri,yu2024context}, we employ a 2D UNet image encoder based on a pretrained EfficientNet-B7~\cite{tan2019efficientnet} backbone.
For depth estimation, we adopt a standard stereo-based approach using MobileStereoNet~\cite{shamsafar2022mobilestereonet}.
The view transformation component integrates the depth network architecture introduced in CGFormer~\cite{yu2024context}, along with our proposed C3DFusion module.
This module features a Temporal 3D Point Feature Alignment mechanism, which enhances the fused geometric representation through two key techniques: historical context blurring and current-centric feature densification. 
The refined features are subsequently passed into the attention module, which consists of 3 deformable attention layers for cross-attention and 2 for self-attention. 
Each attention head uses 8 sampling points per reference.
The voxel processing network comprises two branches: a voxel-based branch that adopts a 3D ResNet~\cite{he2016deep} consisting of 3 stages, each containing 2 residual blocks, followed by a feature pyramid network (FPN)~\cite{lin2017feature}; and a TPV-based branch that employs SwinT~\cite{liu2021swin} as the backbone and is likewise followed by an FPN.

\section{Additional Experiments}
\subsection{Quantitative Results}
\subsubsection{Additional Performance Comparisons}
\input{Sections/Table/sup_results_SemanticKITTI_valid}
\input{Sections/Table/sup_results_KITTI360_valid}
We present the validation results on the SemanticKITTI~\cite{behley2019semantickitti} and SSCBench-KITTI-360~\cite{li2024sscbenchlargescale3dsemantic} datasets in Table~\ref{tab:SemKITTI_val} and Table~\ref{tab:KITTI360_val}, respectively.
For SSCBench-KITTI-360, comparisons are limited to models for which official pretrained weights are available.
Across both datasets, our method demonstrates consistent improvements over prior approaches.
Notably, significant gains are observed in the out-of-view (OOV) regions, as shown in the rightmost columns of each table.
Additionally, Table~\ref{tab:SemKITTI_oov_val}, Table~\ref{tab:KITTI360_oov_val} and Table~\ref{tab:KITTI360 _oov_test} report per-class IoU scores in OOV regions, further demonstrating the strength of our approach.
These results confirm the robustness of the proposed C3DFusion, which is designed to complement the current frame by leveraging historical information.

\input{Sections/Table/sup_results_SemanticKITTI_oov_valid}
\input{Sections/Table/sup_results_KITTI360_oov_valid}
\input{Sections/Table/sup_results_KITTI360_oov_test}

\subsection{Additional Ablation Studies}
\subsubsection{Voxel Aggregation Method}
\input{Sections/Table/sup_voxel_aggregation}
Table~\ref{tab:ablation_temp_fusion} summarizes the five strategies for aggregating voxel features across time steps, which we evaluate as follows: (1) Concat$\rightarrow$Linear: Voxel features from all time steps are concatenated along the channel dimension and projected back to the original dimension using a linear layer.
(2) Concat$\rightarrow$Conv.: Similar to (1), but uses a $1\times1$ convolution instead of a linear layer for dimensionality reduction. 
(3) Learnable Weight: A learnable scalar weight is assigned to each time step, normalized with a softmax function. During inference, the weights remain fixed, resulting in a static aggregation.
(4) Dynamic Weighted Sum: Temporal voxel features are concatenated and passed through a linear layer that outputs adaptive weights for each time step. These weights are used to perform a dynamic weighted sum.
(5) Average (Ours): A simple element-wise average of voxel features across all time steps.

Among these, method (5) achieves the best performance. 
We hypothesize that this is due to the nature of 3D scenes, where a large proportion of voxels are empty and initialized with zeros. 
Since our C3DFusion accumulates point features into voxels, learning-based aggregation methods such as (1)–(4) may suffer from the high presence of zero values, which can hinder effective learning. 
In contrast, the average strategy in (5) fully preserves the information of voxels with a higher number of assigned points, leading to better representation of important regions and achieving the best overall performance.



\subsubsection{Volumetric Contribution of Historical Frame Points Across Time Steps}
In Table~\ref{historical proposal}, we analyze the distribution of unfiltered points and their corresponding aggregated voxels across different time steps on the SemanticKITTI validation set.
As the time step moves further into the past, the proportion of unfiltered points decreases sharply—for instance, 84.76\% in the t–1 frame compared to 15.91\% in the t–6 frame.
In contrast, the proportion of voxels to which these points are assigned shows a more moderate decline (from 4.53\% to 3.27\%). 
This finding suggests that, although points from more distant frames are fewer, they can still affect a broad spatial area by occupying numerous unique voxels.
However, such broad voxel coverage from distant frames may not always be desirable, as depth estimation errors tend to increase with distance~\cite{poggi2020uncertainty}, potentially leading to noisy or inaccurate voxel mapping.
\begin{table}[]
\centering
\setlength\tabcolsep{10pt}\resizebox{.8\linewidth}{!}{
\begin{tabular}{@{}c|c|c@{}}
\toprule
\textbf{\begin{tabular}[c]{@{}c@{}}Time Step\end{tabular}} & \multicolumn{1}{c|}{\textbf{\begin{tabular}[c]{@{}l@{}}Unfiltered Points \\(Proportion)\end{tabular}}} &
\multicolumn{1}{c}{\textbf{\begin{tabular}[c]{@{}l@{}}Aggregated Voxels \\(Proportion)\end{tabular}}} 
\\ \midrule
t-1 & 416,612 (84.76\%) & 11,868 (4.53\%) \\
t-2 & 304,249 (61.90\%) & 11,928 (4.55\%) \\
t-3 & 208,878 (42.50\%) & 11,416 (4.35\%) \\
t-4 & 145,361 (29.57\%) & 10,600 (4.04\%) \\
t-5 & 105,017 (21.37\%) & 9,623 (3.67\%) \\
t-6 & 78,215 (15.91\%) & 8,567 (3.27\%) \\ \bottomrule
\end{tabular}}
\caption{Distribution of unfiltered points and aggregated voxels across different time steps. 
}
\label{historical proposal}
\end{table}


\subsection{Qualitative Analysis}
All analyses are conducted on the SemanticKITTI validation split.

\subsubsection{Temporal LSS Fusion vs. C3DFusion}
\begin{figure*}[hbt]
    \centering    \includegraphics[width=1\linewidth]{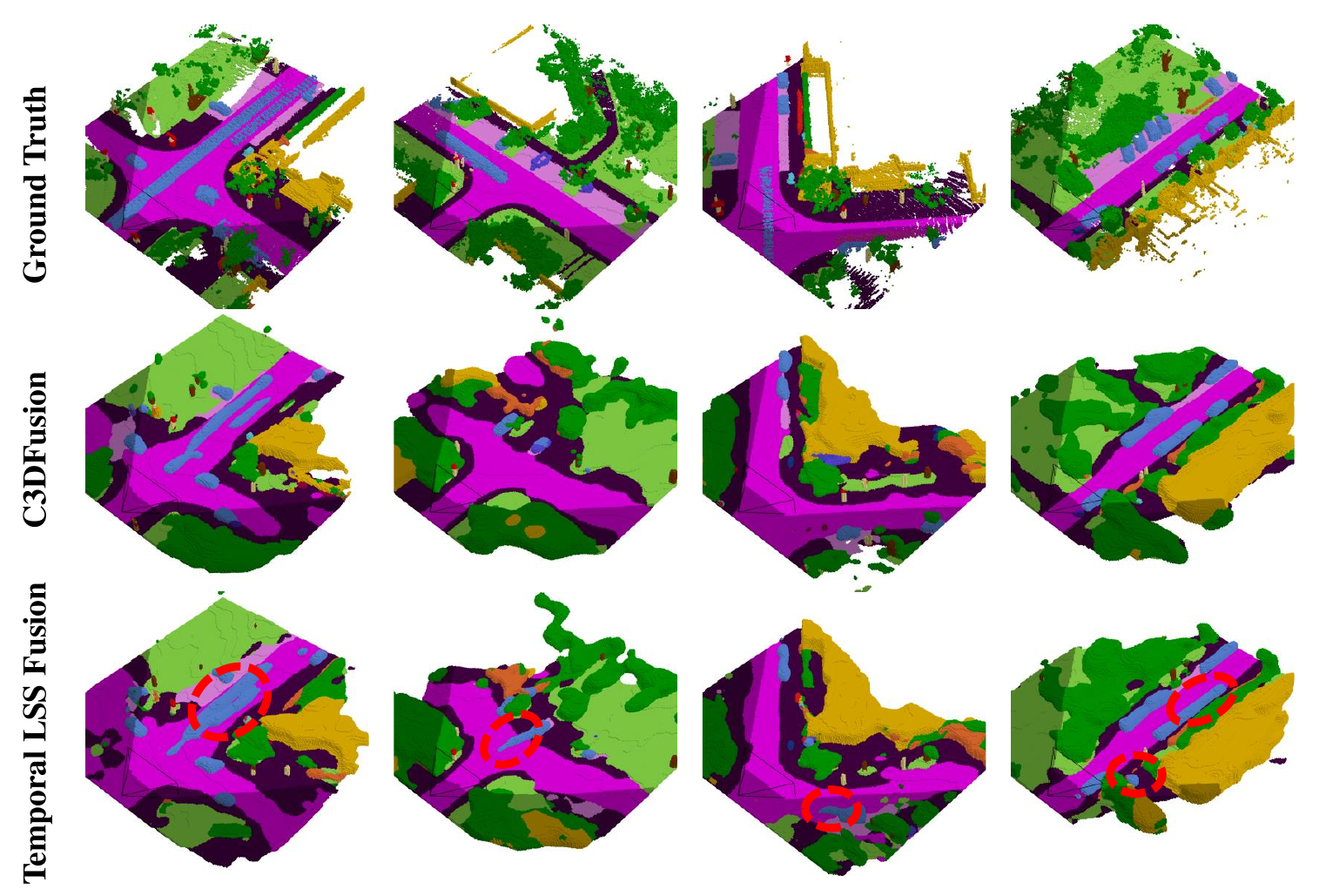}
    \caption{Visual comparison between our model with C3DFusion and the baseline using temporal LSS fusion.}
    \label{lssvsours}
\end{figure*}
Figure~\ref{lssvsours} presents a visual comparison of the prediction outputs from our model equipped with C3DFusion and a baseline model in which C3DFusion is replaced by the temporal LSS fusion method described in Table~3 of the manuscript.
As hypothesized in the Method section, naively fusing long-tailed, sparsely densified LSS features introduces geometric noise—an effect clearly observable in these results.
In particular, temporal LSS fusion leads to inaccurate completions, such as tail-shaped smoothing artifacts (see the bottoms of the second, third, and fourth columns) and unintended merging of independent object completions (see the bottoms of the first and fourth columns).
In contrast, since C3DFusion explicitly merges sparse point clouds within a unified 3D space, it does not suffer from these side effects, further demonstrating its superiority.


\subsubsection{Effect of Current-Centric Feature Densification}
\begin{figure*}[hbt!]
    \centering
    \includegraphics[width=1\linewidth]{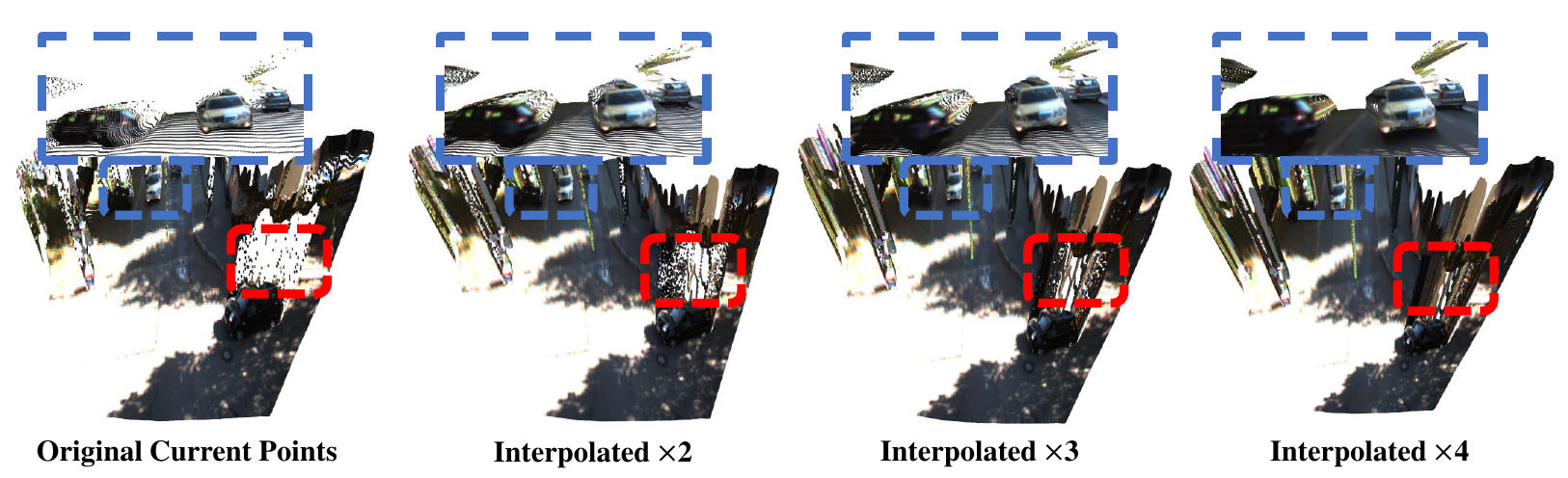}
    \caption{Visual comparison of backprojected 3D point clouds of the current frame with varying interpolation factors for current-centric feature densification.}
    \label{ccfd}
\end{figure*}
Figure~\ref{ccfd} shows the backprojected 3D point cloud visualization of the current frame, with varying interpolation factors used for current-centric feature densification.
As the interpolation factor increases, points representing distant objects become denser, resulting in visually plausible and more refined geometry.
However, we also observe an increase in afterimage artifacts near object edges, which leads to geometric inaccuracies and may degrade SSC performance.
These findings indicate that selecting the degree of current-centric feature densification requires careful consideration, as there is a trade-off between enhancing geometric context and increasing the risk of geometric noise.

\subsubsection{Additional Qualitative Results}
\begin{figure*}[hbt!]
    \centering
    \includegraphics[width=1\linewidth]{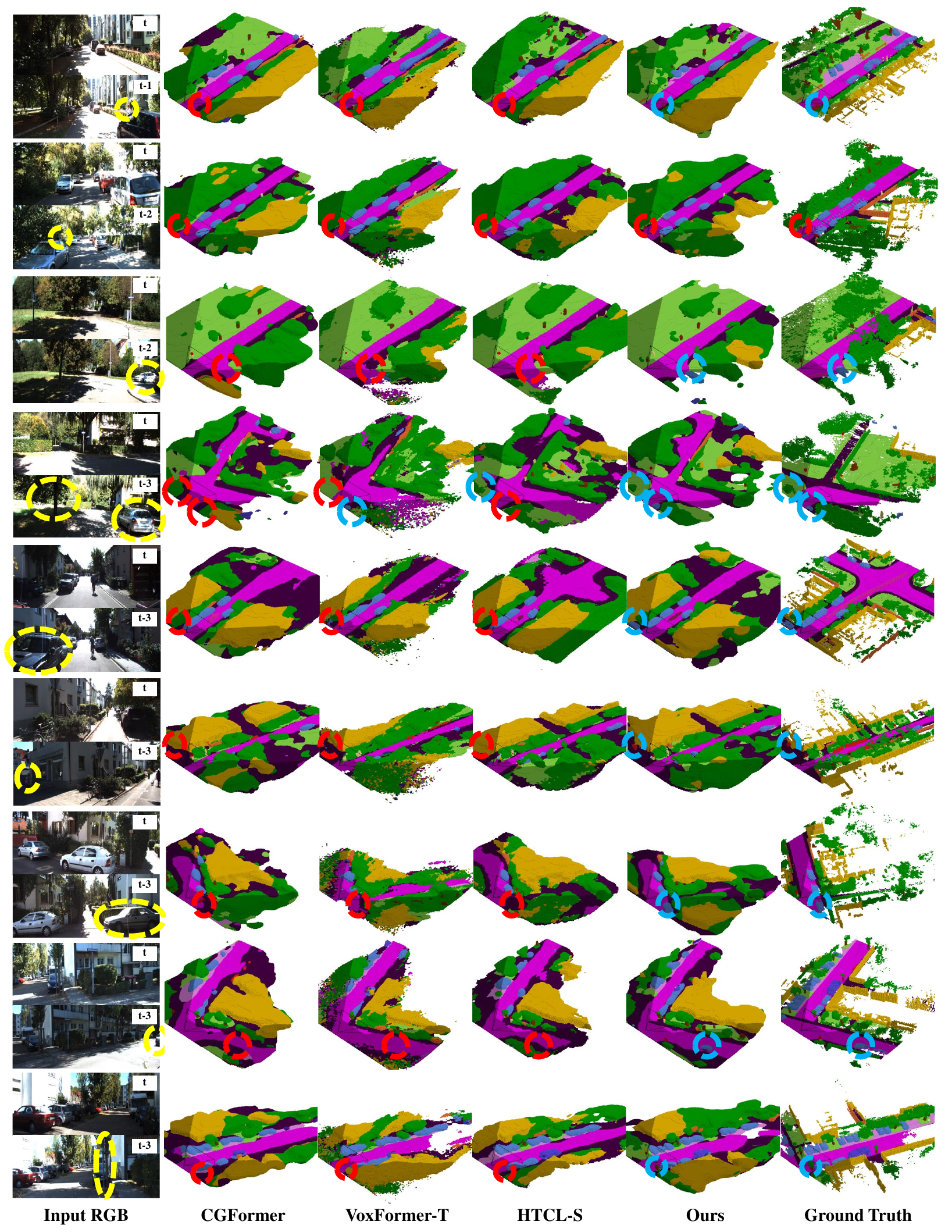}
    \caption{Visual comparison of our model against other recent camera-based methods on the SemanticKITTI validation set.}
    \label{qualitative_supple}
\end{figure*}
In Figure~\ref{qualitative_supple}, we present additional qualitative results comparing our model with existing open-source camera-based approaches.
In these supplementary cases, our model successfully captures objects that are invisible or occluded in the current frame but were observed in previous frames, whereas other models fail to do so.
These results further demonstrate the superior performance of our approach, particularly in reconstructing regions beyond the current camera view.

\section{Pseudocodes}
To facilitate understanding, we provide PyTorch-style~\cite{paszke2019pytorch} pseudocode for historical context blurring and current-centric feature densification in Algorithm~\ref{alg:hcb} and Algorithm~\ref{alg:ccfd}, respectively.

\lstset{
  basicstyle=\ttfamily\scriptsize,
  keywordstyle=\color{blue},
  commentstyle=\color{gray},
  stringstyle=\color{orange},
  breaklines=true,
  breakindent=0pt,
  frame=single,
  language=Python,
  tabsize=2,
  showstringspaces=false,
  numbers=none,        
  numbersep=0pt,     
  xleftmargin=0pt     
}

\begin{center}
\begin{algorithm}[hbt!]
\scriptsize
\caption{PyTorch Style Pseudocode of HCB.}
\vspace{1em} 
\begin{lstlisting}
import torch
import torch.nn as nn

class HCB(nn.Module):
    """
    HCB = Historical Context Blurring.
    This technique assigns higher weights to closer depth values in historical frames by computing inverse depth-based weights.
    """

    def __init__(self):
        super().__init__()

    def forward(self, depth_values):
        # Normalize depth to [0, 1]
        depth_min = depth_values.min()
        depth_max = depth_values.max()
        norm_x = (depth_values - depth_min) / (depth_max - depth_min)

        # Invert so that closer -> higher weight
        HCB_x = 1.0 - norm_x

        return HCB_x
        
\end{lstlisting}
\vspace{1em}
\label{alg:hcb}
\end{algorithm}
\end{center}

\begin{center}
\begin{algorithm}[hbt!]
\scriptsize
\caption{PyTorch Style Pseudocode of CCFD.}
\vspace{1em} 
\begin{lstlisting}
import torch
import torch.nn as nn
import torch.nn.functional as F

class CCFD(nn.Module):
    """
    CCFD = Current-Centric Feature Densification.
    This technique generates dense 3D points from the current frame by upsampling both the image plane grid and the depth map.
    """

    def __init__(self, factor=2):
        super().__init__()
        self.factor = factor

    def create_image_grid(self, H, W):
        xs = torch.linspace(0, W - 1, W)
        xs = xs.view(1, 1, 1, W).expand(1, 1, H, W)
        ys = torch.linspace(0, H - 1, H)
        ys = ys.view(1, 1, H, 1).expand(1, 1, H, W)
        
        image_grid = torch.cat([xs, ys], dim=1) 
        
        return image_grid

    def forward(self, feats, depth):
        B, _, H, W = depth.shape
        image_grid = self.create_image_grid(H, W)
        image_grid = image_grid.repeat(B, 1, 1, 1)

        # Upsample
        up_image_grid = F.interpolate(image_grid,
                size=(H * self.factor, W * self.factor),
                mode='bilinear', align_corners=False)
        up_depth = F.interpolate(depth,
                size=(H * self.factor, W * self.factor),
                mode='bilinear', align_corners=False)
        up_feats = F.interpolate(feats,
                size=(H * self.factor, W * self.factor),
                mode='bilinear', align_corners=False)

        return up_image_grid, up_depth, up_feats

\end{lstlisting}
\vspace{1em}
\label{alg:ccfd}
\end{algorithm}
\end{center}

%% file: Sections/Table/sup_results_SemanticKITTI_valid.tex

\definecolor{custom_road}{RGB}{255, 0, 255}
\definecolor{custom_sidewalk}{RGB}{75, 0, 75}
\definecolor{custom_parking}{RGB}{255, 150, 255}
\definecolor{custom_other_ground}{RGB}{175, 0, 75}
\definecolor{custom_building}{RGB}{255, 200, 0}
\definecolor{custom_car}{RGB}{100, 150, 245}
\definecolor{custom_truck}{RGB}{80, 30, 180}
\definecolor{custom_bicycle}{RGB}{100, 230, 245}
\definecolor{custom_motorcycle}{RGB}{30, 60, 150}
\definecolor{custom_other_veh}{RGB}{0, 0, 255}
\definecolor{custom_vegetation}{RGB}{0, 175, 0}
\definecolor{custom_trunk}{RGB}{135, 60, 0}
\definecolor{custom_terrain}{RGB}{150, 240, 80}
\definecolor{custom_person}{RGB}{255, 30, 30}
\definecolor{custom_bicyclist}{RGB}{255, 40, 200}
\definecolor{custom_motorcyclist}{RGB}{150, 30, 90}
\definecolor{custom_fence}{RGB}{255, 120, 50}
\definecolor{custom_pole}{RGB}{255, 240, 150}
\definecolor{custom_traf_sign}{RGB}{255, 0, 0}

\begin{table*}[hbt!]
\centering
\resizebox{\textwidth}{!}{%
\begin{tabular}{lccc|clclclclclclclclclclclclclclclclclclcl|cc}
\toprule
 &
   &
  \multicolumn{1}{l}{} &
  \multicolumn{1}{l|}{} &
  \multicolumn{2}{l}{} &
  \multicolumn{2}{l}{} &
  \multicolumn{2}{l}{} &
  \multicolumn{2}{l}{} &
  \multicolumn{2}{l}{} &
  \multicolumn{2}{l}{} &
  \multicolumn{2}{l}{} &
  \multicolumn{2}{l}{} &
  \multicolumn{2}{l}{} &
  \multicolumn{2}{l}{} &
  \multicolumn{2}{l}{} &
  \multicolumn{2}{l}{} &
  \multicolumn{2}{l}{} &
  \multicolumn{2}{l}{} &
  \multicolumn{2}{l}{} &
  \multicolumn{2}{l}{} &
  \multicolumn{2}{l}{} &
  \multicolumn{2}{l}{} &
  \multicolumn{2}{l|}{} &
  \multicolumn{2}{l}{} \\
 &
   &
  \multicolumn{1}{l}{} &
  \multicolumn{1}{l|}{} &
  \multicolumn{2}{l}{\rotatebigtext{road} \newline \textcolor{white}{\rotatetinytext{1}} \newline \rotatetinytext{(15.30\%)}} &
  \multicolumn{2}{l}{\rotatebigtext{sidewalk} \newline \textcolor{white}{\rotatetinytext{1}} \newline \rotatetinytext{(11.13\%)}} &
  \multicolumn{2}{l}{\rotatebigtext{parking} \newline \textcolor{white}{\rotatetinytext{1}}\newline \rotatetinytext{(1.12\%)}} &
  \multicolumn{2}{l}{\rotatebigtext{other-grnd.} \newline \textcolor{white}{\rotatetinytext{1}}\newline \rotatetinytext{(0.56\%)}} &
  \multicolumn{2}{l}{\rotatebigtext{building} \newline \textcolor{white}{\rotatetinytext{1}}\newline \rotatetinytext{(14.1\%)}} &
  \multicolumn{2}{l}{\rotatebigtext{car} \newline \textcolor{white}{\rotatetinytext{1}}\newline \rotatetinytext{(3.92\%)}} &
  \multicolumn{2}{l}{\rotatebigtext{truck} \newline \textcolor{white}{\rotatetinytext{1}}\newline \rotatetinytext{(0.16\%)}} &
  \multicolumn{2}{l}{\rotatebigtext{bicycle} \newline \textcolor{white}{\rotatetinytext{1}}\newline \rotatetinytext{(0.03\%)}} &
  \multicolumn{2}{l}{\rotatebigtext{motorcycle} \newline \textcolor{white}{\rotatetinytext{1}}\newline \rotatetinytext{(0.03\%)}} &
  \multicolumn{2}{l}{\rotatebigtext{other-veh.} \newline \textcolor{white}{\rotatetinytext{1}}\newline \rotatetinytext{(0.20\%)}} &
  \multicolumn{2}{l}{\rotatebigtext{vegetation} \newline \textcolor{white}{\rotatetinytext{1}}\newline \rotatetinytext{(39.3\%)}} &
  \multicolumn{2}{l}{\rotatebigtext{trunk} \newline \textcolor{white}{\rotatetinytext{1}}\newline \rotatetinytext{(0.51\%)}} &
  \multicolumn{2}{l}{\rotatebigtext{terrain} \newline \textcolor{white}{\rotatetinytext{1}}\newline \rotatetinytext{(9.17\%)}} &
  \multicolumn{2}{l}{\rotatebigtext{person} \newline \textcolor{white}{\rotatetinytext{1}}\newline \rotatetinytext{(0.07\%)}} &
  \multicolumn{2}{l}{\rotatebigtext{bicyclist} \newline \textcolor{white}{\rotatetinytext{1}}\newline \rotatetinytext{(0.07\%)}} &
  \multicolumn{2}{l}{\rotatebigtext{motorcyclist} \newline \textcolor{white}{\rotatetinytext{1}}\newline \rotatetinytext{(0.05\%)}} &
  \multicolumn{2}{l}{\rotatebigtext{fence} \newline \textcolor{white}{\rotatetinytext{1}}\newline \rotatetinytext{(3.90\%)}} &
  \multicolumn{2}{l}{\rotatebigtext{pole} \newline \textcolor{white}{\rotatetinytext{1}}\newline \rotatetinytext{(0.29\%)}} &
  \multicolumn{2}{l|}{\rotatebigtext{traf.-sign} \newline \textcolor{white}{\rotatetinytext{1}}\newline \rotatetinytext{(0.08\%)}} &
  \multicolumn{2}{c}{\textbf{OOV (val.)}} \\
\textbf{Method} &
  \textbf{Input} &
  \textbf{IoU} &
  \textbf{mIoU} &
  \multicolumn{2}{c}{\colorboxlabel{custom_road}{}} &
  \multicolumn{2}{c}{\colorboxlabel{custom_sidewalk}{}} &
  \multicolumn{2}{c}{\colorboxlabel{custom_parking}{}} &
  \multicolumn{2}{c}{\colorboxlabel{custom_other_ground}{}} &
  \multicolumn{2}{c}{\colorboxlabel{custom_building}{}} &
  \multicolumn{2}{c}{\colorboxlabel{custom_car}{}} &
  \multicolumn{2}{c}{\colorboxlabel{custom_truck}{}} &
  \multicolumn{2}{c}{\colorboxlabel{custom_bicycle}{}} &
  \multicolumn{2}{c}{\colorboxlabel{custom_motorcycle}{}} &
  \multicolumn{2}{c}{\colorboxlabel{custom_other_veh}{}} &
  \multicolumn{2}{c}{\colorboxlabel{custom_vegetation}{}} &
  \multicolumn{2}{c}{\colorboxlabel{custom_trunk}{}} &
  \multicolumn{2}{c}{\colorboxlabel{custom_terrain}{}} &
  \multicolumn{2}{c}{\colorboxlabel{custom_person}{}} &
  \multicolumn{2}{c}{\colorboxlabel{custom_bicyclist}{}} &
  \multicolumn{2}{c}{\colorboxlabel{custom_motorcyclist}{}} &
  \multicolumn{2}{c}{\colorboxlabel{custom_fence}{}} &
  \multicolumn{2}{c}{\colorboxlabel{custom_pole}{}} &
  \multicolumn{2}{c|}{\colorboxlabel{custom_traf_sign}{}} &
  \textbf{IoU} &
  \textbf{mIoU} \\ \midrule
   \multicolumn{44}{c}{\textbf{\textit{Single-Frame-Based}}} \\ \midrule
\multicolumn{1}{l|}{MonoScene} &
  \multicolumn{1}{c|}{Mono} &
   36.86&
   11.08&
  \multicolumn{2}{c}{56.52} &
  \multicolumn{2}{c}{26.72} &
  \multicolumn{2}{c}{14.27} &
  \multicolumn{2}{c}{0.46} &
  \multicolumn{2}{c}{14.09} &
  \multicolumn{2}{c}{23.26} &
  \multicolumn{2}{c}{6.98} &
  \multicolumn{2}{c}{0.61} &
  \multicolumn{2}{c}{0.45} &
  \multicolumn{2}{c}{1.48} &
  \multicolumn{2}{c}{17.89} &
  \multicolumn{2}{c}{2.81} &
  \multicolumn{2}{c}{29.64} &
  \multicolumn{2}{c}{1.86} &
  \multicolumn{2}{c}{1.20} &
  \multicolumn{2}{c}{0.00} &
  \multicolumn{2}{c}{5.84} &
  \multicolumn{2}{c}{4.14} &
  \multicolumn{2}{c|}{2.25} &
  31.07 &
   7.02\\
\multicolumn{1}{l|}{TPVFormer} &
  \multicolumn{1}{c|}{Mono} &
   35.61&
   11.36&
  \multicolumn{2}{c}{56.50} &
  \multicolumn{2}{c}{25.87} &
  \multicolumn{2}{c}{20.60} &
  \multicolumn{2}{c}{0.85} &
  \multicolumn{2}{c}{13.88} &
  \multicolumn{2}{c}{23.81} &
  \multicolumn{2}{c}{8.08} &
  \multicolumn{2}{c}{0.36} &
  \multicolumn{2}{c}{0.05} &
  \multicolumn{2}{c}{4.35} &
  \multicolumn{2}{c}{16.92} &
  \multicolumn{2}{c}{2.26} &
  \multicolumn{2}{c}{30.38} &
  \multicolumn{2}{c}{0.51} &
  \multicolumn{2}{c}{0.89} &
  \multicolumn{2}{c}{0.00} &
  \multicolumn{2}{c}{5.94} &
  \multicolumn{2}{c}{3.14} &
  \multicolumn{2}{c|}{1.52} &
   30.76&
   7.87\\
\multicolumn{1}{l|}{OccFormer} &
  \multicolumn{1}{c|}{Mono} &
   36.50&
   13.46&
  \multicolumn{2}{c}{58.85} &
  \multicolumn{2}{c}{26.88} &
  \multicolumn{2}{c}{19.61} &
  \multicolumn{2}{c}{0.31} &
  \multicolumn{2}{c}{14.40} &
  \multicolumn{2}{c}{25.09} &
  \multicolumn{2}{c}{\underline{25.53}} &
  \multicolumn{2}{c}{0.81} &
  \multicolumn{2}{c}{1.19} &
  \multicolumn{2}{c}{8.52} &
  \multicolumn{2}{c}{19.63} &
  \multicolumn{2}{c}{3.93} &
  \multicolumn{2}{c}{32.62} &
  \multicolumn{2}{c}{2.78} &
  \multicolumn{2}{c}{2.82} &
  \multicolumn{2}{c}{0.00} &
  \multicolumn{2}{c}{5.61} &
  \multicolumn{2}{c}{4.26} &
  \multicolumn{2}{c|}{2.86} &
   30.46&
   8.68\\
\multicolumn{1}{l|}{Symphonies} &
  \multicolumn{1}{c|}{Stereo} &
   41.92&
   14.89&
  \multicolumn{2}{c}{56.37} &
  \multicolumn{2}{c}{27.58} &
  \multicolumn{2}{c}{15.28} &
  \multicolumn{2}{c}{0.95} &
  \multicolumn{2}{c}{21.64} &
  \multicolumn{2}{c}{28.68} &
  \multicolumn{2}{c}{20.44} &
  \multicolumn{2}{c}{2.54} &
  \multicolumn{2}{c}{2.82} &
  \multicolumn{2}{c}{13.89} &
  \multicolumn{2}{c}{25.72} &
  \multicolumn{2}{c}{6.60} &
  \multicolumn{2}{c}{30.87} &
  \multicolumn{2}{c}{3.52} &
  \multicolumn{2}{c}{2.24} &
  \multicolumn{2}{c}{{0.00}} &
  \multicolumn{2}{c}{8.40} &
  \multicolumn{2}{c}{9.57} &
  \multicolumn{2}{c|}{5.76} &
   23.48&
   6.40\\
\multicolumn{1}{l|}{CGFormer} &
  \multicolumn{1}{c|}{Stereo} &
   \underline{45.99}&
   16.87&
  \multicolumn{2}{c}{65.51} &
  \multicolumn{2}{c}{32.31} &
  \multicolumn{2}{c}{20.82} &
  \multicolumn{2}{c}{0.16} &
  \multicolumn{2}{c}{23.52} &
  \multicolumn{2}{c}{34.32} &
  \multicolumn{2}{c}{19.44} &
  \multicolumn{2}{c}{4.61} &
  \multicolumn{2}{c}{2.71} &
  \multicolumn{2}{c}{7.67} &
  \multicolumn{2}{c}{26.93} &
  \multicolumn{2}{c}{8.83} &
  \multicolumn{2}{c}{\underline{39.54}} &
  \multicolumn{2}{c}{2.38} &
  \multicolumn{2}{c}{4.08} &
  \multicolumn{2}{c}{0.00} &
  \multicolumn{2}{c}{9.20} &
  \multicolumn{2}{c}{10.67} &
  \multicolumn{2}{c|}{\underline{7.84}} &
   33.54&
   9.06\\
\multicolumn{1}{l|}{L2COcc-C} &
  \multicolumn{1}{c|}{Stereo} &
   45.46&
   16.72&
  \multicolumn{2}{c}{64.89} &
  \multicolumn{2}{c}{33.82} &
  \multicolumn{2}{c}{22.90} &
  \multicolumn{2}{c}{0.32} &
  \multicolumn{2}{c}{23.65} &
  \multicolumn{2}{c}{33.77} &
  \multicolumn{2}{c}{8.61} &
  \multicolumn{2}{c}{4.14} &
  \multicolumn{2}{c}{{5.22}} &
  \multicolumn{2}{c}{9.14} &
  \multicolumn{2}{c}{\underline{27.64}} &
  \multicolumn{2}{c}{9.06} &
  \multicolumn{2}{c}{39.32} &
  \multicolumn{2}{c}{\textbf{4.30}} &
  \multicolumn{2}{c}{2.28} &
  \multicolumn{2}{c}{0.00} &
  \multicolumn{2}{c}{8.85} &
  \multicolumn{2}{c}{\underline{12.17}} &
  \multicolumn{2}{c|}{7.59} &
   32.24&
   8.55\\
\multicolumn{1}{l|}{ScanSSC} &
  \multicolumn{1}{c|}{Stereo} &
   45.95&
   17.12&
  \multicolumn{2}{c}{66.70} &
  \multicolumn{2}{c}{33.87} &
  \multicolumn{2}{c}{{24.75}} &
  \multicolumn{2}{c}{0.20} &
  \multicolumn{2}{c}{22.42} &
  \multicolumn{2}{c}{34.16} &
  \multicolumn{2}{c}{14.78} &
  \multicolumn{2}{c}{3.81} &
  \multicolumn{2}{c}{2.19} &
  \multicolumn{2}{c}{9.81} &
  \multicolumn{2}{c}{27.59} &
  \multicolumn{2}{c}{8.31} &
  \multicolumn{2}{c}{39.26} &
  \multicolumn{2}{c}{2.87} &
  \multicolumn{2}{c}{\underline{5.93}} &
  \multicolumn{2}{c}{0.00} &
  \multicolumn{2}{c}{10.52} &
  \multicolumn{2}{c}{10.52} &
  \multicolumn{2}{c|}{7.50} &
   33.60&
   9.50\\
\multicolumn{1}{l|}{L2COcc-D} &
  \multicolumn{1}{c|}{Stereo} &
   45.30&
  \underline{18.22} &
  \multicolumn{2}{c}{\textbf{68.51}} &
  \multicolumn{2}{c}{\textbf{36.53}} &
  \multicolumn{2}{c}{\underline{25.33}} &
  \multicolumn{2}{c}{{0.46}} &
  \multicolumn{2}{c}{22.60} &
  \multicolumn{2}{c}{\underline{34.95}} &
  \multicolumn{2}{c}{17.88} &
  \multicolumn{2}{c}{\underline{4.99}} &
  \multicolumn{2}{c}{\textbf{7.11}} &
  \multicolumn{2}{c}{{13.19}} &
  \multicolumn{2}{c}{27.39} &
  \multicolumn{2}{c}{{9.25}} &
  \multicolumn{2}{c}{{42.16}} &
  \multicolumn{2}{c}{3.10} &
  \multicolumn{2}{c}{1.98} &
  \multicolumn{2}{c}{0.00} &
  \multicolumn{2}{c}{{11.17}} &
  \multicolumn{2}{c}{{11.84}} &
  \multicolumn{2}{c|}{{7.74}} &
   31.85&
   10.05\\ \midrule
  \multicolumn{44}{c}{\textbf{\textit{Temporal-Frame-Based}}} \\ \midrule
\multicolumn{1}{l|}{VoxFormer-T} &
  \multicolumn{1}{c|}{Stereo} &
   \underline{44.15}&
   13.35&
  \multicolumn{2}{c}{53.57} &
  \multicolumn{2}{c}{26.52} &
  \multicolumn{2}{c}{19.69} &
  \multicolumn{2}{c}{0.42} &
  \multicolumn{2}{c}{19.54} &
  \multicolumn{2}{c}{26.54} &
  \multicolumn{2}{c}{7.26} &
  \multicolumn{2}{c}{1.28} &
  \multicolumn{2}{c}{0.56} &
  \multicolumn{2}{c}{7.81} &
  \multicolumn{2}{c}{26.10} &
  \multicolumn{2}{c}{6.10} &
  \multicolumn{2}{c}{33.06} &
  \multicolumn{2}{c}{1.93} &
  \multicolumn{2}{c}{1.97} &
  \multicolumn{2}{c}{0.00} &
  \multicolumn{2}{c}{7.31} &
  \multicolumn{2}{c}{9.15} &
  \multicolumn{2}{c|}{4.94} &
  \underline{40.21} &
  \underline{11.58} \\
\multicolumn{1}{l|}{HTCL-S} &
  \multicolumn{1}{c|}{Stereo} &
   45.51&
   17.13&
  \multicolumn{2}{c}{63.70} &
  \multicolumn{2}{c}{32.48} &
  \multicolumn{2}{c}{23.27} &
  \multicolumn{2}{c}{0.14} &
  \multicolumn{2}{c}{24.13} &
  \multicolumn{2}{c}{34.30} &
  \multicolumn{2}{c}{{20.72}} &
  \multicolumn{2}{c}{3.99} &
  \multicolumn{2}{c}{2.80} &
  \multicolumn{2}{c}{11.99} &
  \multicolumn{2}{c}{26.96} &
  \multicolumn{2}{c}{8.79} &
  \multicolumn{2}{c}{37.73} &
  \multicolumn{2}{c}{2.56} &
  \multicolumn{2}{c}{2.30} &
  \multicolumn{2}{c}{0.00} &
  \multicolumn{2}{c}{11.22} &
  \multicolumn{2}{c}{11.49} &
  \multicolumn{2}{c|}{6.95} &
   33.14&
   9.04\\
\multicolumn{1}{l|}{Hi-SOP} &
  \multicolumn{1}{c|}{Stereo} &
   45.56&
   18.19&
  \multicolumn{2}{c}{63.86} &
  \multicolumn{2}{c}{32.71} &
  \multicolumn{2}{c}{\textbf{25.94}} &
  \multicolumn{2}{c}{\underline{1.18}} &
  \multicolumn{2}{c}{24.56} &
  \multicolumn{2}{c}{34.07} &
  \multicolumn{2}{c}{25.25} &
  \multicolumn{2}{c}{4.42} &
  \multicolumn{2}{c}{3.96} &
  \multicolumn{2}{c}{\underline{16.96}} &
  \multicolumn{2}{c}{26.61} &
  \multicolumn{2}{c}{\underline{9.92}} &
  \multicolumn{2}{c}{38.89} &
  \multicolumn{2}{c}{3.36} &
  \multicolumn{2}{c}{\textbf{6.48}} &
  \multicolumn{2}{c}{0.00} &
  \multicolumn{2}{c}{9.30} &
  \multicolumn{2}{c}{11.41} &
  \multicolumn{2}{c|}{6.70} &
  - &
  - \\
\multicolumn{1}{l|}{FlowScene} &
  \multicolumn{1}{c|}{Stereo} &
   45.01&
   18.13&
  \multicolumn{2}{c}{63.72} &
  \multicolumn{2}{c}{32.10} &
  \multicolumn{2}{c}{22.20} &
  \multicolumn{2}{c}{\textbf{1.31}} &
  \multicolumn{2}{c}{\underline{25.63}} &
  \multicolumn{2}{c}{33.33} &
  \multicolumn{2}{c}{\textbf{33.47}} &
  \multicolumn{2}{c}{2.36} &
  \multicolumn{2}{c}{5.09} &
  \multicolumn{2}{c}{\textbf{16.99}} &
  \multicolumn{2}{c}{26.35} &
  \multicolumn{2}{c}{8.68} &
  \multicolumn{2}{c}{36.73} &
  \multicolumn{2}{c}{\underline{3.79}} &
  \multicolumn{2}{c}{{1.92}} &
  \multicolumn{2}{c}{{0.00}} &
  \multicolumn{2}{c}{\underline{12.05}} &
  \multicolumn{2}{c}{11.65} &
  \multicolumn{2}{c|}{7.05} &
  - &
  - \\
\multicolumn{1}{l|}{CF-SSC} &
  \multicolumn{1}{c|}{Stereo} &
  - &
  - &
  \multicolumn{2}{c}{-} &
  \multicolumn{2}{c}{-} &
  \multicolumn{2}{c}{-} &
  \multicolumn{2}{c}{-} &
  \multicolumn{2}{c}{{-}} &
  \multicolumn{2}{c}{-} &
  \multicolumn{2}{c}{-} &
  \multicolumn{2}{c}{-} &
  \multicolumn{2}{c}{-} &
  \multicolumn{2}{c}{-} &
  \multicolumn{2}{c}{{-}} &
  \multicolumn{2}{c}{-} &
  \multicolumn{2}{c}{-} &
  \multicolumn{2}{c}{-} &
  \multicolumn{2}{c}{-} &
  \multicolumn{2}{c}{-} &
  \multicolumn{2}{c}{-} &
  \multicolumn{2}{c}{-} &
  \multicolumn{2}{c|}{-} &
  - &
  - \\
\rowcolor[HTML]{EFEFEF} 
\multicolumn{1}{l|}{\cellcolor[HTML]{EFEFEF}\textbf{Ours}} &
  \multicolumn{1}{c|}{\cellcolor[HTML]{EFEFEF}\textbf{Stereo}} &
  \textbf{49.53} &
  \textbf{19.31} &
  \multicolumn{2}{c}{\cellcolor[HTML]{EFEFEF}\underline{68.04}} &
  \multicolumn{2}{c}{\cellcolor[HTML]{EFEFEF}\underline{35.03}} &
  \multicolumn{2}{c}{\cellcolor[HTML]{EFEFEF}{22.30}} &
  \multicolumn{2}{c}{\cellcolor[HTML]{EFEFEF}{0.46}} &
  \multicolumn{2}{c}{\cellcolor[HTML]{EFEFEF}\textbf{30.22}} &
  \multicolumn{2}{c}{\cellcolor[HTML]{EFEFEF}\textbf{37.05}} &
  \multicolumn{2}{c}{\cellcolor[HTML]{EFEFEF}{12.21}} &
  \multicolumn{2}{c}{\cellcolor[HTML]{EFEFEF}\textbf{6.09}} &
  \multicolumn{2}{c}{\cellcolor[HTML]{EFEFEF}\underline{5.79}} &
  \multicolumn{2}{c}{\cellcolor[HTML]{EFEFEF}{12.69}} &
  \multicolumn{2}{c}{\cellcolor[HTML]{EFEFEF}\textbf{31.99}} &
  \multicolumn{2}{c}{\cellcolor[HTML]{EFEFEF}\textbf{12.21}} &
  \multicolumn{2}{c}{\cellcolor[HTML]{EFEFEF}\textbf{43.51}} &
  \multicolumn{2}{c}{\cellcolor[HTML]{EFEFEF}{1.99}} &
  \multicolumn{2}{c}{\cellcolor[HTML]{EFEFEF}{0.91}} &
  \multicolumn{2}{c}{\cellcolor[HTML]{EFEFEF}{0.00}} &
  \multicolumn{2}{c}{\cellcolor[HTML]{EFEFEF}\textbf{12.43}} &
  \multicolumn{2}{c}{\cellcolor[HTML]{EFEFEF}\textbf{14.83}} &
  \multicolumn{2}{c|}{\cellcolor[HTML]{EFEFEF}\textbf{9.71}} &
   \textbf{44.37}&
  \textbf{17.17} \\ \bottomrule

\end{tabular}%
}
\caption{Quantitative results on SemanticKITTI validation set. ‘OOV’ on the right indicates performance on out-of-view regions, evaluated on the validation set using our implementation. Since CF-SSC~\cite{lu2025one} has no reported performance and no available weights, it is denoted as `-'.
\textbf{Bold} / \underline{underline} highlight the best / second-best, respectively.}
\label{tab:SemKITTI_val}
    
\end{table*}

%% file: Sections/Table/sup_results_KITTI360_valid.tex

\definecolor{custom_road}{RGB}{255, 0, 255}
\definecolor{custom_sidewalk}{RGB}{75, 0, 75}
\definecolor{custom_parking}{RGB}{255, 150, 255}
\definecolor{custom_other_ground}{RGB}{175, 0, 75}
\definecolor{custom_building}{RGB}{255, 200, 0}
\definecolor{custom_car}{RGB}{100, 150, 245}
\definecolor{custom_truck}{RGB}{80, 30, 180}
\definecolor{custom_bicycle}{RGB}{100, 230, 245}
\definecolor{custom_motorcycle}{RGB}{30, 60, 150}
\definecolor{custom_other_veh}{RGB}{0, 0, 255}
\definecolor{custom_vegetation}{RGB}{0, 175, 0}
\definecolor{custom_trunk}{RGB}{135, 60, 0}
\definecolor{custom_terrain}{RGB}{150, 240, 80}
\definecolor{custom_person}{RGB}{255, 30, 30}
\definecolor{custom_bicyclist}{RGB}{255, 40, 20}
\definecolor{custom_motorcyclist}{RGB}{150, 30, 90}
\definecolor{custom_fence}{RGB}{255, 120, 50}
\definecolor{custom_pole}{RGB}{255, 240, 150}
\definecolor{custom_traf_sign}{RGB}{255, 0, 0}
\definecolor{custom_other_struct}{RGB}{255, 150, 0}
\definecolor{custom_other_obj}{RGB}{50, 255, 255}

\begin{table*}[hbt!]
\centering
\resizebox{\textwidth}{!}{%
\begin{tabular}{lccc|cccccccccccccccccccccccccccccccccccc|cc}
\toprule
 &
   &
  \multicolumn{1}{l}{} &
   &
  \multicolumn{2}{c}{} &
  \multicolumn{2}{c}{} &
  \multicolumn{2}{c}{} &
  \multicolumn{2}{c}{} &
  \multicolumn{2}{c}{} &
  \multicolumn{2}{c}{} &
  \multicolumn{2}{c}{} &
  \multicolumn{2}{c}{} &
  \multicolumn{2}{c}{} &
  \multicolumn{2}{c}{} &
  \multicolumn{2}{c}{} &
  \multicolumn{2}{c}{} &
  \multicolumn{2}{c}{} &
  \multicolumn{2}{c}{} &
  \multicolumn{2}{c}{} &
  \multicolumn{2}{c}{} &
  \multicolumn{2}{c}{} &
  \multicolumn{2}{c|}{} &
  \multicolumn{2}{c}{} \\
 &
   &
   &
   &
  \multicolumn{2}{c}{\rotatebigtext{car} \newline \textcolor{white}{\rotatetinytext{1}} \newline \rotatetinytext{(2.85\%)}} &
  \multicolumn{2}{c}{\rotatebigtext{bicycle} \newline \textcolor{white}{\rotatetinytext{1}} \newline \rotatetinytext{(0.01\%)}} &
  \multicolumn{2}{c}{\rotatebigtext{motorcycle} \newline \textcolor{white}{\rotatetinytext{1}}\newline \rotatetinytext{(0.01\%)}} &
  \multicolumn{2}{c}{\rotatebigtext{truck} \newline \textcolor{white}{\rotatetinytext{1}}\newline \rotatetinytext{(0.16\%)}} &
  \multicolumn{2}{c}{\rotatebigtext{other-veh.} \newline \textcolor{white}{\rotatetinytext{1}}\newline \rotatetinytext{(5.75\%)}} &
  \multicolumn{2}{c}{\rotatebigtext{person} \newline \textcolor{white}{\rotatetinytext{1}}\newline \rotatetinytext{(0.02\%)}} &
  \multicolumn{2}{c}{\rotatebigtext{road} \newline \textcolor{white}{\rotatetinytext{1}}\newline \rotatetinytext{(14.98\%)}} &
  \multicolumn{2}{c}{\rotatebigtext{parking} \newline \textcolor{white}{\rotatetinytext{1}}\newline \rotatetinytext{(2.31\%)}} &
  \multicolumn{2}{c}{\rotatebigtext{sidewalk} \newline \textcolor{white}{\rotatetinytext{1}}\newline \rotatetinytext{(6.43\%)}} &
  \multicolumn{2}{c}{\rotatebigtext{other-grnd.} \newline \textcolor{white}{\rotatetinytext{1}}\newline \rotatetinytext{(2.05\%)}} &
  \multicolumn{2}{c}{\rotatebigtext{building} \newline \textcolor{white}{\rotatetinytext{1}}\newline \rotatetinytext{(15.67\%)}} &
  \multicolumn{2}{c}{\rotatebigtext{fence} \newline \textcolor{white}{\rotatetinytext{1}}\newline \rotatetinytext{(0.96\%)}} &
  \multicolumn{2}{c}{\rotatebigtext{vegetation} \newline \textcolor{white}{\rotatetinytext{1}}\newline \rotatetinytext{(41.99\%)}} &
  \multicolumn{2}{c}{\rotatebigtext{terrain} \newline \textcolor{white}{\rotatetinytext{1}}\newline \rotatetinytext{(7.10\%)}} &
  \multicolumn{2}{c}{\rotatebigtext{pole} \newline \textcolor{white}{\rotatetinytext{1}}\newline \rotatetinytext{(0.22\%)}} &
  \multicolumn{2}{c}{\rotatebigtext{traf.-sign} \newline \textcolor{white}{\rotatetinytext{1}}\newline \rotatetinytext{(0.06\%)}} &
  \multicolumn{2}{c}{\rotatebigtext{other-struct.} \newline \textcolor{white}{\rotatetinytext{1}}\newline \rotatetinytext{(4.33\%)}} &
  \multicolumn{2}{c|}{\rotatebigtext{other-obj.} \newline \textcolor{white}{\rotatetinytext{1}}\newline \rotatetinytext{(0.28\%)}} &
  \multicolumn{2}{c}{\textbf{OOV (val.)}}
   \\
  \textbf{Method} &
  \textbf{Input} &
  \textbf{IoU} &
  \textbf{mIoU} &
  \multicolumn{2}{c}{\colorboxlabel{custom_car}{}} &
  \multicolumn{2}{c}{\colorboxlabel{custom_bicycle}{}} &
  \multicolumn{2}{c}{\colorboxlabel{custom_motorcycle}{}} &
  \multicolumn{2}{c}{\colorboxlabel{custom_truck}{}} &
  \multicolumn{2}{c}{\colorboxlabel{custom_other_veh}{}} &
  \multicolumn{2}{c}{\colorboxlabel{custom_person}{}} &
  \multicolumn{2}{c}{\colorboxlabel{custom_road}{}} &
  \multicolumn{2}{c}{\colorboxlabel{custom_parking}{}} &
  \multicolumn{2}{c}{\colorboxlabel{custom_sidewalk}{}} &
  \multicolumn{2}{c}{\colorboxlabel{custom_other_ground}{}} &
  \multicolumn{2}{c}{\colorboxlabel{custom_building}{}} &
  \multicolumn{2}{c}{\colorboxlabel{custom_fence}{}} &
  \multicolumn{2}{c}{\colorboxlabel{custom_vegetation}{}} &
  \multicolumn{2}{c}{\colorboxlabel{custom_terrain}{}} &
  \multicolumn{2}{c}{\colorboxlabel{custom_pole}{}} &
  \multicolumn{2}{c}{\colorboxlabel{custom_traf_sign}{}} &
  \multicolumn{2}{c}{\colorboxlabel{custom_other_struct}{}} &
  \multicolumn{2}{c|}{\colorboxlabel{custom_other_obj}{}} &
  \textbf{IoU} &
  \textbf{mIoU} \\ \midrule
    \multicolumn{42}{c}{\textbf{\textit{Single-Frame-Based}}} \\ \midrule

\multicolumn{1}{l|}{Symphonies} &
  \multicolumn{1}{c|}{Stereo} &
   43.27&
   17.38&
  \multicolumn{2}{c}{29.84} &
  \multicolumn{2}{c}{1.24} &
  \multicolumn{2}{c}{4.36} &
  \multicolumn{2}{c}{\underline{23.48}} &
  \multicolumn{2}{c}{12.05} &
  \multicolumn{2}{c}{7.27} &
  \multicolumn{2}{c}{55.41} &
  \multicolumn{2}{c}{13.47} &
  \multicolumn{2}{c}{33.24} &
  \multicolumn{2}{c}{7.34} &
  \multicolumn{2}{c}{34.76} &
  \multicolumn{2}{c}{8.39} &
  \multicolumn{2}{c}{37.63} &
  \multicolumn{2}{c}{11.42} &
  \multicolumn{2}{c}{14.30} &
  \multicolumn{2}{c}{11.61} &
  \multicolumn{2}{c}{13.46} &
  \multicolumn{2}{c|}{10.92} &
   35.45 &
   12.46 \\
\multicolumn{1}{l|}{CGFormer} &
  \multicolumn{1}{c|}{Stereo} &
  48.28 &
  19.42 &
  \multicolumn{2}{c}{33.42} &
  \multicolumn{2}{c}{0.61} &
  \multicolumn{2}{c}{\underline{5.16}} &
  \multicolumn{2}{c}{22.60} &
  \multicolumn{2}{c}{10.52} &
  \multicolumn{2}{c}{\underline{10.36}} &
  \multicolumn{2}{c}{\underline{64.10}} &
  \multicolumn{2}{c}{16.62} &
  \multicolumn{2}{c}{\underline{38.64}} &
  \multicolumn{2}{c}{\underline{8.21}} &
  \multicolumn{2}{c}{37.85} &
  \multicolumn{2}{c}{\underline{9.43}} &
  \multicolumn{2}{c}{40.98} &
  \multicolumn{2}{c}{\underline{13.41}} &
  \multicolumn{2}{c}{16.13} &
  \multicolumn{2}{c}{12.10} &
  \multicolumn{2}{c}{15.52} &
  \multicolumn{2}{c|}{13.38} &
   44.81 &
   15.24 \\
\multicolumn{1}{l|}{ScanSSC} &
  \multicolumn{1}{c|}{Stereo} &
  \underline{48.78} &
  \underline{19.67} &
  \multicolumn{2}{c}{\underline{33.88}} &
  \multicolumn{2}{c}{\underline{7.83}} &
  \multicolumn{2}{c}{4.85} &
  \multicolumn{2}{c}{22.31} &
  \multicolumn{2}{c}{\underline{13.16}} &
  \multicolumn{2}{c}{9.65} &
  \multicolumn{2}{c}{62.67} &
  \multicolumn{2}{c}{\underline{17.88}} &
  \multicolumn{2}{c}{38.56} &
  \multicolumn{2}{c}{\textbf{8.40}} &
  \multicolumn{2}{c}{\underline{38.28}} &
  \multicolumn{2}{c}{9.14} &
  \multicolumn{2}{c}{\underline{41.69}} &
  \multicolumn{2}{c}{{13.34}} &
  \multicolumn{2}{c}{\underline{16.62}} &
  \multicolumn{2}{c}{\underline{13.17}} &
  \multicolumn{2}{c}{\underline{15.62}} &
  \multicolumn{2}{c|}{\underline{13.67}} &
    \underline{45.31}&
    \underline{15.28}\\ \midrule
    \multicolumn{42}{c}{\textbf{\textit{Temporal-Frame-Based}}} \\ \midrule
\rowcolor[HTML]{EFEFEF} 
\multicolumn{1}{l|}{\cellcolor[HTML]{EFEFEF}\textbf{Ours}} &
  \multicolumn{1}{c|}{\cellcolor[HTML]{EFEFEF}\textbf{Stereo}} &
  \textbf{50.24} &
   \textbf{22.65}&
  \multicolumn{2}{c}{\cellcolor[HTML]{EFEFEF} \textbf{36.24}} & 
  \multicolumn{2}{c}{\cellcolor[HTML]{EFEFEF} \textbf{0.82}} &
  \multicolumn{2}{c}{\cellcolor[HTML]{EFEFEF} \textbf{6.08}} &
  \multicolumn{2}{c}{\cellcolor[HTML]{EFEFEF} \textbf{26.47}} &
  \multicolumn{2}{c}{\cellcolor[HTML]{EFEFEF}\textbf{14.95} } &
  \multicolumn{2}{c}{\cellcolor[HTML]{EFEFEF}\textbf{10.89} } &
  \multicolumn{2}{c}{\cellcolor[HTML]{EFEFEF} \textbf{64.84}} &
  \multicolumn{2}{c}{\cellcolor[HTML]{EFEFEF} \textbf{19.67}} &
  \multicolumn{2}{c}{\cellcolor[HTML]{EFEFEF} \textbf{39.46}} &
  \multicolumn{2}{c}{\cellcolor[HTML]{EFEFEF}{7.97} } &
  \multicolumn{2}{c}{\cellcolor[HTML]{EFEFEF} \textbf{41.03}} &
  \multicolumn{2}{c}{\cellcolor[HTML]{EFEFEF} \textbf{10.27}} &
  \multicolumn{2}{c}{\cellcolor[HTML]{EFEFEF} \textbf{43.55}} &
  \multicolumn{2}{c}{\cellcolor[HTML]{EFEFEF} \textbf{15.24}} &
  \multicolumn{2}{c}{\cellcolor[HTML]{EFEFEF} \textbf{19.98}} &
  \multicolumn{2}{c}{\cellcolor[HTML]{EFEFEF} \textbf{16.49}} &
  \multicolumn{2}{c}{\cellcolor[HTML]{EFEFEF} \textbf{18.56}} &
  \multicolumn{2}{c|}{\cellcolor[HTML]{EFEFEF} \textbf{15.19}} &
   \textbf{54.23} &
   \textbf{24.54} \\ \bottomrule
\end{tabular}
}

\caption{Quantitative results on SSCBench-KITTI-360 validation set. ‘OOV’ on the right indicates performance on out-of-view regions, evaluated on the validation set using our implementation.
\textbf{Bold} / \underline{underline} highlight the best / second-best, respectively.}
\label{tab:KITTI360_val}
    
\end{table*}

%% file: Sections/Table/sup_results_SemanticKITTI_oov_valid.tex

\definecolor{custom_road}{RGB}{255, 0, 255}
\definecolor{custom_sidewalk}{RGB}{75, 0, 75}
\definecolor{custom_parking}{RGB}{255, 150, 255}
\definecolor{custom_other_ground}{RGB}{175, 0, 75}
\definecolor{custom_building}{RGB}{255, 200, 0}
\definecolor{custom_car}{RGB}{100, 150, 245}
\definecolor{custom_truck}{RGB}{80, 30, 180}
\definecolor{custom_bicycle}{RGB}{100, 230, 245}
\definecolor{custom_motorcycle}{RGB}{30, 60, 150}
\definecolor{custom_other_veh}{RGB}{0, 0, 255}
\definecolor{custom_vegetation}{RGB}{0, 175, 0}
\definecolor{custom_trunk}{RGB}{135, 60, 0}
\definecolor{custom_terrain}{RGB}{150, 240, 80}
\definecolor{custom_person}{RGB}{255, 30, 30}
\definecolor{custom_bicyclist}{RGB}{255, 40, 200}
\definecolor{custom_motorcyclist}{RGB}{150, 30, 90}
\definecolor{custom_fence}{RGB}{255, 120, 50}
\definecolor{custom_pole}{RGB}{255, 240, 150}
\definecolor{custom_traf_sign}{RGB}{255, 0, 0}

\begin{table*}[hbt!]
\centering
\resizebox{\textwidth}{!}{%
\begin{tabular}{lccc|clclclclclclclclclclclclclclclclclclcl}
\toprule
 &
   &
  \multicolumn{1}{l}{} &
  \multicolumn{1}{l|}{} &
  \multicolumn{2}{l}{} &
  \multicolumn{2}{l}{} &
  \multicolumn{2}{l}{} &
  \multicolumn{2}{l}{} &
  \multicolumn{2}{l}{} &
  \multicolumn{2}{l}{} &
  \multicolumn{2}{l}{} &
  \multicolumn{2}{l}{} &
  \multicolumn{2}{l}{} &
  \multicolumn{2}{l}{} &
  \multicolumn{2}{l}{} &
  \multicolumn{2}{l}{} &
  \multicolumn{2}{l}{} &
  \multicolumn{2}{l}{} &
  \multicolumn{2}{l}{} &
  \multicolumn{2}{l}{} &
  \multicolumn{2}{l}{} &
  \multicolumn{2}{l}{} &
  \multicolumn{2}{l}{} \\
 &
   &
  \multicolumn{1}{l}{} &
  \multicolumn{1}{l|}{} &
  \multicolumn{2}{l}{\rotatebigtext{road} \newline \textcolor{white}{\rotatetinytext{1}} \newline \rotatetinytext{(15.30\%)}} &
  \multicolumn{2}{l}{\rotatebigtext{sidewalk} \newline \textcolor{white}{\rotatetinytext{1}} \newline \rotatetinytext{(11.13\%)}} &
  \multicolumn{2}{l}{\rotatebigtext{parking} \newline \textcolor{white}{\rotatetinytext{1}}\newline \rotatetinytext{(1.12\%)}} &
  \multicolumn{2}{l}{\rotatebigtext{other-grnd.} \newline \textcolor{white}{\rotatetinytext{1}}\newline \rotatetinytext{(0.56\%)}} &
  \multicolumn{2}{l}{\rotatebigtext{building} \newline \textcolor{white}{\rotatetinytext{1}}\newline \rotatetinytext{(14.1\%)}} &
  \multicolumn{2}{l}{\rotatebigtext{car} \newline \textcolor{white}{\rotatetinytext{1}}\newline \rotatetinytext{(3.92\%)}} &
  \multicolumn{2}{l}{\rotatebigtext{truck} \newline \textcolor{white}{\rotatetinytext{1}}\newline \rotatetinytext{(0.16\%)}} &
  \multicolumn{2}{l}{\rotatebigtext{bicycle} \newline \textcolor{white}{\rotatetinytext{1}}\newline \rotatetinytext{(0.03\%)}} &
  \multicolumn{2}{l}{\rotatebigtext{motorcycle} \newline \textcolor{white}{\rotatetinytext{1}}\newline \rotatetinytext{(0.03\%)}} &
  \multicolumn{2}{l}{\rotatebigtext{other-veh.} \newline \textcolor{white}{\rotatetinytext{1}}\newline \rotatetinytext{(0.20\%)}} &
  \multicolumn{2}{l}{\rotatebigtext{vegetation} \newline \textcolor{white}{\rotatetinytext{1}}\newline \rotatetinytext{(39.3\%)}} &
  \multicolumn{2}{l}{\rotatebigtext{trunk} \newline \textcolor{white}{\rotatetinytext{1}}\newline \rotatetinytext{(0.51\%)}} &
  \multicolumn{2}{l}{\rotatebigtext{terrain} \newline \textcolor{white}{\rotatetinytext{1}}\newline \rotatetinytext{(9.17\%)}} &
  \multicolumn{2}{l}{\rotatebigtext{person} \newline \textcolor{white}{\rotatetinytext{1}}\newline \rotatetinytext{(0.07\%)}} &
  \multicolumn{2}{l}{\rotatebigtext{bicyclist} \newline \textcolor{white}{\rotatetinytext{1}}\newline \rotatetinytext{(0.07\%)}} &
  \multicolumn{2}{l}{\rotatebigtext{motorcyclist} \newline \textcolor{white}{\rotatetinytext{1}}\newline \rotatetinytext{(0.05\%)}} &
  \multicolumn{2}{l}{\rotatebigtext{fence} \newline \textcolor{white}{\rotatetinytext{1}}\newline \rotatetinytext{(3.90\%)}} &
  \multicolumn{2}{l}{\rotatebigtext{pole} \newline \textcolor{white}{\rotatetinytext{1}}\newline \rotatetinytext{(0.29\%)}} &
  \multicolumn{2}{l}{\rotatebigtext{traf.-sign} \newline \textcolor{white}{\rotatetinytext{1}}\newline \rotatetinytext{(0.08\%)}} \\
\textbf{Method} &
  \textbf{Input} &
  \textbf{IoU} &
  \textbf{mIoU} &
  \multicolumn{2}{c}{\colorboxlabel{custom_road}{}} &
  \multicolumn{2}{c}{\colorboxlabel{custom_sidewalk}{}} &
  \multicolumn{2}{c}{\colorboxlabel{custom_parking}{}} &
  \multicolumn{2}{c}{\colorboxlabel{custom_other_ground}{}} &
  \multicolumn{2}{c}{\colorboxlabel{custom_building}{}} &
  \multicolumn{2}{c}{\colorboxlabel{custom_car}{}} &
  \multicolumn{2}{c}{\colorboxlabel{custom_truck}{}} &
  \multicolumn{2}{c}{\colorboxlabel{custom_bicycle}{}} &
  \multicolumn{2}{c}{\colorboxlabel{custom_motorcycle}{}} &
  \multicolumn{2}{c}{\colorboxlabel{custom_other_veh}{}} &
  \multicolumn{2}{c}{\colorboxlabel{custom_vegetation}{}} &
  \multicolumn{2}{c}{\colorboxlabel{custom_trunk}{}} &
  \multicolumn{2}{c}{\colorboxlabel{custom_terrain}{}} &
  \multicolumn{2}{c}{\colorboxlabel{custom_person}{}} &
  \multicolumn{2}{c}{\colorboxlabel{custom_bicyclist}{}} &
  \multicolumn{2}{c}{\colorboxlabel{custom_motorcyclist}{}} &
  \multicolumn{2}{c}{\colorboxlabel{custom_fence}{}} &
  \multicolumn{2}{c}{\colorboxlabel{custom_pole}{}} &
  \multicolumn{2}{c}{\colorboxlabel{custom_traf_sign}{}} \\ \midrule
   \multicolumn{42}{c}{\textbf{\textit{Single-Frame-Based}}} \\ \midrule
\multicolumn{1}{l|}{MonoScene} &
  \multicolumn{1}{c|}{Mono} &
   31.07&
   7.02&
  \multicolumn{2}{c}{49.23} &
  \multicolumn{2}{c}{19.29} &
  \multicolumn{2}{c}{5.57} &
  \multicolumn{2}{c}{\underline{1.55}} &
  \multicolumn{2}{c}{10.00} &
  \multicolumn{2}{c}{5.36} &
  \multicolumn{2}{c}{0.24} &
  \multicolumn{2}{c}{0.01} &
  \multicolumn{2}{c}{0.03} &
  \multicolumn{2}{c}{0.57} &
  \multicolumn{2}{c}{12.17} &
  \multicolumn{2}{c}{0.03} &
  \multicolumn{2}{c}{26.87} &
  \multicolumn{2}{c}{0.07} &
  \multicolumn{2}{c}{0.00} &
  \multicolumn{2}{c}{0.00} &
  \multicolumn{2}{c}{2.03} &
  \multicolumn{2}{c}{0.09} &
  \multicolumn{2}{c}{0.29}\\
\multicolumn{1}{l|}{TPVFormer} &
  \multicolumn{1}{c|}{Mono} &
   30.08&
   7.87&
  \multicolumn{2}{c}{48.92} &
  \multicolumn{2}{c}{18.58} &
  \multicolumn{2}{c}{11.50} &
  \multicolumn{2}{c}{1.60} &
  \multicolumn{2}{c}{10.77} &
  \multicolumn{2}{c}{7.93} &
  \multicolumn{2}{c}{6.62} &
  \multicolumn{2}{c}{0.00} &
  \multicolumn{2}{c}{0.03} &
  \multicolumn{2}{c}{0.70} &
  \multicolumn{2}{c}{12.20} &
  \multicolumn{2}{c}{0.08} &
  \multicolumn{2}{c}{28.28} &
  \multicolumn{2}{c}{0.00} &
  \multicolumn{2}{c}{0.00} &
  \multicolumn{2}{c}{0.00} &
  \multicolumn{2}{c}{2.08} &
  \multicolumn{2}{c}{0.07} &
  \multicolumn{2}{c}{0.09}\\
\multicolumn{1}{l|}{OccFormer} &
  \multicolumn{1}{c|}{Mono} &
  36.42 &
   13.50&
  \multicolumn{2}{c}{59.81} &
  \multicolumn{2}{c}{\underline{27.84}} &
  \multicolumn{2}{c}{\textbf{21.75}} &
  \multicolumn{2}{c}{0.69} &
  \multicolumn{2}{c}{14.54} &
  \multicolumn{2}{c}{\underline{25.19}} &
  \multicolumn{2}{c}{\underline{15.39}} &
  \multicolumn{2}{c}{\underline{1.91}} &
  \multicolumn{2}{c}{2.03} &
  \multicolumn{2}{c}{\textbf{11.61}} &
  \multicolumn{2}{c}{19.39} &
  \multicolumn{2}{c}{3.64} &
  \multicolumn{2}{c}{31.80} &
  \multicolumn{2}{c}{\textbf{3.03}} &
  \multicolumn{2}{c}{\textbf{5.35}} &
  \multicolumn{2}{c}{0.00} &
  \multicolumn{2}{c}{\underline{5.66}} &
  \multicolumn{2}{c}{4.13} &
  \multicolumn{2}{c}{2.76}\\
\multicolumn{1}{l|}{Symphonies} &
  \multicolumn{1}{c|}{Stereo} &
   23.48&
   6.40&
  \multicolumn{2}{c}{47.63} &
  \multicolumn{2}{c}{20.03} &
  \multicolumn{2}{c}{5.77} &
  \multicolumn{2}{c}{1.42} &
  \multicolumn{2}{c}{6.02} &
  \multicolumn{2}{c}{7.80} &
  \multicolumn{2}{c}{0.62} &
  \multicolumn{2}{c}{0.05} &
  \multicolumn{2}{c}{0.19} &
  \multicolumn{2}{c}{1.50} &
  \multicolumn{2}{c}{10.10} &
  \multicolumn{2}{c}{0.10} &
  \multicolumn{2}{c}{18.36} &
  \multicolumn{2}{c}{0.02} &
  \multicolumn{2}{c}{0.17} &
  \multicolumn{2}{c}{{0.00}} &
  \multicolumn{2}{c}{1.42} &
  \multicolumn{2}{c}{0.15} &
  \multicolumn{2}{c}{0.20}\\
\multicolumn{1}{l|}{CGFormer} &
  \multicolumn{1}{c|}{Stereo} &
   33.54&
   9.06&
  \multicolumn{2}{c}{56.30} &
  \multicolumn{2}{c}{23.34} &
  \multicolumn{2}{c}{8.22} &
  \multicolumn{2}{c}{0.23} &
  \multicolumn{2}{c}{12.95} &
  \multicolumn{2}{c}{9.90} &
  \multicolumn{2}{c}{7.92} &
  \multicolumn{2}{c}{0.11} &
  \multicolumn{2}{c}{0.10} &
  \multicolumn{2}{c}{2.17} &
  \multicolumn{2}{c}{13.83} &
  \multicolumn{2}{c}{0.11} &
  \multicolumn{2}{c}{32.25} &
  \multicolumn{2}{c}{0.04} &
  \multicolumn{2}{c}{\underline{0.83}} &
  \multicolumn{2}{c}{0.00} &
  \multicolumn{2}{c}{3.34} &
  \multicolumn{2}{c}{0.13} &
  \multicolumn{2}{c}{0.46}\\
\multicolumn{1}{l|}{L2COcc-C} &
  \multicolumn{1}{c|}{Stereo} &
   32.24&
   8.55&
  \multicolumn{2}{c}{53.63} &
  \multicolumn{2}{c}{23.30} &
  \multicolumn{2}{c}{6.20} &
  \multicolumn{2}{c}{0.28} &
  \multicolumn{2}{c}{13.43} &
  \multicolumn{2}{c}{8.45} &
  \multicolumn{2}{c}{4.19} &
  \multicolumn{2}{c}{0.64} &
  \multicolumn{2}{c}{1.72} &
  \multicolumn{2}{c}{1.74} &
  \multicolumn{2}{c}{12.38} &
  \multicolumn{2}{c}{0.13} &
  \multicolumn{2}{c}{32.82} &
  \multicolumn{2}{c}{0.19} &
  \multicolumn{2}{c}{0.03} &
  \multicolumn{2}{c}{0.00} &
  \multicolumn{2}{c}{2.88} &
  \multicolumn{2}{c}{0.12} &
  \multicolumn{2}{c}{0.28}\\
\multicolumn{1}{l|}{ScanSSC} &
  \multicolumn{1}{c|}{Stereo} &
   33.60&
   9.50&
  \multicolumn{2}{c}{59.07} &
  \multicolumn{2}{c}{25.20} &
  \multicolumn{2}{c}{12.19} &
  \multicolumn{2}{c}{0.45} &
  \multicolumn{2}{c}{12.63} &
  \multicolumn{2}{c}{9.62} &
  \multicolumn{2}{c}{8.11} &
  \multicolumn{2}{c}{0.06} &
  \multicolumn{2}{c}{0.09} &
  \multicolumn{2}{c}{3.31} &
  \multicolumn{2}{c}{13.74} &
  \multicolumn{2}{c}{0.09} &
  \multicolumn{2}{c}{31.90} &
  \multicolumn{2}{c}{0.00} &
  \multicolumn{2}{c}{0.05} &
  \multicolumn{2}{c}{0.00} &
  \multicolumn{2}{c}{3.62} &
  \multicolumn{2}{c}{0.13} &
  \multicolumn{2}{c}{0.32}\\
\multicolumn{1}{l|}{L2COcc-D} &
  \multicolumn{1}{c|}{Stereo} &
   31.85&
   10.05&
  \multicolumn{2}{c}{\underline{59.29}} &
  \multicolumn{2}{c}{26.24} &
  \multicolumn{2}{c}{12.35} &
  \multicolumn{2}{c}{1.60} &
  \multicolumn{2}{c}{13.33} &
  \multicolumn{2}{c}{12.15} &
  \multicolumn{2}{c}{6.07} &
  \multicolumn{2}{c}{0.35} &
  \multicolumn{2}{c}{\underline{2.39}} &
  \multicolumn{2}{c}{3.77} &
  \multicolumn{2}{c}{14.14} &
  \multicolumn{2}{c}{0.21} &
  \multicolumn{2}{c}{\underline{34.55}} &
  \multicolumn{2}{c}{0.60} &
  \multicolumn{2}{c}{0.08} &
  \multicolumn{2}{c}{0.00} &
  \multicolumn{2}{c}{3.18} &
  \multicolumn{2}{c}{0.29} &
  \multicolumn{2}{c}{0.45} \\ \midrule
  \multicolumn{42}{c}{\textbf{\textit{Temporal-Frame-Based}}} \\ \midrule
\multicolumn{1}{l|}{VoxFormer-T} &
  \multicolumn{1}{c|}{Stereo} &
   \underline{40.21}&
   \underline{11.58}&
  \multicolumn{2}{c}{50.45} &
  \multicolumn{2}{c}{24.77} &
  \multicolumn{2}{c}{11.66} &
  \multicolumn{2}{c}{\textbf{1.83}} &
  \multicolumn{2}{c}{\underline{19.78}} &
  \multicolumn{2}{c}{21.83} &
  \multicolumn{2}{c}{2.33} &
  \multicolumn{2}{c}{0.77} &
  \multicolumn{2}{c}{1.42} &
  \multicolumn{2}{c}{3.31} &
  \multicolumn{2}{c}{\underline{22.44}} &
  \multicolumn{2}{c}{\underline{5.26}} &
  \multicolumn{2}{c}{34.13} &
  \multicolumn{2}{c}{0.90} &
  \multicolumn{2}{c}{0.01} &
  \multicolumn{2}{c}{0.00} &
  \multicolumn{2}{c}{4.90} &
  \multicolumn{2}{c}{\underline{9,29}} &
  \multicolumn{2}{c}{\underline{4.92}}  \\
\multicolumn{1}{l|}{HTCL-S} &
  \multicolumn{1}{c|}{Stereo} &
   33.14&
   9.04&
  \multicolumn{2}{c}{54.93} &
  \multicolumn{2}{c}{23.71} &
  \multicolumn{2}{c}{11.21} &
  \multicolumn{2}{c}{0.61} &
  \multicolumn{2}{c}{12.86} &
  \multicolumn{2}{c}{10.69} &
  \multicolumn{2}{c}{5.24} &
  \multicolumn{2}{c}{0.19} &
  \multicolumn{2}{c}{0.10} &
  \multicolumn{2}{c}{4.91} &
  \multicolumn{2}{c}{13.53} &
  \multicolumn{2}{c}{0.08} &
  \multicolumn{2}{c}{30.50} &
  \multicolumn{2}{c}{0.00} &
  \multicolumn{2}{c}{0.06} &
  \multicolumn{2}{c}{0.00} &
  \multicolumn{2}{c}{2.68} &
  \multicolumn{2}{c}{0.19} &
  \multicolumn{2}{c}{0.30} \\
\multicolumn{1}{l|}{Hi-SOP} &
  \multicolumn{1}{c|}{Stereo} &
   -&
   -&
  \multicolumn{2}{c}{-} &
  \multicolumn{2}{c}{-} &
  \multicolumn{2}{c}{-} &
  \multicolumn{2}{c}{-} &
  \multicolumn{2}{c}{-} &
  \multicolumn{2}{c}{-} &
  \multicolumn{2}{c}{-} &
  \multicolumn{2}{c}{-} &
  \multicolumn{2}{c}{-} &
  \multicolumn{2}{c}{-} &
  \multicolumn{2}{c}{-} &
  \multicolumn{2}{c}{-} &
  \multicolumn{2}{c}{-} &
  \multicolumn{2}{c}{-} &
  \multicolumn{2}{c}{-} &
  \multicolumn{2}{c}{-} &
  \multicolumn{2}{c}{-} &
  \multicolumn{2}{c}{-} &
  \multicolumn{2}{c}{-} \\
\multicolumn{1}{l|}{FlowScene} &
  \multicolumn{1}{c|}{Stereo} &
   -&
   -&
  \multicolumn{2}{c}{-} &
  \multicolumn{2}{c}{-} &
  \multicolumn{2}{c}{-} &
  \multicolumn{2}{c}{-} &
  \multicolumn{2}{c}{-} &
  \multicolumn{2}{c}{-} &
  \multicolumn{2}{c}{-} &
  \multicolumn{2}{c}{-} &
  \multicolumn{2}{c}{-} &
  \multicolumn{2}{c}{-} &
  \multicolumn{2}{c}{-} &
  \multicolumn{2}{c}{-} &
  \multicolumn{2}{c}{-} &
  \multicolumn{2}{c}{-} &
  \multicolumn{2}{c}{-} &
  \multicolumn{2}{c}{-} &
  \multicolumn{2}{c}{-} &
  \multicolumn{2}{c}{-} &
  \multicolumn{2}{c}{-} \\
\multicolumn{1}{l|}{CF-SSC} &
  \multicolumn{1}{c|}{Stereo} &
  - &
  - &
  \multicolumn{2}{c}{-} &
  \multicolumn{2}{c}{-} &
  \multicolumn{2}{c}{-} &
  \multicolumn{2}{c}{-} &
  \multicolumn{2}{c}{{-}} &
  \multicolumn{2}{c}{-} &
  \multicolumn{2}{c}{-} &
  \multicolumn{2}{c}{-} &
  \multicolumn{2}{c}{-} &
  \multicolumn{2}{c}{-} &
  \multicolumn{2}{c}{{-}} &
  \multicolumn{2}{c}{-} &
  \multicolumn{2}{c}{-} &
  \multicolumn{2}{c}{-} &
  \multicolumn{2}{c}{-} &
  \multicolumn{2}{c}{-} &
  \multicolumn{2}{c}{-} &
  \multicolumn{2}{c}{-} &
  \multicolumn{2}{c}{-} \\
\rowcolor[HTML]{EFEFEF} 
\multicolumn{1}{l|}{\cellcolor[HTML]{EFEFEF}\textbf{Ours}} &
  \multicolumn{1}{c|}{\cellcolor[HTML]{EFEFEF}\textbf{Stereo}} &
  \textbf{44.37} &
  \textbf{17.17} &
  \multicolumn{2}{c}{\cellcolor[HTML]{EFEFEF}\textbf{62.98}} &
  \multicolumn{2}{c}{\cellcolor[HTML]{EFEFEF}\textbf{30.22}} &
  \multicolumn{2}{c}{\cellcolor[HTML]{EFEFEF}\underline{17.36}} &
  \multicolumn{2}{c}{\cellcolor[HTML]{EFEFEF}{0.26}} &
  \multicolumn{2}{c}{\cellcolor[HTML]{EFEFEF}\textbf{29.23}} &
  \multicolumn{2}{c}{\cellcolor[HTML]{EFEFEF}\textbf{26.86}} &
  \multicolumn{2}{c}{\cellcolor[HTML]{EFEFEF}\textbf{18.05}} &
  \multicolumn{2}{c}{\cellcolor[HTML]{EFEFEF}\textbf{8.02}} &
  \multicolumn{2}{c}{\cellcolor[HTML]{EFEFEF}\textbf{7.71}} &
  \multicolumn{2}{c}{\cellcolor[HTML]{EFEFEF}\underline{11.11}} &
  \multicolumn{2}{c}{\cellcolor[HTML]{EFEFEF}\textbf{27.24}} &
  \multicolumn{2}{c}{\cellcolor[HTML]{EFEFEF}\textbf{11.95}} &
  \multicolumn{2}{c}{\cellcolor[HTML]{EFEFEF}\textbf{41.96}} &
  \multicolumn{2}{c}{\cellcolor[HTML]{EFEFEF}\underline{1.10}} &
  \multicolumn{2}{c}{\cellcolor[HTML]{EFEFEF}{0.14}} &
  \multicolumn{2}{c}{\cellcolor[HTML]{EFEFEF}{0.00}} &
  \multicolumn{2}{c}{\cellcolor[HTML]{EFEFEF}\textbf{7.13}} &
  \multicolumn{2}{c}{\cellcolor[HTML]{EFEFEF}\textbf{15.82}} &
  \multicolumn{2}{c}{\cellcolor[HTML]{EFEFEF}\textbf{9.07}}  \\ \bottomrule

\end{tabular}%
}
\caption{Quantitative results on out-of-view (OOV) regions in the SemanticKITTI validation set. As Hi-SOP~\cite{li2024hierarchical1}, FlowScene~\cite{wang2025learning}, and CF-SSC~\cite{lu2025one} do not provide pretrained weights, their results on OOV regions are marked as `–'. \textbf{Bold} / \underline{underline} highlight the best / second-best, respectively.}
\label{tab:SemKITTI_oov_val}
    
\end{table*}

%% file: Sections/Table/sup_results_KITTI360_oov_valid.tex

\definecolor{custom_road}{RGB}{255, 0, 255}
\definecolor{custom_sidewalk}{RGB}{75, 0, 75}
\definecolor{custom_parking}{RGB}{255, 150, 255}
\definecolor{custom_other_ground}{RGB}{175, 0, 75}
\definecolor{custom_building}{RGB}{255, 200, 0}
\definecolor{custom_car}{RGB}{100, 150, 245}
\definecolor{custom_truck}{RGB}{80, 30, 180}
\definecolor{custom_bicycle}{RGB}{100, 230, 245}
\definecolor{custom_motorcycle}{RGB}{30, 60, 150}
\definecolor{custom_other_veh}{RGB}{0, 0, 255}
\definecolor{custom_vegetation}{RGB}{0, 175, 0}
\definecolor{custom_trunk}{RGB}{135, 60, 0}
\definecolor{custom_terrain}{RGB}{150, 240, 80}
\definecolor{custom_person}{RGB}{255, 30, 30}
\definecolor{custom_bicyclist}{RGB}{255, 40, 20}
\definecolor{custom_motorcyclist}{RGB}{150, 30, 90}
\definecolor{custom_fence}{RGB}{255, 120, 50}
\definecolor{custom_pole}{RGB}{255, 240, 150}
\definecolor{custom_traf_sign}{RGB}{255, 0, 0}
\definecolor{custom_other_struct}{RGB}{255, 150, 0}
\definecolor{custom_other_obj}{RGB}{50, 255, 255}

\begin{table*}[hbt!]
\centering
\resizebox{\textwidth}{!}{%
\begin{tabular}{lccc|cccccccccccccccccccccccccccccccccccc}
\toprule
 &
   &
  \multicolumn{1}{l}{} &
   &
  \multicolumn{2}{c}{} &
  \multicolumn{2}{c}{} &
  \multicolumn{2}{c}{} &
  \multicolumn{2}{c}{} &
  \multicolumn{2}{c}{} &
  \multicolumn{2}{c}{} &
  \multicolumn{2}{c}{} &
  \multicolumn{2}{c}{} &
  \multicolumn{2}{c}{} &
  \multicolumn{2}{c}{} &
  \multicolumn{2}{c}{} &
  \multicolumn{2}{c}{} &
  \multicolumn{2}{c}{} &
  \multicolumn{2}{c}{} &
  \multicolumn{2}{c}{} &
  \multicolumn{2}{c}{} &
  \multicolumn{2}{c}{} &
  \multicolumn{2}{c}{} \\
 &
   &
   &
   &
  \multicolumn{2}{c}{\rotatebigtext{car} \newline \textcolor{white}{\rotatetinytext{1}} \newline \rotatetinytext{(2.85\%)}} &
  \multicolumn{2}{c}{\rotatebigtext{bicycle} \newline \textcolor{white}{\rotatetinytext{1}} \newline \rotatetinytext{(0.01\%)}} &
  \multicolumn{2}{c}{\rotatebigtext{motorcycle} \newline \textcolor{white}{\rotatetinytext{1}}\newline \rotatetinytext{(0.01\%)}} &
  \multicolumn{2}{c}{\rotatebigtext{truck} \newline \textcolor{white}{\rotatetinytext{1}}\newline \rotatetinytext{(0.16\%)}} &
  \multicolumn{2}{c}{\rotatebigtext{other-veh.} \newline \textcolor{white}{\rotatetinytext{1}}\newline \rotatetinytext{(5.75\%)}} &
  \multicolumn{2}{c}{\rotatebigtext{person} \newline \textcolor{white}{\rotatetinytext{1}}\newline \rotatetinytext{(0.02\%)}} &
  \multicolumn{2}{c}{\rotatebigtext{road} \newline \textcolor{white}{\rotatetinytext{1}}\newline \rotatetinytext{(14.98\%)}} &
  \multicolumn{2}{c}{\rotatebigtext{parking} \newline \textcolor{white}{\rotatetinytext{1}}\newline \rotatetinytext{(2.31\%)}} &
  \multicolumn{2}{c}{\rotatebigtext{sidewalk} \newline \textcolor{white}{\rotatetinytext{1}}\newline \rotatetinytext{(6.43\%)}} &
  \multicolumn{2}{c}{\rotatebigtext{other-grnd.} \newline \textcolor{white}{\rotatetinytext{1}}\newline \rotatetinytext{(2.05\%)}} &
  \multicolumn{2}{c}{\rotatebigtext{building} \newline \textcolor{white}{\rotatetinytext{1}}\newline \rotatetinytext{(15.67\%)}} &
  \multicolumn{2}{c}{\rotatebigtext{fence} \newline \textcolor{white}{\rotatetinytext{1}}\newline \rotatetinytext{(0.96\%)}} &
  \multicolumn{2}{c}{\rotatebigtext{vegetation} \newline \textcolor{white}{\rotatetinytext{1}}\newline \rotatetinytext{(41.99\%)}} &
  \multicolumn{2}{c}{\rotatebigtext{terrain} \newline \textcolor{white}{\rotatetinytext{1}}\newline \rotatetinytext{(7.10\%)}} &
  \multicolumn{2}{c}{\rotatebigtext{pole} \newline \textcolor{white}{\rotatetinytext{1}}\newline \rotatetinytext{(0.22\%)}} &
  \multicolumn{2}{c}{\rotatebigtext{traf.-sign} \newline \textcolor{white}{\rotatetinytext{1}}\newline \rotatetinytext{(0.06\%)}} &
  \multicolumn{2}{c}{\rotatebigtext{other-struct.} \newline \textcolor{white}{\rotatetinytext{1}}\newline \rotatetinytext{(4.33\%)}} &
  \multicolumn{2}{c}{\rotatebigtext{other-obj.} \newline \textcolor{white}{\rotatetinytext{1}}\newline \rotatetinytext{(0.28\%)}}
   \\
  \textbf{Method} &
  \textbf{Input} &
  \textbf{IoU} &
  \textbf{mIoU} &
  \multicolumn{2}{c}{\colorboxlabel{custom_car}{}} &
  \multicolumn{2}{c}{\colorboxlabel{custom_bicycle}{}} &
  \multicolumn{2}{c}{\colorboxlabel{custom_motorcycle}{}} &
  \multicolumn{2}{c}{\colorboxlabel{custom_truck}{}} &
  \multicolumn{2}{c}{\colorboxlabel{custom_other_veh}{}} &
  \multicolumn{2}{c}{\colorboxlabel{custom_person}{}} &
  \multicolumn{2}{c}{\colorboxlabel{custom_road}{}} &
  \multicolumn{2}{c}{\colorboxlabel{custom_parking}{}} &
  \multicolumn{2}{c}{\colorboxlabel{custom_sidewalk}{}} &
  \multicolumn{2}{c}{\colorboxlabel{custom_other_ground}{}} &
  \multicolumn{2}{c}{\colorboxlabel{custom_building}{}} &
  \multicolumn{2}{c}{\colorboxlabel{custom_fence}{}} &
  \multicolumn{2}{c}{\colorboxlabel{custom_vegetation}{}} &
  \multicolumn{2}{c}{\colorboxlabel{custom_terrain}{}} &
  \multicolumn{2}{c}{\colorboxlabel{custom_pole}{}} &
  \multicolumn{2}{c}{\colorboxlabel{custom_traf_sign}{}} &
  \multicolumn{2}{c}{\colorboxlabel{custom_other_struct}{}} &
  \multicolumn{2}{c}{\colorboxlabel{custom_other_obj}{}} \\ \midrule
    \multicolumn{40}{c}{\textbf{\textit{Single-Frame-Based}}} \\ \midrule

\multicolumn{1}{l|}{Symphonies} &
  \multicolumn{1}{c|}{Stereo} &
   35.45&
   12.46&
  \multicolumn{2}{c}{19.77} &
  \multicolumn{2}{c}{0.00} &
  \multicolumn{2}{c}{4.94} &
  \multicolumn{2}{c}{17.96} &
  \multicolumn{2}{c}{5.68} &
  \multicolumn{2}{c}{4.09} &
  \multicolumn{2}{c}{39.19} &
  \multicolumn{2}{c}{11.00} &
  \multicolumn{2}{c}{26.85} &
  \multicolumn{2}{c}{5.50} &
  \multicolumn{2}{c}{25.97} &
  \multicolumn{2}{c}{5.59} &
  \multicolumn{2}{c}{30.27} &
  \multicolumn{2}{c}{9.08} &
  \multicolumn{2}{c}{7.94} &
  \multicolumn{2}{c}{6.93} &
  \multicolumn{2}{c}{7.66} &
  \multicolumn{2}{c}{8.41} \\
\multicolumn{1}{l|}{CGFormer} &
  \multicolumn{1}{c|}{Stereo} &
   44.81&
   15.24&
  \multicolumn{2}{c}{\underline{24.68}} &
  \multicolumn{2}{c}{\underline{0.12}} &
  \multicolumn{2}{c}{\underline{6.35}} &
  \multicolumn{2}{c}{\underline{18.21}} &
  \multicolumn{2}{c}{6.95} &
  \multicolumn{2}{c}{\underline{5.24}} &
  \multicolumn{2}{c}{\underline{46.70}} &
  \multicolumn{2}{c}{15.38} &
  \multicolumn{2}{c}{33.11} &
  \multicolumn{2}{c}{7.29} &
  \multicolumn{2}{c}{33.74} &
  \multicolumn{2}{c}{6.61} &
  \multicolumn{2}{c}{37.82} &
  \multicolumn{2}{c}{\underline{10.98}} &
  \multicolumn{2}{c}{9.68} &
  \multicolumn{2}{c}{5.34} &
  \multicolumn{2}{c}{11.18} &
  \multicolumn{2}{c}{\underline{10.24}}\\
\multicolumn{1}{l|}{ScanSSC} &
  \multicolumn{1}{c|}{Stereo} &
   \underline{45.31}&
   \underline{15.28}&
  \multicolumn{2}{c}{24.44} &
  \multicolumn{2}{c}{\underline{0.12}} &
  \multicolumn{2}{c}{5.46} &
  \multicolumn{2}{c}{16.35} &
  \multicolumn{2}{c}{\underline{8.37}} &
  \multicolumn{2}{c}{3.68} &
  \multicolumn{2}{c}{45.31} &
  \multicolumn{2}{c}{\underline{16.93}} &
  \multicolumn{2}{c}{\underline{32.68}} &
  \multicolumn{2}{c}{\underline{7.36}} &
  \multicolumn{2}{c}{\underline{34.26}} &
  \multicolumn{2}{c}{\underline{7.14}} &
  \multicolumn{2}{c}{\underline{38.34}} &
  \multicolumn{2}{c}{10.85} &
  \multicolumn{2}{c}{\underline{9.99}} &
  \multicolumn{2}{c}{\underline{7.27}} &
  \multicolumn{2}{c}{\underline{11.61}} &
  \multicolumn{2}{c}{10.18} \\ \midrule
    \multicolumn{40}{c}{\textbf{\textit{Temporal-Frame-Based}}} \\ \midrule
\rowcolor[HTML]{EFEFEF} 
\multicolumn{1}{l|}{\cellcolor[HTML]{EFEFEF}\textbf{Ours}} &
  \multicolumn{1}{c|}{\cellcolor[HTML]{EFEFEF}\textbf{Stereo}} &
   \textbf{54.23}&
   \textbf{24.54}&
  \multicolumn{2}{c}{\cellcolor[HTML]{EFEFEF} \textbf{42.50}} & 
  \multicolumn{2}{c}{\cellcolor[HTML]{EFEFEF} \textbf{1.87}} &
  \multicolumn{2}{c}{\cellcolor[HTML]{EFEFEF} \textbf{10.72}} &
  \multicolumn{2}{c}{\cellcolor[HTML]{EFEFEF} \textbf{27.59}} &
  \multicolumn{2}{c}{\cellcolor[HTML]{EFEFEF} \textbf{18.94}} &
  \multicolumn{2}{c}{\cellcolor[HTML]{EFEFEF} \textbf{10.97}} &
  \multicolumn{2}{c}{\cellcolor[HTML]{EFEFEF} \textbf{55.20}} &
  \multicolumn{2}{c}{\cellcolor[HTML]{EFEFEF} \textbf{25.15}} &
  \multicolumn{2}{c}{\cellcolor[HTML]{EFEFEF} \textbf{39.99}} &
  \multicolumn{2}{c}{\cellcolor[HTML]{EFEFEF} \textbf{10.14}} &
  \multicolumn{2}{c}{\cellcolor[HTML]{EFEFEF} \textbf{44.71}} &
  \multicolumn{2}{c}{\cellcolor[HTML]{EFEFEF} \textbf{14.92}} &
  \multicolumn{2}{c}{\cellcolor[HTML]{EFEFEF} \textbf{48.64}} &
  \multicolumn{2}{c}{\cellcolor[HTML]{EFEFEF} \textbf{20.60}} &
  \multicolumn{2}{c}{\cellcolor[HTML]{EFEFEF} \textbf{27.24}} &
  \multicolumn{2}{c}{\cellcolor[HTML]{EFEFEF} \textbf{22.60}} &
  \multicolumn{2}{c}{\cellcolor[HTML]{EFEFEF} \textbf{22.92}} &
  \multicolumn{2}{c}{\cellcolor[HTML]{EFEFEF} \textbf{21.60}} \\ \bottomrule
\end{tabular}
}

\caption{Quantitative results on out-of-view regions in the SSCBench-KITTI-360 validation set. \textbf{Bold} / \underline{underline} highlight the best / second-best, respectively.}
\label{tab:KITTI360_oov_val}
    
\end{table*}

%% file: Sections/Table/sup_results_KITTI360_oov_test.tex

\definecolor{custom_road}{RGB}{255, 0, 255}
\definecolor{custom_sidewalk}{RGB}{75, 0, 75}
\definecolor{custom_parking}{RGB}{255, 150, 255}
\definecolor{custom_other_ground}{RGB}{175, 0, 75}
\definecolor{custom_building}{RGB}{255, 200, 0}
\definecolor{custom_car}{RGB}{100, 150, 245}
\definecolor{custom_truck}{RGB}{80, 30, 180}
\definecolor{custom_bicycle}{RGB}{100, 230, 245}
\definecolor{custom_motorcycle}{RGB}{30, 60, 150}
\definecolor{custom_other_veh}{RGB}{0, 0, 255}
\definecolor{custom_vegetation}{RGB}{0, 175, 0}
\definecolor{custom_trunk}{RGB}{135, 60, 0}
\definecolor{custom_terrain}{RGB}{150, 240, 80}
\definecolor{custom_person}{RGB}{255, 30, 30}
\definecolor{custom_bicyclist}{RGB}{255, 40, 20}
\definecolor{custom_motorcyclist}{RGB}{150, 30, 90}
\definecolor{custom_fence}{RGB}{255, 120, 50}
\definecolor{custom_pole}{RGB}{255, 240, 150}
\definecolor{custom_traf_sign}{RGB}{255, 0, 0}
\definecolor{custom_other_struct}{RGB}{255, 150, 0}
\definecolor{custom_other_obj}{RGB}{50, 255, 255}

\begin{table*}[hbt!]
\centering
\resizebox{\textwidth}{!}{%
\begin{tabular}{lccc|cccccccccccccccccccccccccccccccccccc}
\toprule
 &
   &
  \multicolumn{1}{l}{} &
   &
  \multicolumn{2}{c}{} &
  \multicolumn{2}{c}{} &
  \multicolumn{2}{c}{} &
  \multicolumn{2}{c}{} &
  \multicolumn{2}{c}{} &
  \multicolumn{2}{c}{} &
  \multicolumn{2}{c}{} &
  \multicolumn{2}{c}{} &
  \multicolumn{2}{c}{} &
  \multicolumn{2}{c}{} &
  \multicolumn{2}{c}{} &
  \multicolumn{2}{c}{} &
  \multicolumn{2}{c}{} &
  \multicolumn{2}{c}{} &
  \multicolumn{2}{c}{} &
  \multicolumn{2}{c}{} &
  \multicolumn{2}{c}{} &
  \multicolumn{2}{c}{} \\
 &
   &
   &
   &
  \multicolumn{2}{c}{\rotatebigtext{car} \newline \textcolor{white}{\rotatetinytext{1}} \newline \rotatetinytext{(2.85\%)}} &
  \multicolumn{2}{c}{\rotatebigtext{bicycle} \newline \textcolor{white}{\rotatetinytext{1}} \newline \rotatetinytext{(0.01\%)}} &
  \multicolumn{2}{c}{\rotatebigtext{motorcycle} \newline \textcolor{white}{\rotatetinytext{1}}\newline \rotatetinytext{(0.01\%)}} &
  \multicolumn{2}{c}{\rotatebigtext{truck} \newline \textcolor{white}{\rotatetinytext{1}}\newline \rotatetinytext{(0.16\%)}} &
  \multicolumn{2}{c}{\rotatebigtext{other-veh.} \newline \textcolor{white}{\rotatetinytext{1}}\newline \rotatetinytext{(5.75\%)}} &
  \multicolumn{2}{c}{\rotatebigtext{person} \newline \textcolor{white}{\rotatetinytext{1}}\newline \rotatetinytext{(0.02\%)}} &
  \multicolumn{2}{c}{\rotatebigtext{road} \newline \textcolor{white}{\rotatetinytext{1}}\newline \rotatetinytext{(14.98\%)}} &
  \multicolumn{2}{c}{\rotatebigtext{parking} \newline \textcolor{white}{\rotatetinytext{1}}\newline \rotatetinytext{(2.31\%)}} &
  \multicolumn{2}{c}{\rotatebigtext{sidewalk} \newline \textcolor{white}{\rotatetinytext{1}}\newline \rotatetinytext{(6.43\%)}} &
  \multicolumn{2}{c}{\rotatebigtext{other-grnd.} \newline \textcolor{white}{\rotatetinytext{1}}\newline \rotatetinytext{(2.05\%)}} &
  \multicolumn{2}{c}{\rotatebigtext{building} \newline \textcolor{white}{\rotatetinytext{1}}\newline \rotatetinytext{(15.67\%)}} &
  \multicolumn{2}{c}{\rotatebigtext{fence} \newline \textcolor{white}{\rotatetinytext{1}}\newline \rotatetinytext{(0.96\%)}} &
  \multicolumn{2}{c}{\rotatebigtext{vegetation} \newline \textcolor{white}{\rotatetinytext{1}}\newline \rotatetinytext{(41.99\%)}} &
  \multicolumn{2}{c}{\rotatebigtext{terrain} \newline \textcolor{white}{\rotatetinytext{1}}\newline \rotatetinytext{(7.10\%)}} &
  \multicolumn{2}{c}{\rotatebigtext{pole} \newline \textcolor{white}{\rotatetinytext{1}}\newline \rotatetinytext{(0.22\%)}} &
  \multicolumn{2}{c}{\rotatebigtext{traf.-sign} \newline \textcolor{white}{\rotatetinytext{1}}\newline \rotatetinytext{(0.06\%)}} &
  \multicolumn{2}{c}{\rotatebigtext{other-struct.} \newline \textcolor{white}{\rotatetinytext{1}}\newline \rotatetinytext{(4.33\%)}} &
  \multicolumn{2}{c}{\rotatebigtext{other-obj.} \newline \textcolor{white}{\rotatetinytext{1}}\newline \rotatetinytext{(0.28\%)}}
   \\
  \textbf{Method} &
  \textbf{Input} &
  \textbf{IoU} &
  \textbf{mIoU} &
  \multicolumn{2}{c}{\colorboxlabel{custom_car}{}} &
  \multicolumn{2}{c}{\colorboxlabel{custom_bicycle}{}} &
  \multicolumn{2}{c}{\colorboxlabel{custom_motorcycle}{}} &
  \multicolumn{2}{c}{\colorboxlabel{custom_truck}{}} &
  \multicolumn{2}{c}{\colorboxlabel{custom_other_veh}{}} &
  \multicolumn{2}{c}{\colorboxlabel{custom_person}{}} &
  \multicolumn{2}{c}{\colorboxlabel{custom_road}{}} &
  \multicolumn{2}{c}{\colorboxlabel{custom_parking}{}} &
  \multicolumn{2}{c}{\colorboxlabel{custom_sidewalk}{}} &
  \multicolumn{2}{c}{\colorboxlabel{custom_other_ground}{}} &
  \multicolumn{2}{c}{\colorboxlabel{custom_building}{}} &
  \multicolumn{2}{c}{\colorboxlabel{custom_fence}{}} &
  \multicolumn{2}{c}{\colorboxlabel{custom_vegetation}{}} &
  \multicolumn{2}{c}{\colorboxlabel{custom_terrain}{}} &
  \multicolumn{2}{c}{\colorboxlabel{custom_pole}{}} &
  \multicolumn{2}{c}{\colorboxlabel{custom_traf_sign}{}} &
  \multicolumn{2}{c}{\colorboxlabel{custom_other_struct}{}} &
  \multicolumn{2}{c}{\colorboxlabel{custom_other_obj}{}} \\ \midrule
    \multicolumn{40}{c}{\textbf{\textit{Single-Frame-Based}}} \\ \midrule

\multicolumn{1}{l|}{Symphonies} &
  \multicolumn{1}{c|}{Stereo} &
   34.39&
   11.93&
  \multicolumn{2}{c}{18.04} &
  \multicolumn{2}{c}{4.63} &
  \multicolumn{2}{c}{\underline{3.16}} &
  \multicolumn{2}{c}{12.70} &
  \multicolumn{2}{c}{3.07} &
  \multicolumn{2}{c}{2.83} &
  \multicolumn{2}{c}{43.15} &
  \multicolumn{2}{c}{10.44} &
  \multicolumn{2}{c}{27.96} &
  \multicolumn{2}{c}{4.15} &
  \multicolumn{2}{c}{27.05} &
  \multicolumn{2}{c}{3.66} &
  \multicolumn{2}{c}{27.09} &
  \multicolumn{2}{c}{14.01} &
  \multicolumn{2}{c}{8.82} &
  \multicolumn{2}{c}{8.10} &
  \multicolumn{2}{c}{4.69} &
  \multicolumn{2}{c}{3.17} \\
\multicolumn{1}{l|}{CGFormer} &
  \multicolumn{1}{c|}{Stereo} &
   44.72&
   \underline{15.61}&
  \multicolumn{2}{c}{22.26} &
  \multicolumn{2}{c}{\underline{4.76}} &
  \multicolumn{2}{c}{1.67} &
  \multicolumn{2}{c}{\underline{21.28}} &
  \multicolumn{2}{c}{2.58} &
  \multicolumn{2}{c}{\underline{3.91}} &
  \multicolumn{2}{c}{\underline{55.11}} &
  \multicolumn{2}{c}{14.17} &
  \multicolumn{2}{c}{\underline{35.15}} &
  \multicolumn{2}{c}{4.60} &
  \multicolumn{2}{c}{37.55} &
  \multicolumn{2}{c}{5.12} &
  \multicolumn{2}{c}{\underline{35.68}} &
  \multicolumn{2}{c}{21.62} &
  \multicolumn{2}{c}{11.01} &
  \multicolumn{2}{c}{8.96} &
  \multicolumn{2}{c}{7.29} &
  \multicolumn{2}{c}{\underline{3.77}}\\
\multicolumn{1}{l|}{ScanSSC} &
  \multicolumn{1}{c|}{Stereo} &
   \underline{45.09}&
   15.44&
  \multicolumn{2}{c}{\underline{22.47}} &
  \multicolumn{2}{c}{3.85} &
  \multicolumn{2}{c}{0.90} &
  \multicolumn{2}{c}{16.18} &
  \multicolumn{2}{c}{\underline{4.93}} &
  \multicolumn{2}{c}{\textbf{4.03}} &
  \multicolumn{2}{c}{52.68} &
  \multicolumn{2}{c}{\underline{15.85}} &
  \multicolumn{2}{c}{34.52} &
  \multicolumn{2}{c}{\underline{5.01}} &
  \multicolumn{2}{c}{\underline{37.95}} &
  \multicolumn{2}{c}{\underline{5.42}} &
  \multicolumn{2}{c}{35.58} &
  \multicolumn{2}{c}{\underline{22.07}} &
  \multicolumn{2}{c}{\underline{11.34}} &
  \multicolumn{2}{c}{\underline{9.69}} &
  \multicolumn{2}{c}{\underline{7.66}} &
  \multicolumn{2}{c}{3.21} \\ \midrule
    \multicolumn{40}{c}{\textbf{\textit{Temporal-Frame-Based}}} \\ \midrule
\rowcolor[HTML]{EFEFEF} 
\multicolumn{1}{l|}{\cellcolor[HTML]{EFEFEF}\textbf{Ours}} &
  \multicolumn{1}{c|}{\cellcolor[HTML]{EFEFEF}\textbf{Stereo}} &
   \textbf{52.41}&
   \textbf{23.72}&
  \multicolumn{2}{c}{\cellcolor[HTML]{EFEFEF} \textbf{34.21}} & 
  \multicolumn{2}{c}{\cellcolor[HTML]{EFEFEF} \textbf{10.24}} &
  \multicolumn{2}{c}{\cellcolor[HTML]{EFEFEF} \textbf{10.71}} &
  \multicolumn{2}{c}{\cellcolor[HTML]{EFEFEF} \textbf{27.95}} &
  \multicolumn{2}{c}{\cellcolor[HTML]{EFEFEF} \textbf{8.75}} &
  \multicolumn{2}{c}{\cellcolor[HTML]{EFEFEF} 3.88} &
  \multicolumn{2}{c}{\cellcolor[HTML]{EFEFEF} \textbf{60.84}} &
  \multicolumn{2}{c}{\cellcolor[HTML]{EFEFEF} \textbf{20.60}} &
  \multicolumn{2}{c}{\cellcolor[HTML]{EFEFEF} \textbf{42.49}} &
  \multicolumn{2}{c}{\cellcolor[HTML]{EFEFEF} \textbf{6.62}} &
  \multicolumn{2}{c}{\cellcolor[HTML]{EFEFEF} \textbf{45.99}} &
  \multicolumn{2}{c}{\cellcolor[HTML]{EFEFEF} \textbf{12.80}} &
  \multicolumn{2}{c}{\cellcolor[HTML]{EFEFEF} \textbf{45.63}} &
  \multicolumn{2}{c}{\cellcolor[HTML]{EFEFEF} \textbf{30.29}} &
  \multicolumn{2}{c}{\cellcolor[HTML]{EFEFEF} \textbf{29.00}} &
  \multicolumn{2}{c}{\cellcolor[HTML]{EFEFEF} \textbf{31.08}} &
  \multicolumn{2}{c}{\cellcolor[HTML]{EFEFEF} \textbf{16.84}} &
  \multicolumn{2}{c}{\cellcolor[HTML]{EFEFEF} \textbf{12.70}} \\ \bottomrule
\end{tabular}
}

\caption{Quantitative results on out-of-view regions in the SSCBench-KITTI-360 test set. \textbf{Bold} / \underline{underline} highlight the best / second-best, respectively.}
\label{tab:KITTI360 _oov_test}
    
\end{table*}

%% file: Sections/Table/sup_voxel_aggregation.tex
\begin{table}[hbt!]
\centering
\setlength\tabcolsep{8pt}\resizebox{.7\linewidth}{!}{
\begin{tabular}{@{}cll|clcl@{}}
\toprule
\multicolumn{3}{c|}{\textbf{Voxel Aggregation Method}}               & \multicolumn{2}{c}{\textbf{IoU}}                           & \multicolumn{2}{c}{\textbf{mIoU}}                          \\ \midrule
\multicolumn{3}{c|}{Concat$\rightarrow$Linear}                                  & \multicolumn{2}{c}{48.85}                                  & \multicolumn{2}{c}{18.41}                                  \\
\multicolumn{3}{c|}{Concat$\rightarrow$Conv.}                                  & \multicolumn{2}{c}{49.02}                                  & \multicolumn{2}{c}{18.39}                                  \\
\multicolumn{3}{c|}{{Learnable Weight}} & \multicolumn{2}{c}{{49.08}} & \multicolumn{2}{c}{{18.79}} \\
\multicolumn{3}{c|}{Dynamic Weighted Sum}                                  & \multicolumn{2}{c}{49.22}                                  & \multicolumn{2}{c}{18.97}                                  \\
\multicolumn{3}{c|}{\cellcolor[HTML]{EFEFEF}\textbf{Average}}                                  & \multicolumn{2}{c}{\cellcolor[HTML]{EFEFEF}\textbf{49.53}}                                  & \multicolumn{2}{c}{\cellcolor[HTML]{EFEFEF}\textbf{19.31}}                                  \\ \bottomrule
\end{tabular}}
\caption{Ablation study on the voxel aggregation method.}
\label{tab:ablation_temp_fusion}
\end{table}